\documentclass{cup-pan}
\usepackage[utf8]{inputenc}
\usepackage{booktabs}
\usepackage{blindtext}
\usepackage[export]{adjustbox}[2011/08/13]
\usepackage{makecell}

\usepackage{fancyhdr,graphicx,amsmath,amssymb}
\usepackage[ruled,linesnumbered]{algorithm2e}
\usepackage{booktabs}
\usepackage{tablefootnote}
\usepackage{natbib}
\usepackage{caption}

\newcommand{\descrcell}[2]{%
  \scriptsize
  \begin{tabular}[t]{@{}c@{}}\normalsize#1\\\normalsize#2\end{tabular}%
}

\title{Truly Sparse Neural Networks at Scale}

\author[1]{Selima Curci}
\author[2,1]{Decebal Constantin Mocanu}
\author[1]{Mykola Pechenizkiyi}

\affil[1]{ Department of Mathematics and Computer Science, Eindhoven University of
Technology}
\affil[2]{Faculty of Electrical Engineering, Mathematics and Computer Science, University
of Twente}

\begin{document}
\maketitle

\begin{abstract}
Recently, sparse training methods have started to be established as a de facto approach for training and inference efficiency in artificial neural networks. Yet, this training efficiency is just in theory. In practice, everyone uses a binary mask to simulate sparsity since the typical deep learning software and hardware are optimized for dense matrix operations. In this paper, we take an orthogonal approach, and we show that we can train truly sparse neural networks to harvest their full potential. To achieve this goal, we introduce three novel contributions \footnote{The code is available at \href{https://github.com/SelimaC/large-scale-sparse-neural-networks}{https://github.com/SelimaC/large-scale-sparse-neural-networks}}, specially designed for sparse neural networks: (1) a parallel training algorithm and its corresponding sparse implementation from scratch, (2) an activation function with non-trainable parameters to favour the gradient flow, and (3) a hidden neurons importance metric to eliminate redundancies. All in one, we are able to break the record and to train the largest neural network ever trained in terms of representational power -- reaching the bat brain size. The results show that our approach has state-of-the-art performance while opening the path for an environmentally friendly artificial intelligence era.

\end{abstract}

\section{Introduction}
\label{sec:intro}
Artificial Neural Networks (ANNs) succeeded in a broad range of application domains (\cite{deeplearningbengionature}) due to their ability to learn complex transformations from data while achieving superior generalisation performance. However, current state-of-the-art networks are typically highly overparameterised (e.g. \cite{zhang2016understanding}) and demand extensive computational resources to be trained, which become a bottleneck where such resources are limited (\cite{Kepner_2018}). Reducing the memory footprint and training time of ANNs are active areas of research, crucial to handle the rapid expansion of machine learning which has resulted in enormous datasets, with millions to billions of examples and features, but also to decrease the high environmental impact of the energy-hungry deep learning algorithms. Taking inspiration from nature, a solution to improve neural network scaling is to use sparse connectivity. The traditional \textit{dense-to-sparse} training paradigm (known mainly as network pruning) \citep{Mozer, braindamage, han2015learning, frankle2018lottery} offer computational benefits just in the inference phase as first, it trains a dense network in order to prune unimportant connections and to obtain a sparsely connected neural network.

Therefore, to obtain scalable and efficient ANNs, contrary to general practice, artificial neural networks, like biological neural networks, should not have fully connected layers also in the training phase. Recently, a new \textit{sparse-to-sparse} training paradigm (or simply, sparse training) began to establish inside the research community, with several studies focus on developing memory and computational efficiency from the start by training directly sparse neural networks from scratch. The first attempt \citep{Mocanu2016xbm} has used just static sparsity limiting the capacity of the model sparse connectivity graph to fit the data distribution. This concept has been revised and drastically improved by introducing the Sparse Evolutionary Training (SET) algorithm with dynamic (or adaptive) sparsity in (\cite{Mocanu_2018}). Currently, the sparse training concept has started to be a \textit{de facto} approach for efficient training of ANNs, as demonstrated in \citep{Bellec2017DeepRT,  Dettmers2019SparseNF, mostafa2019parameter, Evci2019RiggingTL, Evci2020GradientFI, top, NEURIPS2020_b4418237}. These algorithms search for an optimal \textit{sparse} \textit{topology} according to some salience criteria, while simultaneously optimising the model weights. Here, the \textit{topology} of a neural network refers to the way the neurons are connected, and it is a crucial factor in network functioning, and learning (\cite{Miikkulainen2010}). The resulting networks have a significantly lower number of parameters by design, and they have empirically shown to outstretch higher generalisation power than their dense counterparts in a number of cases, especially in the case of multilayer perceptrons and recurrent neural networks (\cite{Evci2020GradientFI, Bourgin2019CognitiveMP, liu2020topological, Liu2021SelfishSR}). Besides this, intrinsically sparse models allow, in theory, real scalable deep learning solutions in low-resource devices, standard computers, and in the cloud.

The main limitation to achieve this theoretical scalability level is given by the fact that all state-of-the-art deep learning frameworks are based on very well-optimised dense matrix multiplications on Graphics Processing Units (GPUs), while sparse matrix operations are practically ignored. The only notable exception is given by the NVIDIA A100 GPU which was released in 2020 (\cite{nvidiaampere}) and supports a hardware fixed 2:4 sparsity level (i.e. 50\% sparsity level). Within these frameworks, one can only simulate the sparsity by using a binary mask over the connections; therefore, the model will carry on training dense matrices. As follows, until optimised hardware for sparse operations appears, one would have to focus on optimising the algorithms.

In (\cite{liu2019sparse}), the authors developed an efficient implementation of sparse multilayer perceptrons (MLPs) trained with SET. For the first time, they built sparse MLP models with over one million artificial neurons on commodity hardware, only utilising one CPU core. Still their sparse framework is completely sequential and cannot yet compete against advanced professional frameworks designed to accelerate the learning of dense neural networks.
\begin{figure}[!htbp]
\centering
\includegraphics[width=0.9\textwidth]{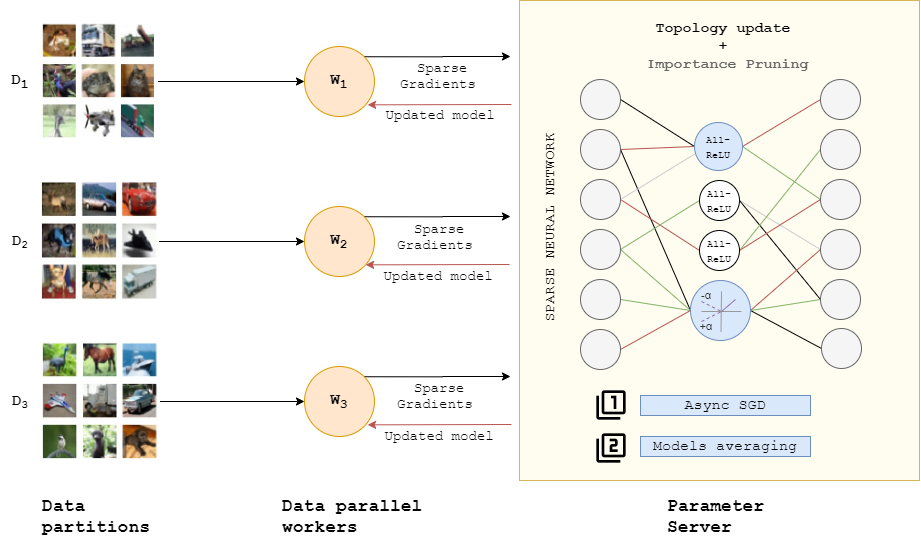}
\vskip 0.25cm
\captionsetup{justification=centering}
\caption{A graphical high-level overview of the proposed methods to efficiently train truly sparse neural networks. }
\label{fig:overview}
\end{figure}

Additionally, there is the need for revisiting various aspects (e.g. optimizers and activation functions) of sparse neural networks training since most of the literature has mainly focused on dense models. The choice of the activation function for deep neural networks has a critical impact on the performance of the training procedure. An inappropriate selection can lead to the loss of information of the input during forward propagation and the exponential vanishing/exploding of gradients during backpropagation (\cite{ hayou2019impact}). It is crucial to question weather the activation functions currently used for densely connected networks still behave reliably in the sparse context. SReLU is a relatively little-known activation function suggested in (\cite{jin2015deep}) and it has proven to outperform ReLU (\cite{relu}) when training sparse networks over different datasets (\cite{Mocanu_2018, adam}) as it improves the networks gradient flow (\cite{Tessera2021KeepTG}). However, this activation function requires to learn four additional parameters per neuron, which becomes a non-negligible number and introduces a serious computational overhead if we want to train models with millions or billions of hidden units.

To alleviate the aforementioned limitations, this paper proposes four new contributions which advance the scalability of neural networks by exploiting sparsity:
\begin{itemize}
\item We introduce \textit{Weight Averaging Sparse Asynchronous Parallel SGD} (\textit{WASAP-SGD}), a parallel algorithm to train truly sparse neural networks and expand their feasible size on commodity hardware, without any GPU support.
\item We propose a variant of ReLU, called \textit{ALternated Left ReLU} (\textit{All-ReLU}), to achieve performance comparable to SReLU for SET without the additional overhead for training its associated parameters. 
\item We introduce the concept of neuron importance and a method (\textit{Importance Pruning}) to blend it into the sparse training procedure, which allows us to shrink even more the number of weights and to accelerate sparse training considerably.
\item We developed a customised and modularized software framework for sparse neural networks to test our theoretical contributions. It allows us to break the record and to train as a proof-of-concept the largest neural network model ever trained, i.e. 50 million neurons.
\end{itemize}
The three approaches (\textit{WASAP-SGD}, \textit{All-ReLU} and \textit{Importance Pruning}) represent independent contributions to sparse neural network literature, but also they can be used together as complementary methods to improve further the performance of sparse models, as we illustrate in our experimental results.

\section{Results}
\label{sec:results}
Our work is focused on the most straightforward type of neural networks, MLPs, as they count for 61\% of a typical Google TPU (Tensor Processing Unit) workload for production neural networks applications, while convolutional neural networks represent merely 5\% (\cite{jouppi2017indatacenter}). Despite the numerous algorithms available to train sparse neural networks from scratch, we decided to base our evaluation on the SET algorithm, given its simplicity and good performance on a broad range of domains. Unlike the other sparse training techniques mentioned in \autoref{subsubsection:sparsetosparse} that calculate and store information for all the parameters, including the non-existing ones, SET is memory-efficient because it uses information just from the existing parameters, and it does not require high computational complexity. These are all favourable advantages to our goal of developing large scale sparse neural networks. We evaluate and discuss the performance of our proposed methods on sparse MLP models by considering five publicly available datasets listed in Table \ref{table:datasets}.

\subsubsection*{Problem formulation}
Given a dataset $\mathcal{D} = {(x_i, y_i)}_{i=1}^n$ and a network $f(x; \theta)$ with $L$ layers parameterized by
$\theta$ (weights $\mathbf{w}$ and biases $\mathbf{b}$). We train the network to minimize the loss function $ \sum L(f(x; \theta), y)$. The motivation of sparse neural networks is to use a fraction of parameters to reparameterize the whole network, while preserving the performance as much as possible. Hence, a sparse neural network can be denoted as $f_s(x; \theta_s)$ with a given sparsity level. Initially, the network is uniformly initialized with a sparse distribution in which the sparsity level $\mathcal{S}_l$ of each layer $l$ is controlled by a parameter $\epsilon$ (see \cite{Mocanu_2018} for details) and stays constant during the training. More precisely, for each layer $l$ the connections are defined in a sparse adjacency weights matrix $\mathbf{W}^{(l)} = [[w_{11}, w_{12}, \dots, w_{1n_{l}}], \dots, [w_{n_{l-1}1}, w_{n_{l-1}2}, \dots, w_{n_{l-1}n_{l}}]]$ in which the elements are either \textit{null} ($w_{ij} = 0$) when
there is no connection between neuron $i$ and neuron $j$ or have a connection weight ($w_{ij} \neq 0$) when the connection between i and j exists. Initially each $\mathbf{W}^{(l)}$ is a Erd\H{o}s-Rényi random graph (\cite{Erdos:1959:pmd}).

\subsection{Proposed contributions}

\subsubsection*{WASAP-SGD method}
We propose a novel parallel training method with two phases based on a data parallelism strategy (where the learning phase of a model is partitioned by input samples) to improve the scalability of sparse neural networks. The algorithm, called \textbf{WASAP-SGD}, is based on \textit{asynchronous SGD} (\cite{dist}) training for the first phase and \textit{Stochastic Weight Averaging} (\cite{Izmailov2018AveragingWL}) for the second phase. This two-phase method helps in filling the gap between sparse and dense neural networks' performances (accuracy, running time and generalization).

\begin{figure}[!htbp]
\centering
\includegraphics[width=0.5\textwidth]{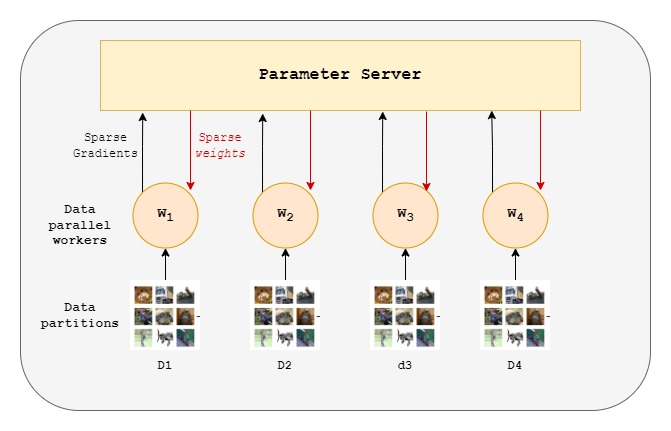}
\caption{Sparse model replicas asynchronously fetch sparse parameters and push sparse gradients to the parameter server with atomic read and write operations. }
\label{fig:parameter server}
\end{figure}

We consider a system with \textit{K} workers, which repeatedly compute gradient contributions based on independently drawn data mini-batches from the given dataset  $\mathcal{D}$. We also consider a \textit{shared parameter server}, which communicates with each of the workers independently, to give state information and get updates that it applies according to the algorithm it follows. The master and each worker have a replica of the sparse model to be trained. Moreover, each worker has access to a subset of the training data, as depicted in Figure \ref{fig:parameter server}.

\begin{figure}[!htbp]
\centering
\includegraphics[width=0.7\textwidth]{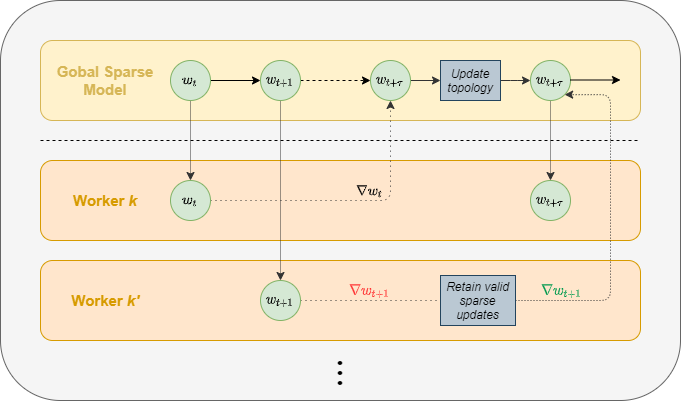}
\vskip 0.25cm
\captionsetup{justification=centering}
\caption{Worker \textit{k'} fetches parameters $w_{t+1}$ and push gradients
$\Delta w_{t+1}$ to the parameter server. These gradients may contain non-valid updates, since in that time frame the global model may have performed the topology evolution, hence they need to be corrected. }
\label{fig:retain}
\end{figure}

In \autoref{algorithm:wasap} we show the pseudocode for \textit{WASAP-SGD}, describing how standard \textit{asynchronous SGD} using a parameter server is extended with a local training phase followed by a \textit{sparse} model averaging step to improve its generalization performance. Moreover, it is designed to include the topology adaptation step of sparse networks. The training is carried out \textit{asynchronously} by all workers. We adopt a simple SGD update rule with momentum, which has shown to be effective for training intrinsically sparse models:
\begin{equation}
\mathbf{w_{t+1}} = \mathbf{w_t} + \mu (\mathbf{w_{t}} - \mathbf{w_{t-1}}) - \eta_t \nabla \mathbf{w_t}
\label{sgdmomentum}
\end{equation}
The master must periodically pause the asynchronous update to carry on the weight evolution algorithm on the sparse model to generate the new topology. Each update must include a minor modification, since individual weights may be outdated due to the topology evolution (as illustrated in \autoref{fig:retain}).

Then, to improve the model generalization performance, each worker locally updates its sparse replica for the next phase (\textit{phase two}). Once \textit{phase two} is concluded, the $K$ different models are averaged:
\begin{equation}
\theta_{s}^f = \frac{1}{K} \sum_{i=1}^{K} \theta_s^i
\end{equation}
The averaging step does not preserve the sparsity level ${S}$, since each worker has updated its topology independently from each other. Hence, the final model $\theta_{s}^f$ will have a different sparsity level $\mathcal{S}'_{(l)}$, for each layer $l$, where $\mathcal{S}'_{(l)} \ge \mathcal{S}, \forall l=1, \dots, L$ . Thus, unimportant connections, accounting for a certain fraction $\mathcal{S}'_{(l)} - \mathcal{S}$, will be pruned in each layer. More precisely, the unimportant connections are pruned based on their magnitude, corresponding to the largest
negative weights and the smallest positive weights in $\mathbf{W}^{(l)}$.

\subsubsection*{All-ReLU}
The new proposed activation function, \textbf{All-ReLU} (Alternated Left ReLU), is designed for training sparse MLPs and is able to accelerate training, without adding any additional computational complexity. All-ReLU is inspired by the S-shaped rectified linear activation unit (SReLU) presented in \cite{jin2015deep}. The intuition behind it came from analysing the SReLU parameters as well as the input distribution of the learned topology. Like SReLU, this function can improve the networks gradient flow, and consequently achieve better accuracy. However, since All-ReLU does not require to train additional parameters, it can be considered as simple and fast as ReLU to use.

Given an ANN with $L$ layers, our proposed Alternate Left ReLU (All-ReLU) is defined as follows for each layer $l$:
\begin{equation}
f_l(x_i) = \begin{cases} 
- \alpha x_i & x_i\leq 0 \And l \% 2 == 0 \\
\alpha x_i & x_i\leq 0 \And l \% 2 == 1 \\
x_i & x_i > 0 
\end{cases}
\end{equation}
where $x_i$ is the input value, $\alpha$ is the slope for the negative side of the input and $\%$ represent the \textit{modulo} operation. The input layer ($l=1$) and the output layer ($l=L$) are excluded. We believe that the proposed activation function can accelerate convergence by breaking symmetry during training and preserving the gradient flow through the network, hence leading to better performance for sparse models.

\subsubsection*{Importance Pruning}
To substantially reduce the size of neural networks, we propose a novel method, for selecting the most important neurons, based on their strength (importance). In graph theory, the node strength is the sum of weights of links connected to the node (\cite{graphteory}). Taking inspiration from this graph measure, we determine the importance of each neuron based on the summation of absolute weights of its incoming connection. For each neuron $j$ in layer $l$ we define its importance as follows:

\begin{equation}
I_j^{(l)} = \sum_{i \in \Gamma_j^{(l-1)}} |w_{ij}^{(l)}|
\end{equation}
where $\Gamma_j^{(l-1)}$ is the set of all neurons connected to neuron $j$, i.e. $\Gamma_j^{(l-1)} = \{i | 1 \le i \le n^{(l-1)}, \forall i \in \mathbb{N} \hspace{5pt} \land \hspace{5pt} w_{ij}^{(l)} \neq 0\}$, given that $n^{(l-1)}$ is the number of hidden neurons from the previous layer $l-1$ and $w_{ij}^{(l)}$ denotes the weight of connection linking neuron $i$ to $j$ in two consecutive layers as defined in $\mathbf{W}^{(l)}$.
This neuron importance metric can rapidly identify the main \textit{hubs} of the sparse network, i.e. nodes that are positioned to make strong contributions to global network performance.

This metric can be easily integrated during training and we named this procedure \textit{Importance Pruning}, once the topology is stable, to reduce overfitting with almost no loss in accuracy and substantially lessen the number of parameters up to 80\% with respect to the initial sparse network and, as a consequence, decrease the overall training running time. In \autoref{algorithm:pruning}, we provide an example of how \textit{Importance Pruning} can be integrated with dynamic sparse training. The pseudocode refers to the SET algorithm; however, it could be easily replaced by any other sparse-to-sparse training technique.

\subsubsection*{Large scale Sparse Neural Network framework}
We extended the sparse framework presented in \cite{liu2019sparse} by implementing the theoretical contributions presented in this paper. The initial implementation was sequential and it was not able to obtain the same accuracy as Keras on some datasets such as CIFAR10. Our focus was on MLP as we did not have the human resources to develop RNNs and CNNs from scratch and we let this as future work. With respect to the speed of our approaches, it is worth to highlight that first we substantially improved the training time of a truly sparse MLP from its previous implementation in \cite{liu2019sparse} with no parallelisation by replacing Cython with Numba (\cite{numba}) and adopting \texttt{32-bit} float precision instead of \texttt{64-bit}. These minor changes ensure minimum resource requirements. WASAP-SGD is designed using Python 3.7, one of the most popular Machine Learning languages, combined with Message Passing Interface (MPI). 

With our framework, we were able to build the largest Sparse MLP model, in terms of the number of neurons, ever trained on a single machine on the cloud (with no GPU). Note that the enormous models mentioned in literature have been usually trained in a distributed fashion or on several GPUs. Since a high-dimensional dataset for this task is hard to find, we use an artificial one to train a model with 50 million neurons on a machine with 96 virtual cores and 768 GB of RAM. This experiment demonstrates how the Sparse MLP can achieve what its dense counterpart cannot, due to memory error.

\begin{table}[!htpb]
\centering
\resizebox{0.8\textwidth}{!}{%
\begin{tabular}{lllllllll}
\toprule
\textbf{Experiment} & \textbf{Dataset} & \textbf{Dataset properties} & & \\ \cmidrule{3-9}
& & \textbf{Domain} & \textbf{Features} & 
\textbf{\descrcell{Train}{samples [\#]}} & 
\textbf{\descrcell{Train}{samples [MB]}} &
\textbf{\descrcell{Test}{samples [\#]}} & 
\textbf{\descrcell{Test}{samples [MB]}} &
\textbf{Classes} \\ \midrule
\textbf{MLPs} & Leukemia  \citep{ncbi} & Microarray & 54675 & 1397 & & 699 & & 18\\
& Higgs \citep{higgs} & Physics particles & 28 & 105000 & 12 & 50000 & 6 & 2 \\
& Madelon  \citep{madelon} & Artificial data & 500 & 2000 & & 600 & & 2 \\ 
& FashionMNIST \citep{fashionmnist} & Images & 784 & 60000 & 188 & 10000 & 32 & 10 \\ 
& CIFAR10 \citep{cifar10} & RGB Images & 3072 & 50000 & 615 & 10000 & 123 & 10 \\ 
\bottomrule
\end{tabular}
}
\vskip 0.25cm
\captionsetup{justification=centering}
\caption{List of dataset used for the experiments.}
\label{table:datasets}
\end{table}

\subsection{Performance on Sequential Trained Sparse MLPs}
\label{subsection:mlps}
\begin{figure}[!htbp]
\centering
   \subfloat[\label{figure:memor_vs_error_train} Train results]{%
   \centering
      \includegraphics[ width=0.5\textwidth]{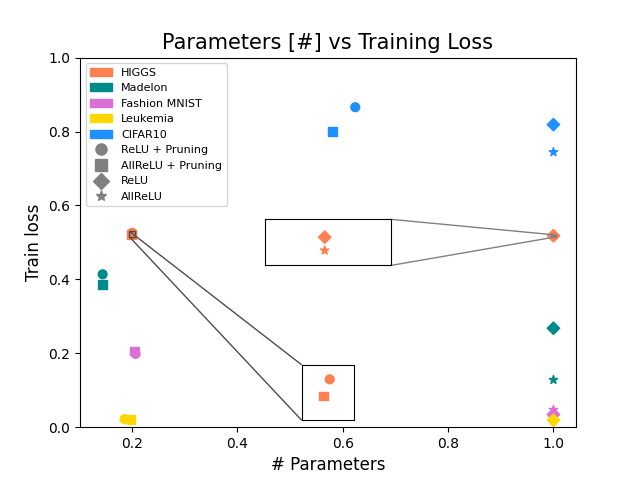}}
   \subfloat[\label{figure:memor_vs_error_test} Test results]{%
   \centering
      \includegraphics[ width=0.5\textwidth]{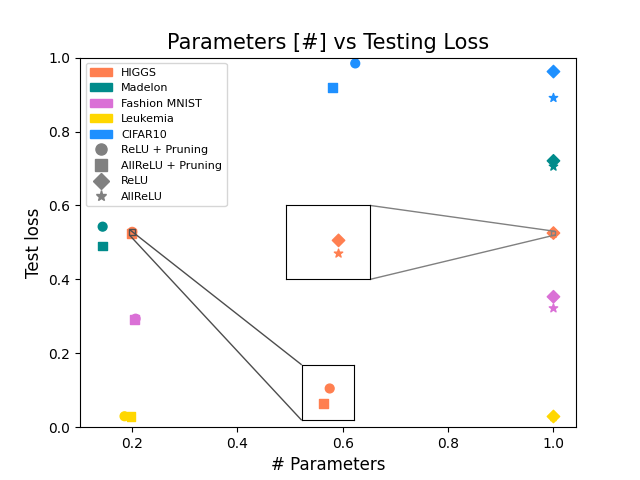}}
	\captionsetup{justification=centering}
\caption{Relative model size (expressed in number of parameters [\#]) against the relative error (loss) to highlight the differences with and without \textit{Importance Pruning}.}
\label{figure:memor_vs_error}
\end{figure}

\begin{figure}	
	\centering
	\begin{subfigure}[t]{0.9\textwidth}
		\centering
		\includegraphics[width=0.9\textwidth]{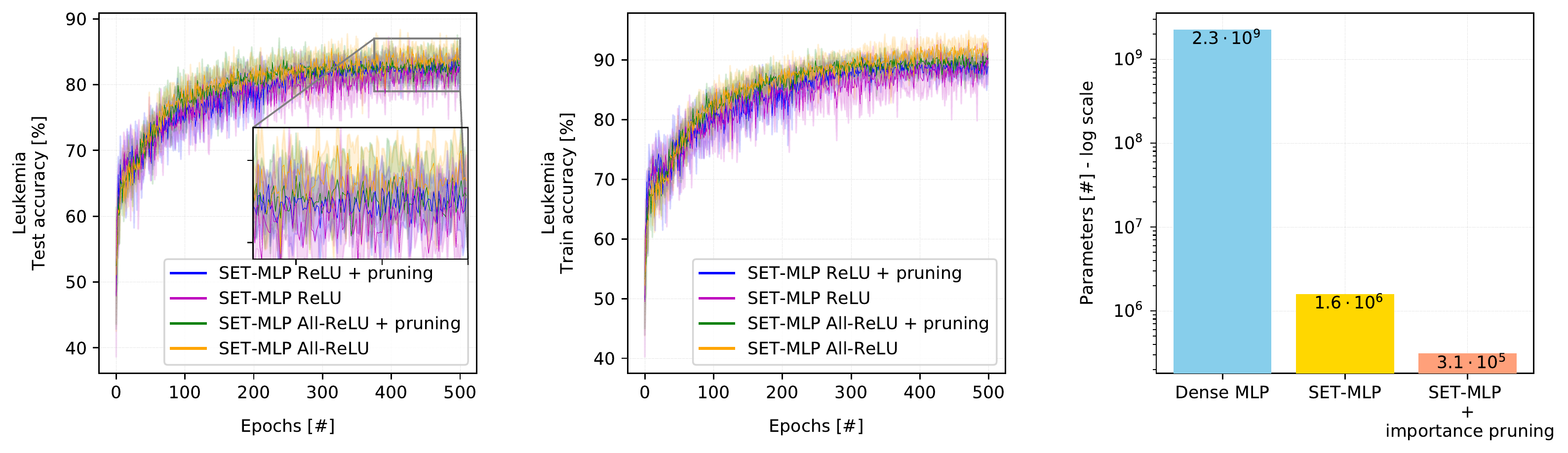}
		\caption{Learning curves and parameter numbers comparison for Leukemia dataset.}\label{fig:1a_evaluation}		
	\end{subfigure}
	\vspace{2mm}
	\begin{subfigure}[t]{0.9\textwidth}
		\centering
		\includegraphics[width=0.9\textwidth]{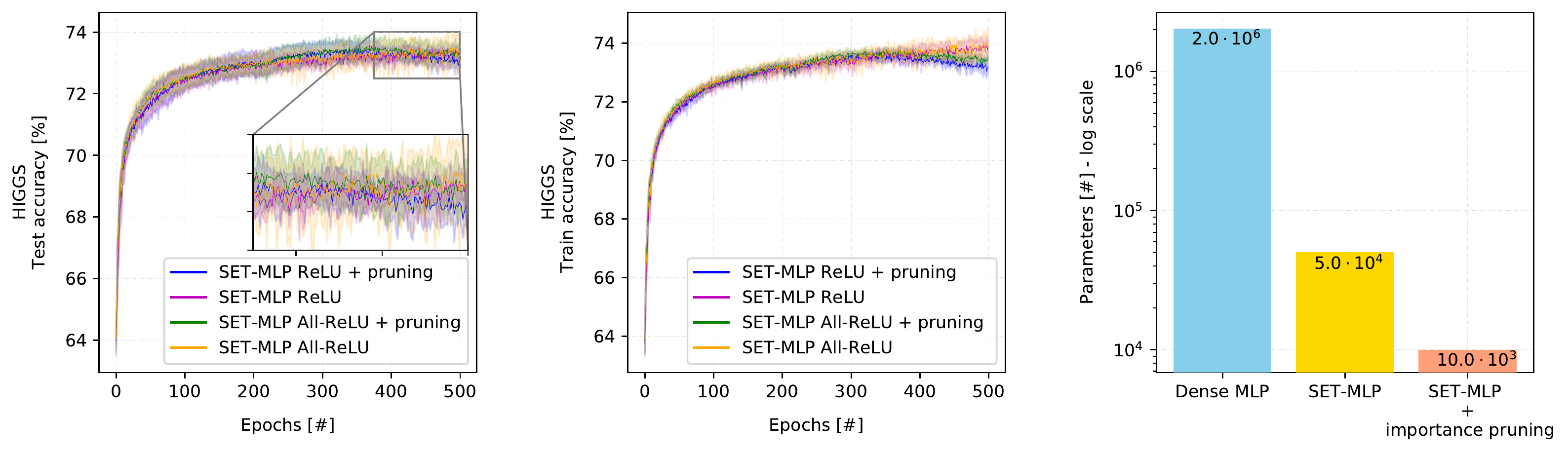}
		\caption{Learning curves and parameter numbers comparison for HIGGS dataset.}\label{fig:1b_evaluation}
	\end{subfigure}
		\vspace{2mm}
	\begin{subfigure}[t]{0.9\textwidth}
		\centering
		\includegraphics[width=0.9\textwidth]{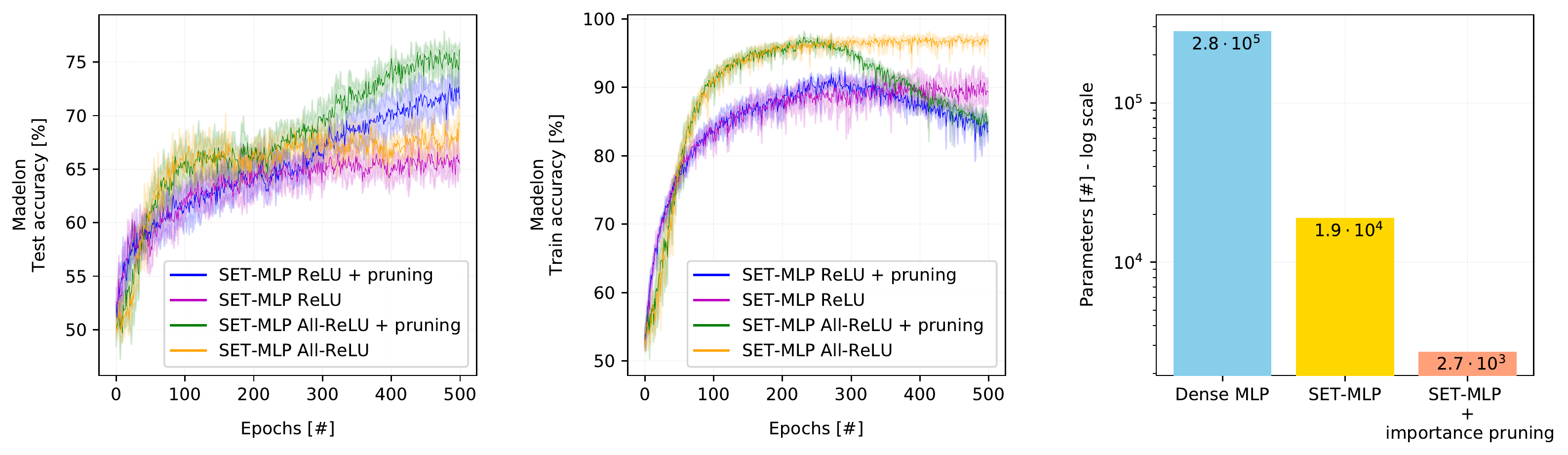}
		\caption{Learning curves and parameter numbers comparison for Madalon dataset.}\label{fig:1c_evaluation}
	\end{subfigure}
	\vspace{2mm}
	\begin{subfigure}[t]{0.9\textwidth}
		\centering
		\includegraphics[width=0.9\textwidth]{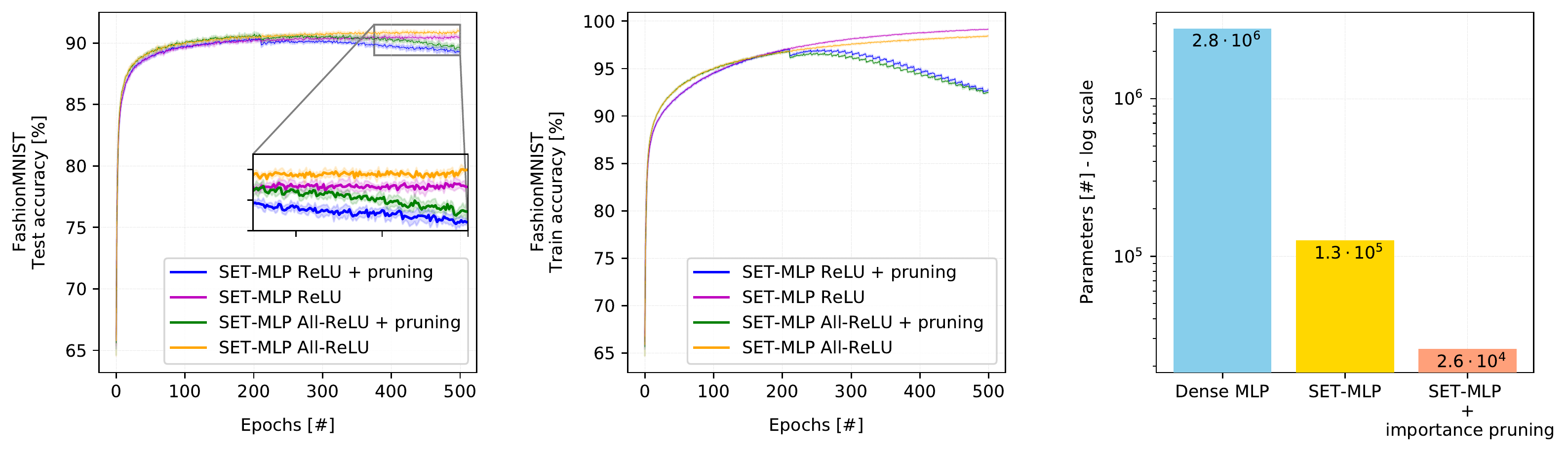}
		\caption{Learning curves and parameter numbers comparison for FashionMNIST dataset.}\label{fig:1d_evaluation}
	\end{subfigure}
	\vspace{2mm}
	\begin{subfigure}[t]{0.9\textwidth}
		\centering
		\includegraphics[width=0.9\textwidth]{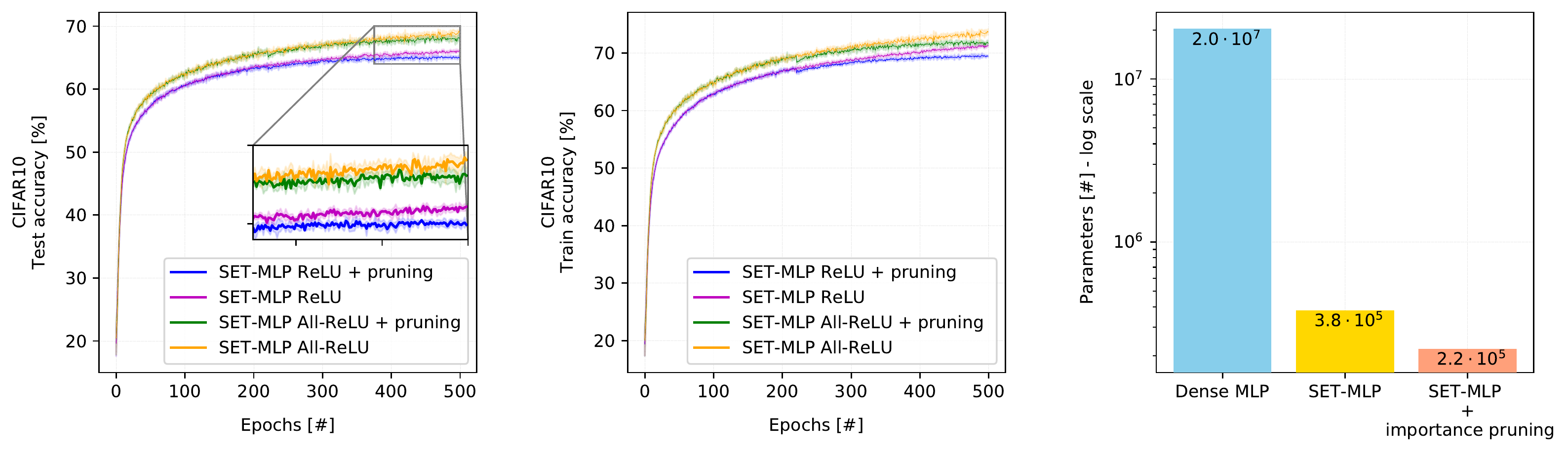}
		\caption{Learning curves and parameter numbers comparison for CIFAR10 dataset.}\label{fig:1e_evaluation}
	\end{subfigure}
	\captionsetup{justification=centering}
	\caption{Evaluation of the proposed methods on five different datasets. These results are obtained by standard sequential training with \textit{momentum} SGD.}\label{figure:evaluation}
\end{figure}

\begin{figure}	
	\centering
	\begin{subfigure}[t]{0.9\textwidth}
		\centering
		\includegraphics[width=0.9\textwidth]{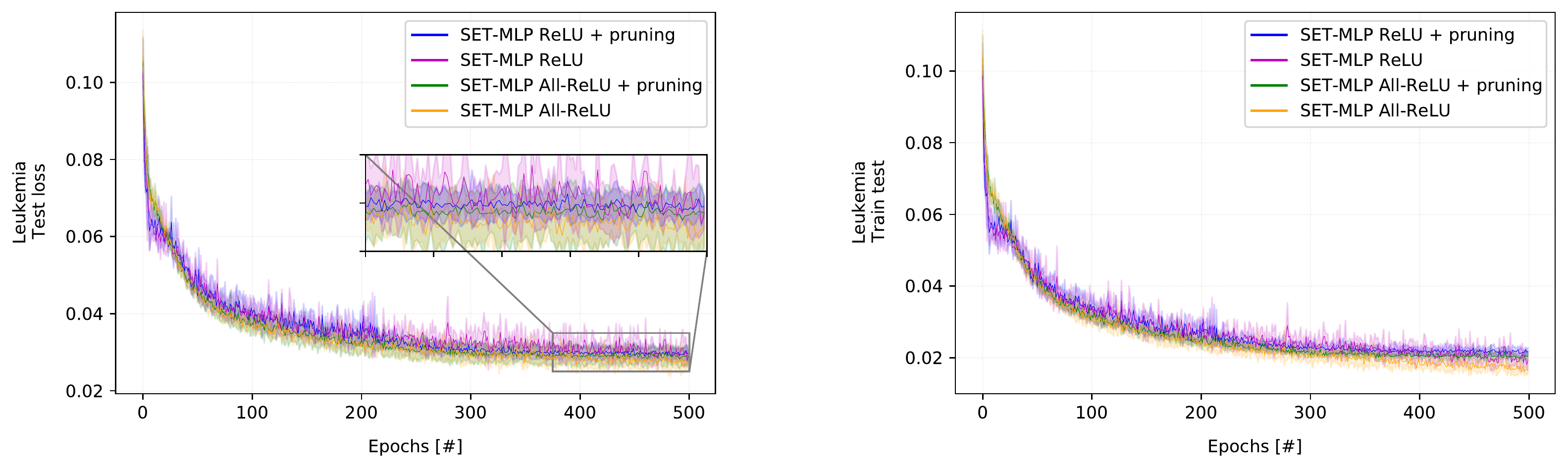}
		\caption{Train and test error for Leukemia dataset.}\label{fig:1a}		
	\end{subfigure}
	\vspace{2mm}
	\begin{subfigure}[t]{0.9\textwidth}
		\centering
		\includegraphics[width=0.9\textwidth]{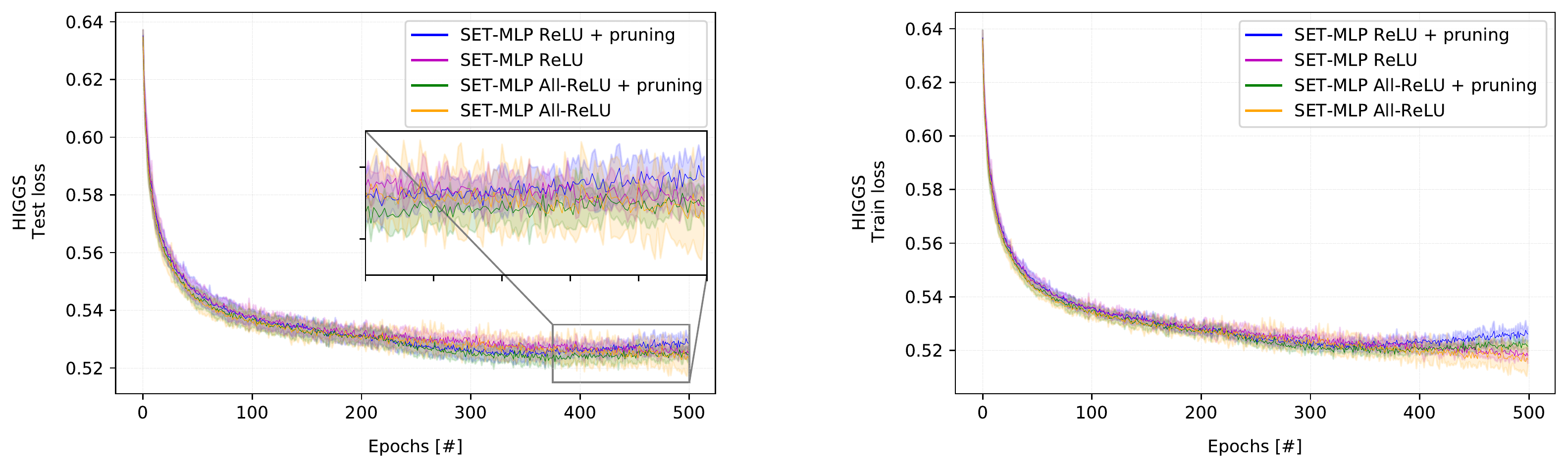}
		\caption{Train and test error for HIGGS dataset.}\label{fig:1b}
	\end{subfigure}
		\vspace{2mm}
	\begin{subfigure}[t]{0.9\textwidth}
		\centering
		\includegraphics[width=0.9\textwidth]{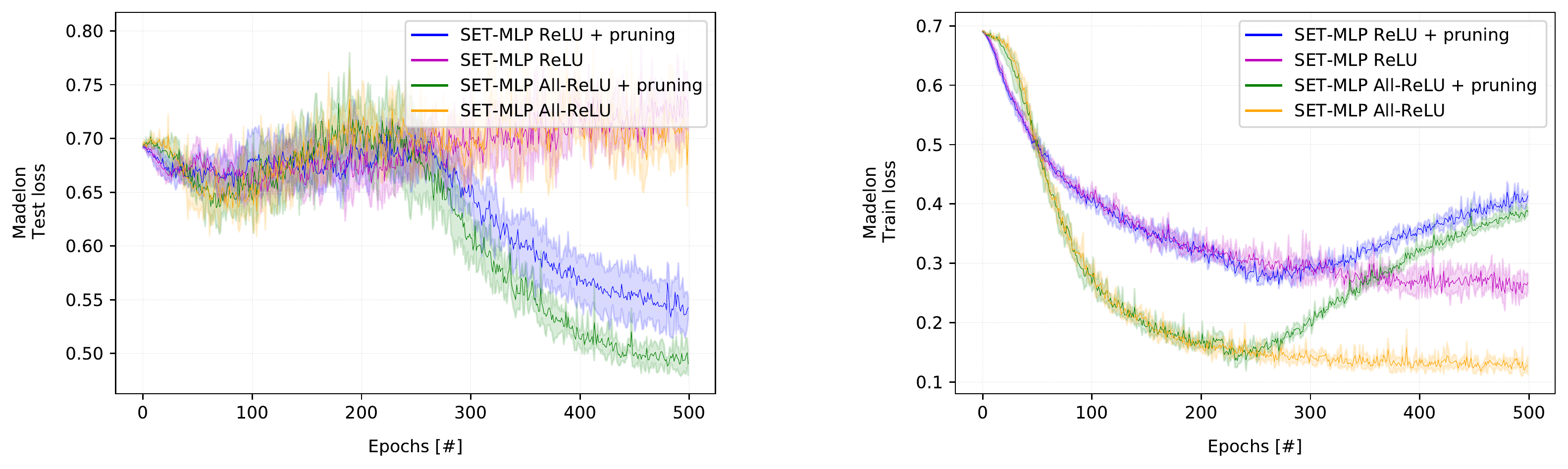}
		\caption{Learning curves and parameter numbers comparison for Madalon dataset.}\label{fig:1c}
	\end{subfigure}
	\vspace{2mm}
	\begin{subfigure}[t]{0.9\textwidth}
		\centering
		\includegraphics[width=0.9\textwidth]{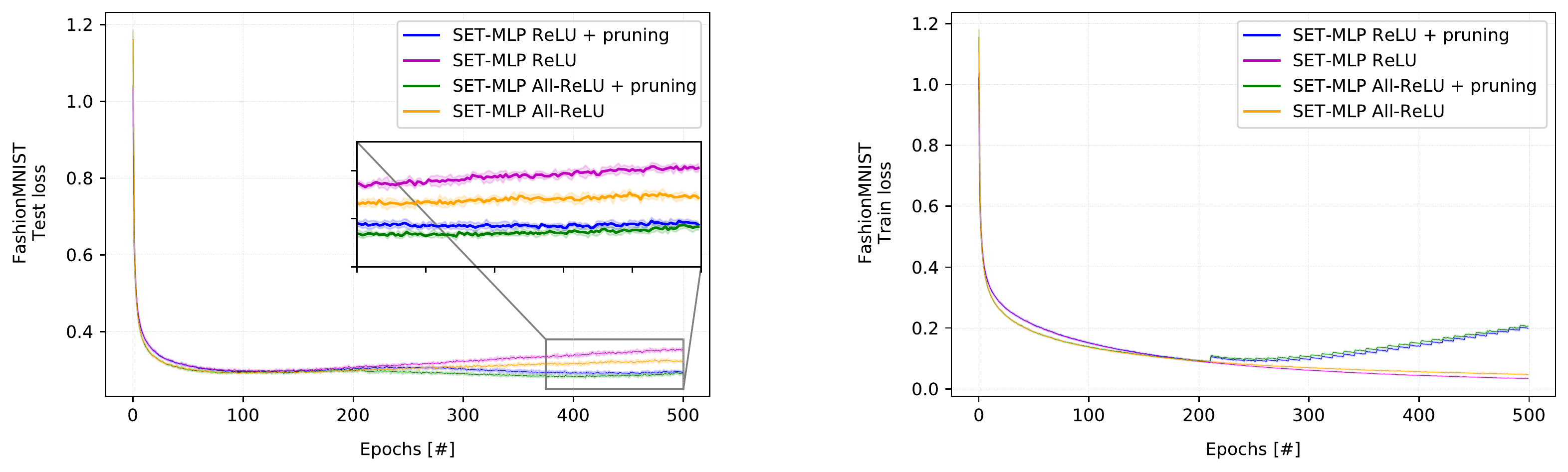}
		\caption{Train and test error for FashionMNIST dataset.}\label{fig:1d}
	\end{subfigure}
	\vspace{2mm}
	\begin{subfigure}[t]{0.9\textwidth}
		\centering
		\includegraphics[width=0.9\textwidth]{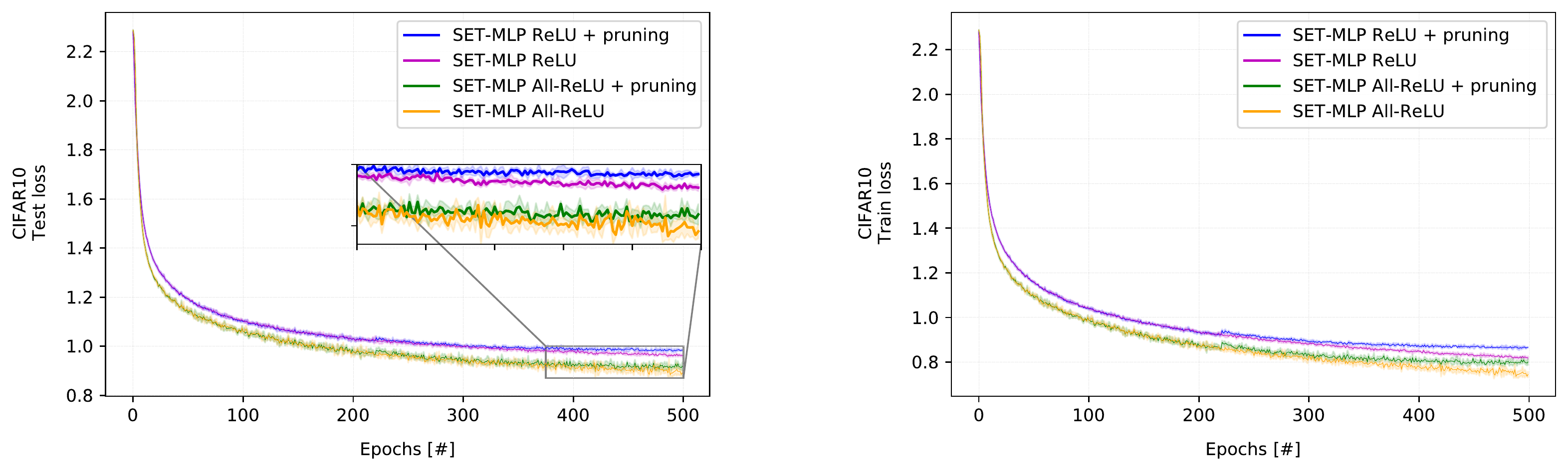}
		\caption{Train and test error for CIFAR10 dataset.}\label{fig:1e}
	\end{subfigure}
	\captionsetup{justification=centering}
	\caption{Train and test errors of the experiments results presented in \autoref{figure:evaluation} (sequential training with \textit{momentum} SGD).}\label{figure:losses}
\end{figure}

This section summarises the performance for the comparison between All-ReLU and ReLU on sparse MLP models, and their integration with \textit{Importance Pruning} to speed up the training. All SET-MLP variants have been run using our own truly sparse implementation framework and just one CPU core. \autoref{table:results} lists the maximum accuracy for each method/dataset combination, along with the total training time (expressed in minutes), the number of parameters at the beginning ($start\_n^W$) and at the end of the training ($end\_n^W$). This is particularly interesting to report when \textit{Importance Pruning} is applied to the sparse models for understanding the benefits in terms of memory footprint. The resulting learning curves and errors of our experiments are shown in \autoref{figure:evaluation} and \autoref{figure:losses}, for both testing and training sets. For all figures, we obtain the mean accuracy by averaging the best test accuracy from 5 trials over 500 epochs. Moreover, for each model, we display the resulting number of parameters for the dense MLP version, the basic SET-MLP and SET-MLP where the neuron importance metric is adopted for taking out unimportant hidden units. Since a key result is the relative space improvement from the \textit{Importance Pruning}, in \autoref{figure:memor_vs_error}, we plot the relative model size (expressed in number of parameters) against the relative error to highlight the differences with and without \textit{Importance Pruning} (where \autoref{figure:memor_vs_error_train} summarizes the final training results and \autoref{figure:memor_vs_error_test} for the final testing results).

\begin{table}[!htbp]
\resizebox{1\textwidth}{!}{%
\begin{tabular}{llllllllll}

\toprule
\textbf{Dataset} & \textbf{Architecture} & \textbf{NN model} & \textbf{CPU} & \textbf{Activation} & \textbf{Importance} & & \textbf{Results} & \\ \cmidrule{7-10}
& & &\textbf{cores} & &\textbf{Pruning}& \textbf{Accuracy [\%]} & \textbf{$\mathbf{start\_n^W}$ [\#]} & \textbf{$\mathbf{end\_n^W}$ [\#]} & \textbf{Training [min]} \\ \midrule
\textbf{Leukemia} & 54675-27500-27500-18 & \textbf{SET-MLP} & 1 & ReLU & no & 85.98 & 1684944 & 1684944 & $\sim$ 375\\
& & & & ReLU & yes & 85.40 & 1582214 & 312900 & $\sim$ 266.5\\
& & & & All-ReLU ($\alpha=0.75$) & no & \textbf{86.42} & 1582518 & 1582518 & $\sim$ 375\\
& & & & All-ReLU ($\alpha=0.75$) & yes & 85.69 & 1582581 & 313047 & $\sim$ 266.5\\
& & \textbf{Dense MLP \tablefootnote{With our hardware settings (CPU Intel Core i7-9750H, 2.60 GHz $\times$ 6, RAM 32 GB, Hard disk 1000 GB, NVIDIA GeForce GTX 1650 4GB), it was not possible to train the dense MLP on Leukemia due to memory error.}} & 12 & ReLU & - & n/a & 2260362518 & 2260362518 & n/a\\
& & & & All-ReLU & - & n/a & 2260362518 & 2260362518 & n/a\\
\textbf{Higgs} & 28-1000-1000-1000-2 & \textbf{SET-MLP} & 1 & ReLU & no & 73.59 & 50224 & 50244 & $\sim$ 216.67\\
& & & & ReLU & yes & 73.50 & 50246 & 10065 & $\sim$ 182.3\\
& & & & All-ReLU ($\alpha=0.05$)& no & 73.67 & 50165 & 50165 & $\sim$ 216.67\\
& & & &All-ReLU ($\alpha=0.05$)& yes & \textbf{73.76} & 50165 & 9992 & $\sim$ 140.3 \\
& & \textbf{Dense MLP} & 12 & ReLU & - & 70.59 & 2033002 & 2033002 & $\sim$ 182.3\\
& & & & All-ReLU & - & 70.10 & 2033002 & 2033002 & $\sim$ 140.13\\
\textbf{Madelon} & 500-400-100-400-2 & \textbf{SET-MLP }& 1 & ReLU & no & 68.50 & 19000 & 19000 & $\sim$ 3.6\\
& & & & ReLU & yes & 75.00 & 19000 & 2739 & $\sim$ 3.2\\
& & & & All-ReLU ($\alpha=0.5$) & no & 71.33 & 19011 & 19011 & $\sim$ 3.6\\
& & & & All-ReLU ($\alpha=0.5$)& yes & \textbf{77.00} & 19000 & 2737 & $\sim$ 3.2\\
& & \textbf{Dense MLP} & 12 & ReLU & - & 59.66 & 281702 & 281702 & $\sim$ 3.62\\
& & & & All-ReLU ($\alpha=0.5$) & - & 62.00 & 281702 & 281702 & $\sim$ 3.62\\
\textbf{FashionMNIST} & 784-1000-1000-1000-10 & \textbf{SET-MLP} & 1 & ReLU & no & 90.48 & 126302 & 126302 & $\sim$ 137.5\\
& & & & ReLU & yes & 89.43 & 126302 & 26111 & $\sim$ 96.25\\
& & & & All-ReLU ($\alpha=0.6$) & no & \textbf{91.38} & 126302 & 126302 & $\sim$ 137.5\\
& & & & All-ReLU ($\alpha=0.6$)& yes & 90.12 & 126302 & 25759 & $\sim$ 96.25\\
& & \textbf{Dense MLP} & 12 & ReLU & - & 90.85 & 2797010 & 2797010 & $\sim$ 95.05\\
& & & & All-ReLU ($\alpha=0.25$) & - & 90.73 & 2797010 & 2797010 & $\sim$ 95.05\\
\textbf{CIFAR10} & 3072-4000-1000-4000-10 & \textbf{SET-MLP} & 1 & ReLU & no & 67.05 & 381758 & 381758 & $\sim$ 590\\
& & & & ReLU & yes & 65.21 & 381758 & 238114 &$\sim$ 530\\
& & & & All-ReLU ($\alpha=0.75$)& no & \textbf{69.83} & 381425 & 381425 & $\sim$ 590\\
& & & & All-ReLU ($\alpha=0.75$) & yes & 68.55 & 380318 & 221323 & $\sim$ 530\\
& & \textbf{Dense MLP} & 12 & ReLU & - & 64.94 & 20337010 & 20337010 & $\sim$ 530.7\\
& & & & All-ReLU ($\alpha=0.25$) & - & 67.96 & 20337010 & 20337010 & $\sim$ 530.7\\
\bottomrule
\end{tabular}}
\vskip 0.25cm
\caption{On each dataset, we report the best classification accuracy and error obtained by each model on the test data over five different runs for 500 epochs. $start\_n^W$ represents the number of weights in the model at the beginning of the training, while $end\_n^W$ represents the number of parameters in the final model. \textit{Importance Pruning} (y/n) indicates if the proposed pruning strategy based on our neuron importance metric is activated. \textit{Training} reports the overall running time needed for training the models. Furthermore, the table reports the performance of fully connected MLPs (Dense MLPs) with both activation functions. The networks have been trained using momentum SGD in its standard sequential version. It is worth mentioning that the Dense MLP has been run using Keras using all CPU cores and SET-MLP using our own implementation and just one CPU core. }
\label{table:results}
\end{table}

We can observe that All-ReLU consistently outperforms ReLU on all datasets, indicating that the novel activation function associated with a sparse-to-sparse training algorithm helps to model better the data distribution. Moreover, when \textit{Importance Pruning} is activated, we can notice a significant reduction in the number of parameters, which leads to a remarkable speedup in running time, with almost no loss in terms of accuracy. Looking at the CIFAR10 dataset (\autoref{fig:1e_evaluation}), we can see that the new activation function is capable of boosting the accuracy on test data by more than 2\%. SET-MLP on CIFAR10, after 500 epochs when using SReLU reaches about 70.30 \% accuracy. This result suggests that All-ReLU indeed fills the performance gap with SReLU successfully. The model version with \textit{Importance Pruning} can achieve comparable performances while training roughly 40\% fewer parameters (where the reduced number of parameters refers to the final model and it is obtained by  gradually reducing the connections during training) and gaining a speedup of 60 minutes. In this case, the importance metric seems to be more stable when adopting All-ReLU, resulting in a minor loss in performance. A similar outstanding result is obtained with Madelon (\autoref{fig:1c_evaluation}), where All-ReLU obtains 3\% increase in accuracy with no \textit{Importance Pruning} and about 2\% with \textit{Importance Pruning} (where the model uses 80\% fewer parameters). Here, the \textit{Importance Pruning} method has improved the performance significantly. 

\vskip 0.5cm
\begin{figure}[!htbp]
\centering
   \subfloat[\label{figure:g1a} CIFAR10]{%
   \centering
      \includegraphics[ width=0.3\textwidth]{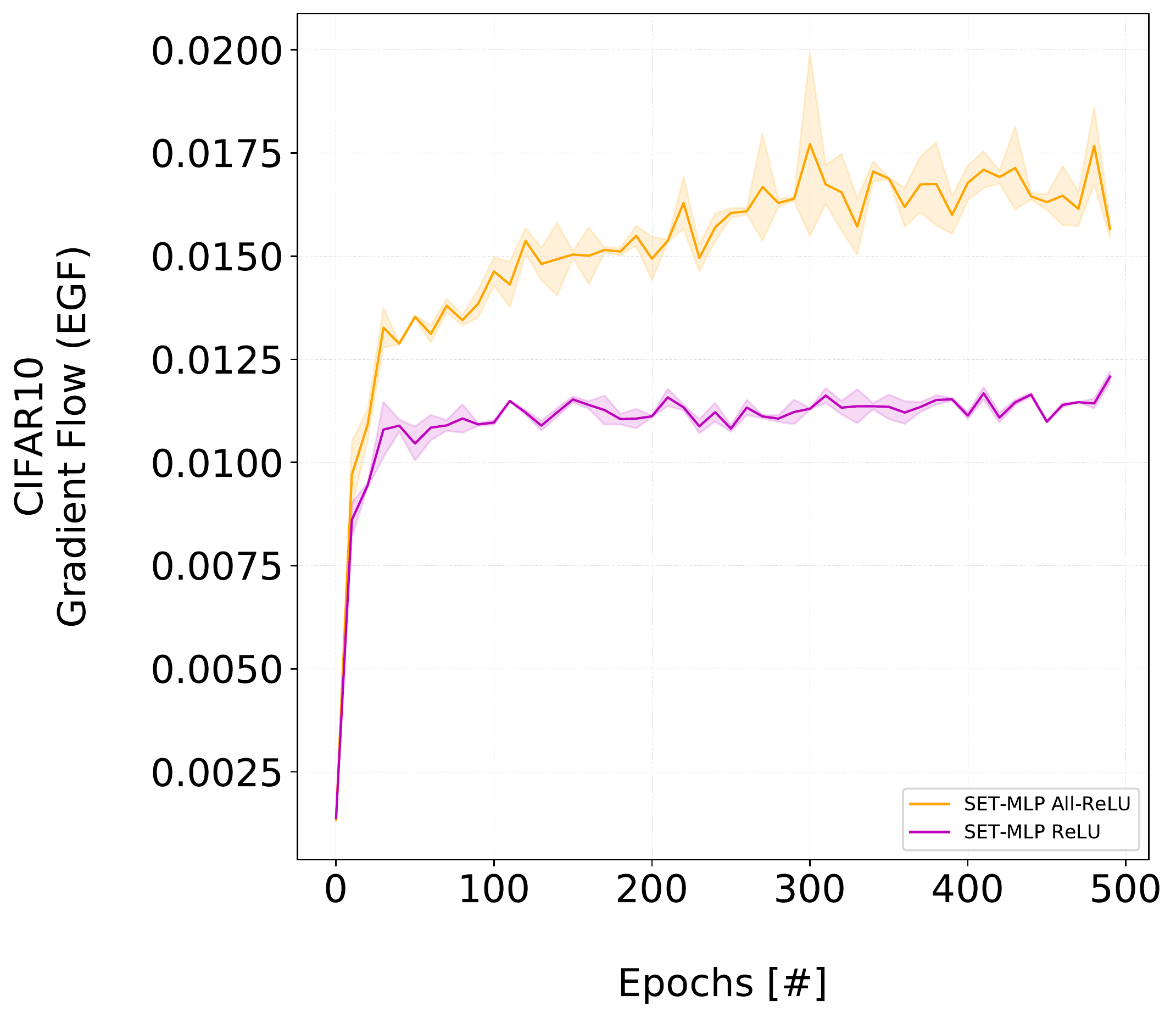}}
\hspace{\fill}
   \subfloat[\label{figure:g1b} FashionMNIST ]{%
   \centering
      \includegraphics[ width=0.3\textwidth]{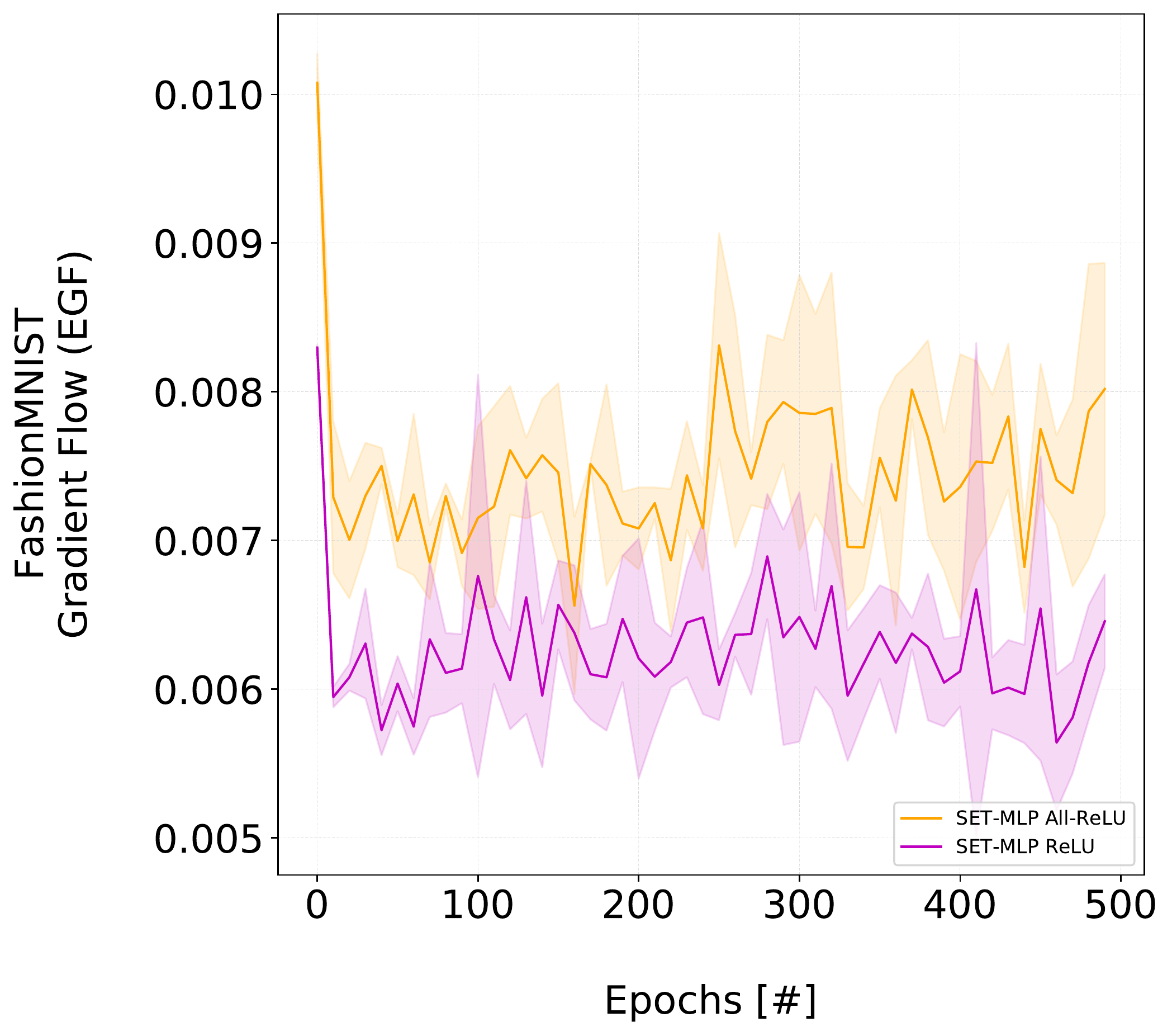}}
\hspace{\fill}
   \subfloat[\label{figure:g1c} Madelon]{%
   \centering
      \includegraphics[width=0.3\textwidth]{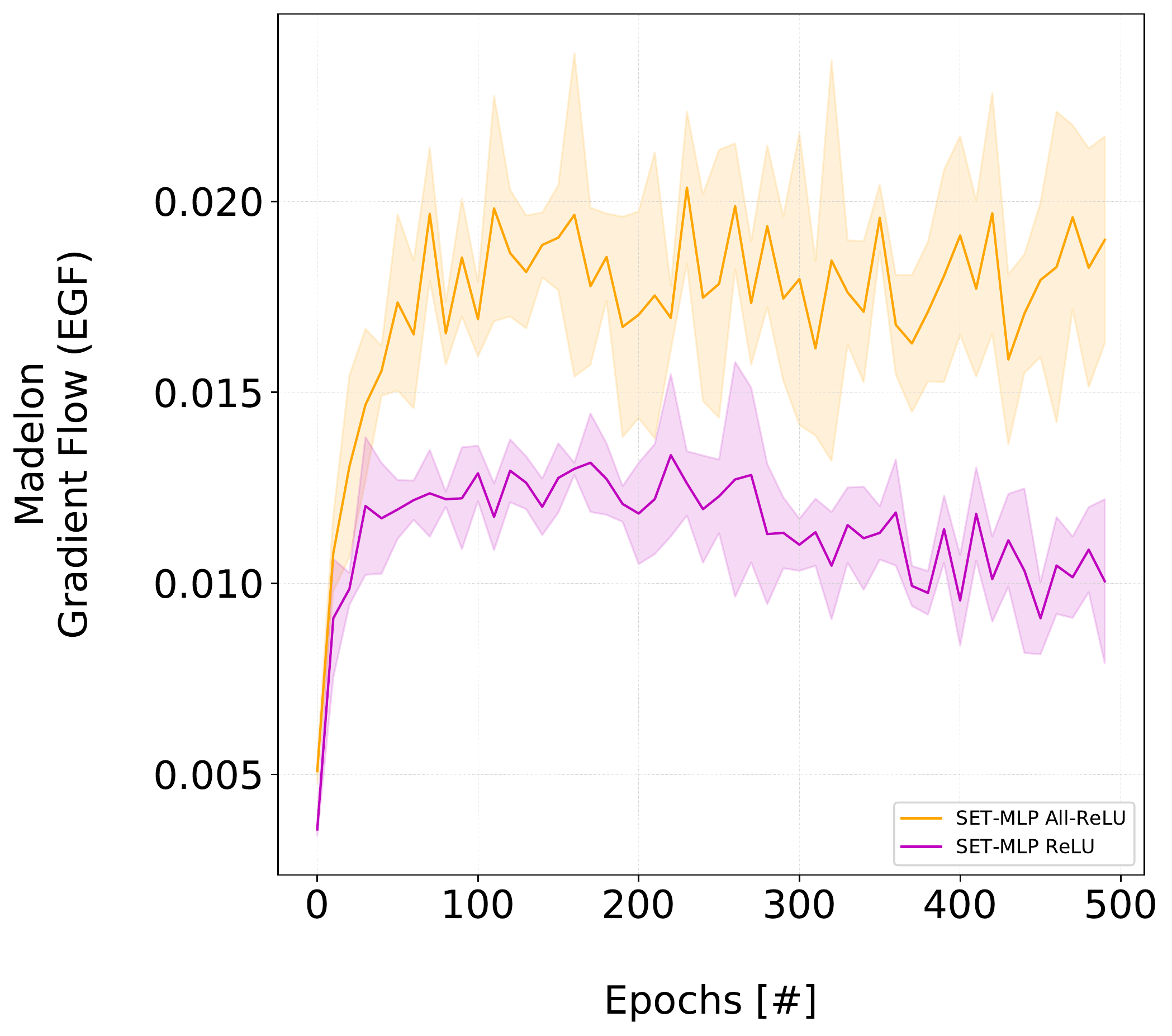}}
\vskip 0.25cm
\captionsetup{justification=centering}
\caption{Gradient Flow for sparse MLPs with three hidden layers on CIFAR10 (a), FashionMNIST (b) and Madelon (c) trained with All-ReLU and ReLU.}
\label{figure:gflow}
\end{figure}

For FashionMNIST, All-ReLU can surpass both ReLU (\autoref{fig:1d_evaluation}) and SReLU. The latter achieves around 90.10 \% accuracy when trained for 500 epochs with the same settings, while All-ReLU achieves 91.38 \%. The smaller version obtained via Importance Pruning ends with 20\% of the original parameters, gaining a 41 minutes speedup. We attain similar performance for Leukemia and Higgs, although the increase in accuracy is more predominant for image data and Madelon. Lastly, in \autoref{figure:gflow}, we show the gradient flow for the sparse models trained with All-ReLU and ReLU on CIFAR10 (\autoref{figure:g1a}), FashionMNIST (\autoref{figure:g1b}) and Madelon (\autoref{figure:g1c}). We recall that gradient flow is the first-order approximation of the decrease in the loss expected after a gradient step, hence the higher, the better. All-ReLU visibly improves this metric, which is associated with efficient training of sparse neural networks (\cite{Tessera2021KeepTG, wang2020picking, Evci2020GradientFI}).

\subsection{Performance on Parallel Trained Sparse MLPs}
\label{subsection:parallelmlps}
The available algorithms for parallelisation of dense neural networks are not suitable for sparse neural networks training. Hence they are not considered in our experimental evaluation of the proposed method. We did not test WASAP-SGD with Madelon and Leukemia because there is no added value in parallelising the training under the conditions where the data and/or the model size are minimal. In \autoref{table:running_time} we summarise the performance of the proposed parallel algorithm on three different datasets where All-ReLU is adopted, with and without \textit{Importance Pruning}. For completeness, we report the results for a variant where \textit{phase one} is synchronous, called WASSP-SGD (\textit{Weight Averaging Sparse Synchronous Parallel SGD}). In this way, we want to empirically demonstrate that our asynchronous version is more suitable to train sparse models. Furthermore, for easy comparison, we again report the accuracy and training time for the sequential version (baseline). Since we run the experiments on a machine with six physical cores, we employed five workers and one master (parameter server). The time reduction does not improve significantly as the number of workers surpasses the number of physical processors.

The synchronous variant WASSP-SGD is implemented by following the suggestions from (\cite{goyal2017accurate}), such as their gradual warmup and linear scaling rule for the learning rate. Conversely, for our WASAP-SGD, where the first phase is asynchronous, we observed it benefits from larger learning rates for the first few epochs, followed by fixed learning rates.
\begin{table}[!htbp]
\resizebox{\textwidth}{!}{%
\begin{tabular}{cccccccccc}
\toprule
\textbf{Dataset} & \textbf{Framework} & \textbf{\descrcell{Importance}{Pruning} }& \textbf{\descrcell{CPU}{cores} }&
\textbf{GPU} &
\textbf{Accuracy [\%]} & \textbf{\descrcell{Training}{time [min]}} & \textbf{\descrcell{CPU}{usage [\%]}} &
\textbf{\descrcell{Memory}{usage [MB]}}  \\
\midrule

\textbf{Higgs} & WASSP-SGD & no & 12 & no & 74.26 & $\sim$ 142.52 & $\sim$ 62\% & $\sim$ 2800 MB & \\
& WASSP-SGD & yes & 12 & no & 74.16 & $\sim$ 113.33 & $\sim$ 62\% & $\sim$ 2800 MB &\\
& WASAP-SGD & no & 12 & no & 74.47 & $\sim$ 138.52 & $\sim$ 62\% & $\sim$ 2800 MB &\\
& WASAP-SGD & yes & 12 & no & 74.88 & $\sim$ 108.33 & $\sim$ 62\% & $\sim$ 2800 MB &\\

& Sequential & no & 1 & no & 73.67 & $\sim$ 216.67 & $\sim$ 24\% & $\sim$ 1700 MB &\\
& Sequential & yes & 1 & no & 73.76 & $\sim$ 182.3 & $\sim$ 24\% & $\sim$ 1700 MB &\\

& \textit{Keras CPU} & no & 1 & no & 74.28 & 
$\sim$ 350 & $\sim$ 27 \% & $\sim$ 2100 MB &\\
& \textit{Keras CPU} & no & 12 & no & 74.28 & 
$\sim$ 191.7 & $\sim$ 65 \% & $\sim$ 2300 MB &\\
& \textit{Keras GPU} & no & 12 & yes &  74.28 & 
$\sim$ 78.3 & $\sim$ 25 \% & $\sim$ 2400 MB &\\

\textbf{FashionMNIST} & WASSP-SGD & no & 12 & no & 90.52 & $\sim$ 90.6 & $\sim$ 65\% & $\sim$ 3300 MB &\\
& WASSP-SGD & yes & 12 & no & 89.96 & $\sim$ 77.6 & $\sim$ 65\% & $\sim$ 3300 MB &\\

& WASAP-SGD & no & 12 & no & 91.23 & $\sim$ 87.25 & $\sim$ 65\% & $\sim$ 3300 MB &\\
& WASAP-SGD & yes & 12 & no & 90.15 & $\sim$ 74.85 & $\sim$ 65\% & $\sim$ 3300 MB &\\

& Sequential & no & 1 & no & 91.38 & $\sim$ 137.5 & $\sim$ 20 \% & $\sim$ 1900 MB &\\
& Sequential & yes & 1 & no & 90.12 & $\sim$ 96.25 & $\sim$ 20 \% & $\sim$ 1900 MB &\\

& \textit{Keras CPU} & no & 1 & no & - & 
$\sim$ 316 & $\sim$ 20 \% & $\sim$ 2800 MB &\\
& \textit{Keras CPU} & no & 12 & no & - & 
$\sim$ 135.6 & $\sim$ 76 \% & $\sim$ 2800 MB &\\
& \textit{Keras GPU} & no & 12  & yes & - & 
$\sim$ 65 & $\sim$ 26 \% & $\sim$ 3000 MB &\\

\textbf{CIFAR10} & WASSP-SGD & no & 12 & no &  67.13 & $\sim$ 309.9 & $\sim$ 70 \% & $\sim$ 6500 MB &\\
& WASSP-SGD & yes & 12 & no & 66.88 & 
$\sim$ 279.9 & $\sim$ 70 \% & $\sim$ 6500 MB &\\
& WASAP-SGD & no & 12 & no & 69.03 & $\sim$ 281.9 & $\sim$ 70 \% & $\sim$ 6500 MB &\\
& WASAP-SGD & yes & 12 & no & 68.51 & $\sim$ 246.5 & $\sim$ 70 \% & $\sim$ 6500 MB &\\

& Sequential & no & 1 & no & 69.83 & $\sim$ 590 & $\sim$ 25 \% & $\sim$ 2200 MB &\\
& Sequential & yes & 1 & no & 68.55 & 
$\sim$ 530 & $\sim$ 25 \% & $\sim$ 2200 MB &\\

& \textit{Keras CPU} & no & 1 & no & - & 
$\sim$ 1000 & $\sim$ 20 \% & $\sim$ 4700 MB &\\
& \textit{Keras CPU} & no & 12 & no & - & 
$\sim$ 530.7 & $\sim$ 80 \% & $\sim$ 4700 MB &\\
& \textit{Keras GPU} & no & 12 & yes & - & 
$\sim$ 195 & $\sim$ 20 \% & $\sim$ 4800 MB &\\

\bottomrule
\end{tabular}}
\vskip 0.25cm
\caption{The table reports the accuracy, average running time and average resource utilisation over five different runs for 500 epochs when using parallel training with WASAP-SGD and our proposed sparse implementation framework. For completeness, we report the performance for the synchronous (phase 1) version of the algorithm as well. Here, this synchronous version is called WASSP-SGD (\textit{Weight Averaging Sparse Synchronous Parallel SGD}). Moreover, we include the performance of sequential training for facilitating the comparison. \textit{Importance Pruning} (y/n) indicates if the proposed pruning strategy based on our neuron importance metric is activated. Note that in this setting, the memory usage will decrease during training. To have an idea on how these performances can be compared against a state-of-the-art framework, we report the running time and memory statistics when using the Keras implementation of SET-MLP with a mask over the parameters. These statistics are provided for three different configurations of the hardware.}
\label{table:running_time}
\end{table}






In \autoref{table:running_time}, we also show the average running time using the Keras implementation of SET-MLP with a mask over the parameters. The statistics are provided for three different configurations of the hardware: training on one CPU core only, training with no constrains (all 12 CPUs cores) and GPU training. These numbers allow us to make a comparison between our proposed sparse framework and a popular deep learning library like Keras. By looking at the results in \autoref{table:running_time}, we can observe that the proposed parallel algorithm exhibits persistently better convergence when the first phase is carried out asynchronously (WASAP-SGD). The same outcome holds in terms of training time. If we compare the running times of \textit{WASAP-SGD} against the one from the sequential version, we gain an improvement of about half among all datasets, without introducing a notable increase in memory footprint. Our parallel method is able to outperform Keras CPU-based wall-clock running time for training of sparse MLPs significantly. We would like to emphasise that our sparse implementation together with parallelisation and \textit{Importance Pruning} gets quite close to GPU training time (see \autoref{table:running_time}). Plus, this is the kind of comparison that people typically do not make, because (for dense networks) does not make sense to compare CPU with GPU. Moreover, as we could notice from the results in \autoref{subsection:mlps} for the Leukemia dataset, this computational advantage of GPUs manifests its limitation when it comes to training huge models. In this case, Keras is not able to allocate the dense tensors, resulting in a memory error. Additionally, it is worth mentioning that GPU training utilises more resource than CPU training and classic GPU training, as it also uses the CPU besides GPU. Hence, we believe that our approach has more potential to tackle large scale deep learning models.

\subsection{Extreme large sparse MLPs}
\label{subsection:extreme}
To the best of our knowledge, the largest public (claimed) dense neural network has 160 billion (B) parameters, where a parameter roughly corresponds to a synapse in the human brain. Given the human brain is estimated to have about 100 trillion synapses, that neural network could be said to be about 0.16\% of the human brain. State-of-the-art networks in a generic sense contain around 16M neurons. For comparison, the 16 million (M) neurons number when compared with the 100B neurons in a human brain—only represents 0.016\% of human brain size. Now, it is worth noting that size is not the only thing that matters. Most advanced models today perform worse when their network size is increased blindly. While it is true that DNN capabilities increase with their network size, there is also a fair amount of engineering work that goes into making larger networks accurate. Hopefully, sparsity will help in overcoming some challenges when training such large models.

In this scenario, we tried to push the limit of ANNs on a virtual machine with 96 cores and 768GB of RAM to train extreme large sparse MLP models. We are attempting to enter a region where neural networks have never been explored. Hence, we believe that any small finding is important. Because of limited time and resources, we did not repeat experiments to get statistical confidence. We just run them once for few epochs, on different regimes, to collect statistics such as matrix initialisation time, training time per epoch, inference time and topology evolution time.

Our main goal was to train a sparse model with more than 125 million neurons because we believe that this is the latest state-of-the-art (just for inference) size, according to (\cite{sparseinference}). For the sake of clarity, we mention that these results ignore entirely the training focusing just on inference. We want to go one step further and train such models. Herein, we have discovered that we are very close of reaching the limits of our proposed implementation framework of training extremely large sparse neural networks, but it is good to know the limits in order to know how to proceed further.

There are not many available datasets with a high number of features and a reasonable amount of data points to exploit data parallelism. For this reason, we created an artificial dataset by adopting the function \texttt{make\_classification} \footnote{More details about the function are available \href{https://scikit-learn.org/stable/modules/generated/sklearn.datasets.make_classification.html}{here}.} from Scikit-learn, a free software machine learning library for the Python programming language. The algorithm is adapted from (\cite{Guyon2003DesignOE}) and was designed to generate the "Madelon" dataset. We generated a binary classification task with 10000 samples, where each sample has 65536 features. We used 30\% of the data as test data and 70\% as train data. The models are trained using our parallel algorithm WASAP-SGD with momentum (set to 0.9), weight decay and dropout (set to 0.4).
Moreover, the batch size is set to 128, and the learning rate is 0.01. The statistics for different architectures are presented in table \autoref{table:extremeresults}. Moreover, we report the number of workers, the number of parameters and the value for $\epsilon$, which controls the sparsity level. We stress that the training is performed on a single machine with no GPU, while popular state-of-the-art models are usually trained with distributed algorithms on multi GPUs.

\begin{table}[!htbp]
\resizebox{\textwidth}{!}{%
        \begin{tabular}{lllllllll}
            \toprule
            \textbf{Architecture}  & 
            \textbf{Epsilon $\mathbf{\epsilon}$} & \textbf{Neurons [\#]} & 
            \textbf{Parameters [\#]} & 
            \textbf{Workers [\#]} &  
            
            \textbf{\descrcell{Weight}{initialization [min]}} & 
            \textbf{\descrcell{Training}{[min]}}
            & \textbf{\descrcell{Testing}{[min]}} &  
             \textbf{\descrcell{Weight}{evolution [min]}}     \\ \midrule
            
              \textbf{65536-0.5M-0.5M-2} & 10 &  1 M  & 20.6 M & 16& $\sim$ 1 &   $\sim$ 6  & $\sim$ 2.5 & $\sim$ 1 \\
              
              \textbf{65536-2.5M-2.5M-2} & 5 & 5 M & 50.3 M & 16 & $\sim$ 2  &  $\sim$ 10  &  $\sim$ 6 & $\sim$ 2\\
           
              \textbf{65536-5M-5M-2} & 5 & 10 M & 100.3 M & 16 & $\sim$ 3  & $\sim$ 20 &  $\sim$ 11 & $\sim$ 5\\
            
              \textbf{65536-5M$\times$ 4-2} & 1 & 20 M & 40.3 M & 8 & $\sim$ 5  & $\sim$ 26 &  $\sim$ 18 & $\sim$ 6\\
      
              \textbf{65536-5M $\times$ 10-2} & 1 & 50 M & 100.3M & 8 & $\sim$ 9 & $\sim$ 52 &  $\sim$ 20 & $\sim$ 6\\
                      
             \bottomrule
        \end{tabular}}
        \caption{The table reports the running time (per epoch) when training extreme large sparse models with WASAP-SGD on the big artificial dataset. The training is performed with 16 workers. We trained the model for a few epochs, to make sure the loss is decreasing, hence the networks are learning.}
        \label{table:extremeresults}
\end{table}

While trying to build these large sparse models, multiple challenges and technical difficulties have emerged:
\begin{description}
    \item[Inference bottleneck:] Up to this moment, our work was focused on the training phase of neural networks since inference did not play an important role for common sized networks. However, once we started building extreme large models, we immediately noticed how important it is to optimise this phase as well. We tried to overcome the bottleneck by parallelising the inference in batches with python \texttt{mutiprocessing}. We are aware that there exist more sophisticated approaches, but this simple solution could already decrease the running time considerably.
    
    \item[MPI overflow:] Parallelization in Python integrates Message Passing Interface via \texttt{mpi4py} module. When we first selected this framework, we did not consider that mpi4py does not support the parallelisation of objects greater than $2^{31}$ bytes, limiting the size of sparse matrices to be created. This limitation is probably becoming more noticeable nowadays, given the importance of Big Data analysis, and in \cite{BIGMPI}, the authors developed BigMPI4py, a Python module that wraps mpi4py, supporting object sizes beyond this boundary. 
    
    \item[Matrix initialisation time:] When building such larges models, weights initialisation starts to play a significant role. If the sparse matrix initialisation is not implemented efficiently, this may cause a bottleneck. In this regard, we had to vectorise this step in order to reduce the weight initialization running time.
    
    \item[Memory allocation issues:] Sparsity allows for creating a bigger model (in terms of the number of hidden units) than when one adopts fully connected layers. However, this advantage has its limitations as well. When adopting a data parallelism strategy, the sparse model is replicated among all workers to accelerate the training procedure. When the model becomes extremely large, the number of workers which can fit in memory decreases (as can be noticed in \autoref{table:extremeresults}{}). At this point, the training should become distributed by having each workers running on a different node in a network. Alternatively, other strategies may be explored to overcome the memory issues. Our implementation adopts MPI standard; hence it could be automatically run in a distributed fashion.  Nevertheless, the feasible size of sparse models is much larger than their dense counterparts. In this regard, the largest dense network we can build on the virtual machine has around 600000 neurons. 
\end{description}

We note that the weight evolution step is able to scale successfully without adding too much overhead. Moreover, we want to briefly outline that we managed to train a model with 10 million neurons ($\epsilon=1$) via sequential training on Leukemia. Each training epoch takes around 33 minutes, and 18 seconds, inference takes 12 minutes and 9 seconds, and evolution time 30 seconds. Similarly, we could train a model with 50 million neurons where one epoch takes about 1 hour and 15 minutes. In the end, we were not able to train sparse models with more than 50 million neurons due to the challenges listed above. After a point, to grow the number of neurons, we had to increase the sparsity as well as decrease the number of workers. Nonetheless, from these extreme experiments, we could learn some of the limitations of our approach and the chosen technologies. Lastly, it is important to highlight that these limitations come from the implementation itself and they are not limitations of the three theoretical contributions.

\section{Discussion}
\label{sec:discussion}

To improve the scalability of ANNs by exploiting sparsity, three main contributions have been introduced in this work. Each of them has been implemented in a truly sparse manner with sparse matrices and operations. First, we introduced \textit{WASAP-SGD}, a new parallel algorithm to efficiently train sparse neural networks asynchronously. This type of communication protocol has proven to be more effective than synchronous training for sparse models, both from a running time and convergence point of view. Furthermore, the last training phase ensures higher generalization performance. For example, for CIFAR10, we could bring down the sequential training time of \textit{590} minutes to \textit{282} without \textit{Importance Pruning} and to \textit{246} with \textit{Importance Pruning} while the GPU training time is around \textit{200}. It should be pointed out that all communications are intrinsically sparse, reducing the overhead significantly. From our experimental evaluation, we argue that sparsity could more easily overcome the typical negative traits of asynchrony. At the same time, the communication overhead is mitigated since the processes in the system exchange sparse updates. Lastly, the concept of staleness-adaptive AsyncPSGD and delayed compensation strategies have been explored, but they did not improve statistical efficiency.

Secondly, we proposed a new activation function called \textit{All-ReLU} to boost sparse MLPs performances. All-ReLU has shown promising results, outperforming ReLU in all five datasets, across various domains, without adding any extra computational complexity to the training procedure. To present some outstanding results, it has shown to significantly increase accuracy on test data for CIFAR10 and Madalon by more than 2\% when compared to ReLU. The benefit on CIFAR10 becomes even more visible if we train the network longer. In this case, All-ReLU increases the accuracy by 3.4 \% when compared to the classic ReLU. Moreover, if the model is trained longer, All-ReLU gets on par results (72-73\%) with SReLU, achieving an accuracy close to the 73-74\% reported in original SET \citep{Mocanu_2018}. For image datasets, we hypothesized that the higher performance of our activation function might be caused by its ability to capture the feature shift that is common in this data. Our results are in line with independent parallel literature on sparse neural networks, as we demonstrate that All-ReLU achieves better performance by enriching the gradient flow during the training process. These findings make stronger the contribution of both works and the generality of the concept.

Lastly, we proposed a metric to define neuron importance which can be employed to remarkably shrink the number of parameters via \textit{Importance Pruning}, an active pruning strategy. The sparsity level of the network can be increased without performance loss using our proposed method, which reduces computation time and memory requirements. The combination of the new activation and \textit{Importance Pruning} has been tested across all datasets, resulting in better or comparable outcomes when pruning is included. The most interesting case has been revealed for Madelon data, where the combination of the two methodologies has significantly improved performance up to 77\%. This happens because the artificial dataset contains many redundant features which are eliminated, and it might imply that the neuron importance metric is useful for implicit feature selection.

The three methods are complementary and can be combined to obtain large scale sparse MLPs. In \autoref{subsection:extreme} we performed some experiments to train extremely large networks (in terms of neurons), and we managed to train a sparse MLP with 50M neurons on a single machine. Thanks to this investigation, we could identify many important limitations in order to open several new research directions. Our contributions allow advancing the state-of-the-art in representational power (i.e. number of neurons) of artificial neural networks. Currently, up to our knowledge, the largest ANNs, built on supercomputers, accommodate the size of a frog's brain (about 16 million neurons)(\cite{goodfellow2016deep}). After some technical challenges are overcome, with sparse neural networks, we may create on the same supercomputers ANNs close to the human brain size (about 80 billion neurons).

There are several directions for future work. The concept of staleness-adaptive AsyncPSGD for the first training phase has been under-explored for a high number of workers. Although these adaptations do not seem to help in our experiments, continuing to investigate asynchrony-aware SGD is of interest for very sparse large models. Future research directions also include investigating the nature of sparse training with more extensive experiments on various model architecture, as CNNs and Transformers. The latter would likely benefit the most since they use large dense layers. Additionally, it would be intriguing to consider a decentralized architecture with no parameter server involved. From an implementation point of view, it would be great to develop the parallelization in C++, in order to achieve better performance and overcome some sloppy characteristics of Python, or to incorporate customized sparse matrix operations developed for GPUs \citep{Gale2020SparseGK}. Lastly, we need to adopt distributed settings if we want to outstretch the size of ANNs and approach the human brain's size. A clear limitation for \textit{All-ReLU} consists in choosing the slope $\alpha$. Although we provide some practical advice (see \autoref{subsection:gridsearch}), it would be interesting to find a way to tune this parameter before training automatically.

We believe that our research opens the path for obtaining better performance for current state-of-the-art sparse training research in terms of accuracy, computational requirements, and energy costs. With regard to the latter, people use artificial intelligence for climate change, but they do not improve deep learning to save energy. With our work, we hope to raise more awareness concerning this problem and show that it is possible to pursue sustainable supercomputing. Finally, it can pave the way to develop much larger neural networks with billion of neurons which can help us to tackle challenging problems in complex domains such as health care.

\section{Methods}
\label{sec:methods}

\subsection*{Sparse Neural Networks Training}
\label{subsection:sparsetraining}
Fully connected neural networks have been shown to have a substantial number of redundant parameters, and, in some
cases, more than 95\% of the parameters can be predicted
from the remaining ones without accuracy loss (\cite{denil2013predicting}). In early work on sparsification, Optimal Brain Damage \cite{braindamage} and Optimal Brain Surgeon (\cite{optimalbrainsurgeon}) use gradient methods to sparsify networks during training. They observed that a sparse network demonstrated several advantages over its dense counterparts, such as better generalisation, reduced memory footprint and faster inference time. Additionally, the methodology of sparsification includes Hebbian pruning (\cite{droput_a_simple_way}), matrix decomposition (\cite{Liu2015SparseCN}),
and graph techniques (\cite{doi:10.1137/1.9780898719918, Kepner2017EnablingMD, Kepner_2018,Kumar2018IBMPA, Prabhu2018DeepEN}).

\subsubsection*{Dense-to-Sparse Training}
\label{subsubsection:densetosparse}
Recently, more and more studies attempted to obtain memory and computational efficiency methods for the inference phase of deep neural networks. Numerous post-pruning techniques (dense-to-sparse training) have been proposed to reduce the number of parameters and speed up the inference phase across a broad range of neural network architectures \citep{han2015learning, Srinivas2015DatafreePP, Han2016DeepCC, narang2017exploring, zhu2017prune, zhou2020deconstructing}; yet, these approaches require to fully train the dense model first. Several methods strove to learn the sparse networks during training \citep{louizos2018learning, wen2018learning, liu2020dynamic, Kusupati2020SoftTW, Yuan2021GrowingED}. However, these techniques begin with (or, are using at some moment during training) a fully-connected model, and as a consequence, they are not memory efficient. Another viable way is \textit{one-shot pruning}, which aims to find sparse neural networks by pruning once before the main training phase \citep{lee2019snip, lee2020signal, wang2020picking}. In this setting, at least one iteration of the dense model requires to be trained to identify the sparse sub-networks, and therefore the pruning process is unfeasible for memory-limited scenarios. Additionally, this method cannot meet the performance of dynamic sparse training, especially at extreme sparsity levels (\cite{wang2020picking}).

\subsubsection*{Sparse-to-Sparse Training}
\label{subsubsection:sparsetosparse}
The aforementioned issues can be naturally overcome by training intrinsically sparse neural networks from scratch to obtain the efficiency for both the training and inference phases. With respect to \textit{Static Sparse Training}, \cite{Mocanu2016xbm} introduced intrinsically sparse networks by exploring the scale-free and small-world topological properties in Restricted Boltzmann Machines. Later, some works focus on designing sparse CNNs based on Expander graphs and show comparable performance (\cite{Prabhu2018DeepEN, Kepner_2019}). In (\cite{Alford2019TrainingBO}) they tests pruning-based topologies, as well as RadiX-Nets from (\cite{Kepner_2019}). Once more, results show that these static sparse networks obtain accuracies comparable to dense networks, but extreme levels of sparsity cause instability in training. A training technique that allows for sparsity throughout the entire training process in a \textit{dynamic} fashion (sparse-to-sparse training) was first introduced in \cite{phdmocanu, Mocanu_2018}, with a simple and effective procedure called Sparse Evolutionary Training (SET) which uses magnitude-based pruning and random growth at the end of each training epoch. After that, in Deep Rewiring (DeepR) (\cite{Bellec2017DeepRT}), the authors rigorously combined dynamic sparse parameterisation with stochastic parameter updates for training; however, this approach is computationally expensive and difficult to deploy on large models. More recent work like (\cite{Mostafa2019ParameterET}) includes the cross-layer redistribution of weights, while in  (\cite{Dettmers2019SparseNF}) they present a similar approach which uses gradient information and momentum, significantly improving performances on various Convolutional Neural Networks (CNNs) models. Nevertheless, these additional calculations can result in a significant amount of extra computation. Very recently, based on the Lottery Ticket Hypothesis (\cite{frankle2018lottery}), RigL (\cite{Evci2019RiggingTL, top}) was introduced as a novel method for training sparse models without the need of a "lucky initialisation"; it can match and sometimes exceed the performance of pruning based approaches. Lastly, in (\cite{Liu2021SelfishSR}), the authors propose an approach to successfully train sparse Recurrent Neural Networks (RNNs) with a stable number of floating-point operations (FLOPs) and a fixed parameters count.

\subsection*{Parallel Training of Deep Neural Networks}
\begin{figure}[!htbp]
\centering
\includegraphics[width=0.95\textwidth]{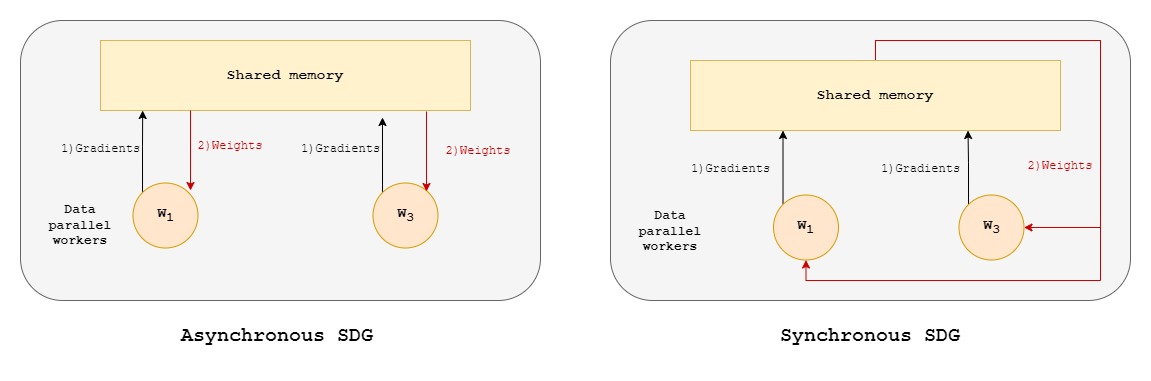}
\caption{Asynchronous vs. Synchronous SGD in standard shared memory architecture. }
\label{fig:asyncvssync}
\end{figure}

Accelerating training for Deep Neural Networks (DNNs) is a daunting challenge and techniques range from distributed algorithms to hardware optimisations. Stochastic Gradient Descent (SGD) (\cite{SGD}) together with backpropagation, and in particular, its mini-batch variants (\cite{bottou}) are the de-facto methods to train DNNs. SGD, however, is inherently sequential with a dependency across iterations and, this dependency limits parallelism. In (\cite{BenNun2018DemystifyingPA}), the authors provide an extensive survey about the vast catalogue of parallelisation approaches in deep learning. There are three prominent strategies to partition the learning phase of a model: partitioning by input samples (\textit{data parallelism}), by network structure (\textit{model parallelism}), and by layer (\textit{pipelining}). Data parallelism can be easily implemented, and it is, therefore, the most widely used implementation strategy on multi-GPUs (\cite{LI201695}). We have explored this option focusing on CPUs only, where each core utilises the same sparse model to train on different data subsets. The replicas communicate updates through a centralised \textit{parameter server} (shared memory) which maintains the current state of all parameters for the sparse model. In this architecture, there is no synchronisation between CPU cores during the forward pass, but the gradients must be synchronised for the weights update. As for the parallelisation of SGD algorithms, one can choose to do it in either a synchronous or asynchronous way (\autoref{fig:asyncvssync}).

\subsubsection*{Synchronous Parallel SGD}
\label{subsubsection:sync}
In a \textit{shared parameter server} system, the local workers can compute the gradients over their mini-batch of data, and then add the gradients to the global model; this approach is commonly referred to as \textit{Synchronous} Parallel SGD (SyncPSGD) due to its barrier-based nature. In its standard form, SyncPSGD has scalability issues due to the waiting time that is inherent in the aggregation when independent workers compute with different speed (\cite{mindthestepasyncpsgd}).

Recent work in \cite{goyal2017accurate, 32Kimagenet}, explores the various limitation of this approach, as in general models trained with large mini-batches often do not generalise well. To overcome the communication overhead, local SGD has recently attracted increasing research interest \cite{zhang2016parallel, mcmahan2017communicationefficient, yu2018parallel, stich2019local} where data is partitioned among computation nodes, and the nodes compute local updates with periodically exchanging the model among the workers to perform averaging. To address the current generalisation issue of large-batch training and improve the scalability of SyncPSGD, in \cite{lin2018dont} the authors proposed Post-Local SGD, where the classic large mini-batch SGD is followed by a local SGD phase which allows workers to independently update their models for a few steps before synchronising. Moreover, (\cite{gupta2020stochastic}) introduces a variant where they adapt Stochastic Weight Averaging \cite{dist} to accelerate DNNs training. In the second phase, each worker refines its network separately, and at the end, they average the weights of the resulting models to produce the final result. 

\subsubsection*{Asynchronous Parallel SGD}
\label{subsubsection:async}
An alternative type of parallelisation is Asynchronous Parallel SGD (AsyncPSGD) \cite{dist}, in which workers fetch and update the shared model independently of each other. Hence, the training procedure has no barrier imposed. Notably, for sparse problems, Hogwild method (\cite{niu2011hogwild}) shows that updating only the relevant parameters without any synchronisation could guarantee a nearly linear speedup with the number of processors (\cite{niu2011hogwild}). Although AsyncPSGD can achieve faster speed due to the absence of waiting overhead, the lack of coordination implies that gradients may be computed on \textit{stale} (old) version of the weights, which leads to statistical inefficiency. This problem is well-known, and some researchers have analysed its negative effect on the convergence speed \cite{avron2015revisiting,lian2019asynchronous}. Another critical factor to monitor when considering different scales of asynchrony is that it introduces momentum to the SGD update, called the \textit{implicit momentum} (\cite{mitliagkas2016asynchrony}). 

Recent studies in \cite{lan2018asynchronous,mindthestepasyncpsgd, zheng2020asynchronous} proposes different staleness-adaptive SGD algorithms to reduce the negative impact of asynchrony and approach the performance of sequential SGD. Moreover, they allow for fine-tuning the implicit \textit{momentum} and increase the number of workers while maintaining statistical efficiency.

\subsubsection*{Parallel Training of Sparse Networks}
\label{subsubsection:parallesparsenet}
To reduce the communication overhead in parallel distributed DNN training, various quantisation techniques and sparse gradient updates have been developed \cite{wangni2017gradient,alistarh2017qsgd, stich2018sparsified}. In this regard, one significant advantage of sparse models is that the sparse gradient communication is automatically at hand. Related work on parallelisation for sparse DNN is presented in (\cite{dataparallelsparseNNMITchallenge}) as a solution to the Sparse DNN Challenge posed by MIT/IEEE/Amazon (\cite{Kepner2019SparseDN}). However, their work is focused on sparse neural networks created using RadiX-Net (\cite{Kepner_2019}) and trained on GPU which do not evolve the topology over time. Moreover, their solution is implemented in Tensorflow, where sparse layers are represented by dense matrices with a mask over weights. Similarly, in (\cite{sparseinference}), the authors devised a parallel strategy for large sparse neural networks (up to 125 millions of neurons), but even if they employ truly sparse matrices their approach is focused on the inference phase only. Further related work for sparse neural network inference is presented in (\cite{9286206, Pawowski2020CombinatorialTF, HasanzadehMofrad2020StudyingTE}). In short, none of the available parallel training algorithms is designed for training truly sparse dynamic neural networks.

\subsection*{Activation Functions}
Activation functions determine the output of a deep learning model, its accuracy, and also the computational efficiency, which can make or break a large scale neural network. They are essential for an artificial neural network to help the model learn complex non-linear patterns in the data.
\subsubsection*{ReLU}
Rectified Linear Unit (ReLU) is one of the most widely
used activation functions in neural networks (\cite{Glorot2011DeepSR, RectifiedLinearUnits}), which can effectively solve the problem of vanishing gradient (small gradient prevents the weight from altering its value) and slow training time of saturated activation function. Its mathematical expression is:
\begin{equation}
\label{equation:relu}
f(x_i) = \begin{cases} 
0 & x_i\leq 0 \\
x_i & x_i > 0 
\end{cases}
\end{equation}

and its derivatives can be easily calculated with a meagre computational cost; this is a desirable advantage for choosing ReLU to speed up the training. When the input value $x_i$ is lower than zero, the resulting derivative will also be zero, leading to a disconnection of the neuron (zero-sparsity). Disconnecting some neurons may reduce overfitting; however, this will hinder the neural network from learning in some cases due to the neurons death problem. The ReLU function also keeps the mean activation value to be greater than zero, which makes it difficult for the network to determine the direction with the fastest gradient drop in the backpropagation process, thus affecting the network convergence. This "free" sparsity (in terms of neuron activations) obtained by adopting ReLU may represent an advantage for training fully-connected layers. However, this might be untrue for sparsely connected neurons since, in these settings, the fact they are not capable of capturing some significant aspect of the data on the negative side and the dead neurons problem could lead to a higher impact on performance overall. 

\subsubsection*{Leaky ReLU}
Leaky ReLU (LReLU) is a modification of ReLU which replaces the zero part of the domain in $[-\inf,0]$ by a low slope $\alpha$. The reason for using LReLU instead of ReLU is that constant zero gradients can also result in slow learning, as when a saturated neuron uses a sigmoid activation function. Additionally, others do not even activate. According to the authors in (\cite{Maas2013RectifierNI}), this sacrifice of the zero-sparsity might bring worse results than when the neurons are entirely deactivated, suggesting the leaky rectifiers' non-zero gradient does not substantially impact training optimisation. 

\subsubsection*{PReLU}
Parametric Rectified Linear Unit (PReLU) is proposed by (\cite{he2015delving}) and generalizes the traditional rectified unit. The authors reported that its performance was considerably better than ReLU in large-scale image classification tasks. It is the same as Leaky ReLU with the exception that the slope parameter is learned during training via backpropagation.

\subsubsection*{Activation Functions in Sparse Networks}
\label{subsubsection:snnetactivations}
It is crucial to question weather the activation functions currently used for densely connected networks still behave reliably in the sparse context. S-shaped rectified linear activation unit (SReLU) is a relatively little-known activation function suggested in (\cite{jin2015deep}) and used by Mocanu et al. (\cite{Mocanu_2018}) in their implementation of the SET algorithm. When compared with the most common activation functions, SReLU in SET models has shown to perform best at all sparsity levels (\cite{adam}) in various domain.

Very recently, in (\cite{Tessera2021KeepTG}) the authors have shown that SReLU and PReLU are more suitable activation function for sparse networks as they improve the networks gradient flow and achieve better accuracy when compared to the other activations. In this regard, they proposed a normalised measure of gradient flow called \textit{Effective Gradient Flow} (EGF), which is better suited to examining the training dynamics of sparse networks. Their results related to activation functions are in line with our findings. Previous papers were already suggesting that low gradient flow is an exacerbated issue in sparse networks \cite{wang2020picking, Evci2020GradientFI}, but they did not investigate its relation with activation functions.

\subsection*{Neuron Importance}
\label{subsection:importance}
The importance of hidden units in neural networks is still an open problem, crucial to understand neural networks' behaviour and to enhance the explainability of these black-box models. Many papers on dense deep networks speculate about the significance of a neuron towards a prediction. They tend to use the activation value of the hidden unit or its product with the gradient as a proxy for feature importance (e.g. \cite{zagoruyko2016paying}); however, both metrics can have undesirable outcomes. To overcome these problems, in (\cite{dhamdhere2018important}), the authors proposed the notion of \textit{conductance} to gain a better understanding of neuron relevance through extensive ablation studies.

We argue that a neuron importance metric can be straightforwardly identified in sparse neural networks trained with a sparse training algorithm where the topology evolves overtime to find the best weight configuration. The proposed metric is shown to be valuable since it can remove a big chunk of unimportant units and related connections with almost no loss in accuracy. More importantly, this metric can be simultaneously derived for all neurons without requiring expensive computations. To define the importance of a neuron, we borrow some terminology and ideas from graph theory and hence, we introduce them here. In network science, a hub is a high-degree node that occupies a central role in the overall organisation of a network. Hubs have a significantly larger number of links in comparison with other nodes in the network (\cite{barabasi2016network}). They can be found in many real networks, such as the brain (\cite{hubsbrain}) or the Internet. The loss of such well-connected hubs can be extremely devastating to network function. Given the role of hubs and their significance to networks, their locations and functions in the brain are of clear interest to neuroscientists. Accordingly, we would expect to find a similar biological structure in sparse ANNs as well.

\newpage
\section*{Data availability}
The data used in this paper are public datasets, freely available online, as reflected by their corresponding citations from \autoref{table:datasets}.

\section*{Code availability}
Prototype
software implementations of the models used in this study are freely available \href{https://github.com/SelimaC/large-scale-sparse-neural-networks}{online}.

\section*{Acknowledgement}
We thank the Google Cloud Platform Research Credits program for granting us the necessary resources to run the Extreme large sparse MLPs experiments.  

\section*{Competing interests}
The authors declare no competing interests.


\begin{algorithm}[!htbp]

\SetAlgoLined
\KwResult{Trained sparse model $\theta_s^f$ }
\KwIn{Number of workers $K$, $t=0$, $t'=0$, $phase=1$, $epoch=0$, sparsity level $\mathcal{S}$
\newline Step size $\eta$ and momentum $\mu$ 
\newline Training dataset $\mathcal{D}$ wit labels $\mathcal{I}$ ; Mini-batch size $\mathcal{B}$
\newline SGDSparseUpdate($\cdot$), a function that updates the weights using momentum SGD 
\newline TopologyEvolutionStep($\cdot$), a function that updates the sparse topology
\newline RetainValidUpdates($\cdot$), a function that retain only gradients applicable to the topology defined by $\theta_s^t$
\newline Epoch $\tau_1$ and $\tau_2$, at which to exit phase one and phase two respectively} 
\vspace{10pt}
\textbf{Phase 1:} \\
\underline{Worker \textit{k}} in [1,...,K]\\
/* Each worker shuffle its data partition $\mathcal{I}^{(k)}_{t}$ after each local epoch */ \\
\While{phase == 1}{
Sample a mini-batch $B^k$ from $\mathcal{I}^{(k)}_{t}$\\
Calculate worker gradient: $\nabla w^{(k)}_{t} = \frac{1}{\mathcal{B}} \sum_{i \in \mathcal{B}^k} \nabla w^i$ \\
Send gradients $\nabla w^{(k)}_{t}$ and time step $t$ to PS\\
Receive updated model from PS and time step $t'$\\
Update time step: $t=t'$ \\
}
\underline{Parameter server PS}\\
\While{$epoch \le \tau_1$}{
Receive gradients $\nabla w^{(k)}_{t}$ and time step $t$ from a ready worker $k$\\
Retain valid updates: $g$ = RetainValidUpdates($\nabla w^{(k)}_{t}$, $\theta_s^{t'}$)\\
Update model: $\theta_s^{t'+1}$ = $\theta_s^{t'}$ + SGDSparseUpdate($g$, $\eta$, $\mu$) \\

\If{ $t' \% (n \div \mathcal{B})== 0$}{
Update sparse topology: $\theta_s^{t'+1}$ = TopologyEvolutionStep($\theta_s^{t'}$) \\
Update epoch: $epoch=epoch+1$\\
}
Send updated model $\theta_s^{t'+1}$ and time step $t'$ to worker $k$\\
Update time step: $t'= t'+ 1$\\
}
Switch phase: $phase = 2$ \\
\vspace{10pt}
\textbf{Phase 2:} \;
/* Local training of K sparse models that evolve their topology separately */ \\
\While{$epoch \le \tau_2$}{
Each worker shuffles its data partition $\mathcal{I}^{(k)}_{t}$ \\
\For{\textit{k} in [1,...,K] in parallel}{
Sample a mini-batch $B^k$ from $\mathcal{I}^{(k)}_{t}$ \\
Calculate worker gradient: $\nabla w^{(k)}_{t} = \frac{1}{B} \sum_{i \in B^k} \nabla w^i$ \\
Update model: $\theta_s^{t+1}$ = $\theta_s^t$ + SGDSparseUpdate($\nabla w^{(k)}_{t}$, $\eta$, $\mu$) \\
Update sparse topology: $\theta_s^{t+1}$ = TopologyEvolutionStep($\theta_s^t$) \\
}
Update time step: t=t+1; epoch=epoch+1 \\
}
We get K different models at the end of phase 2\\
Produce averaged model $\theta_{s}^f$ and select a fraction $\mathcal{S}$ of weights with bigger magnitude for each layer \\

\caption{WASAP-SGD}
\label{algorithm:wasap}
\end{algorithm}

\begin{algorithm}[!htbp]
\SetAlgoLined
\KwResult{Trained sparse model }

\KwIn{An ANN model with $L$ layers
\newline Weight $\theta_s$, sparsity $\mathcal{S}$
\newline pruning rate $\zeta$, pruning step $p$, starting pruning epoch $\tau$, and threshold $t$ }
\% Sparse initialization\;
\For{each fully-connected (FC) layer l}{
replace \textit{l} with a Sparse Connected Layer having a Erd\H{o}s-Rényi topology
}
\% Training\;
\For{each training epoch e}{
Perform standard training procedure\;
Perform weights update\;

\If{e \% $p$ == 0 and $ e \ge \tau$}{
\% Perform Importance Pruning\;
\For{each SC layer of the ANN }{
Calculate importance (I) per each neuron\;
Remove incoming weights of neurons where $I < t$
}
}
\% Weight pruning-regrowing cycle\;
\For{each SC layer of the ANN }{
Remove a fraction $\zeta$ of the smallest positive weights\;
Remove a fraction $\zeta$ of the largest negative weights\;
Add randomly new weights in the equivalent amount as the one removed previously\;
}
}
\caption{Sparse evolutionary training (SET) with Importance Pruning}
\label{algorithm:pruning}
\end{algorithm}

\newpage

\bibliography{refs}

\newpage

\section{Supplementary Information}

\subsection{From SReLU to All-ReLU}
\label{srelutoallrelu}

\begin{figure*}[!htbp]
\subfloat{%
\includegraphics[width=0.3\textwidth]{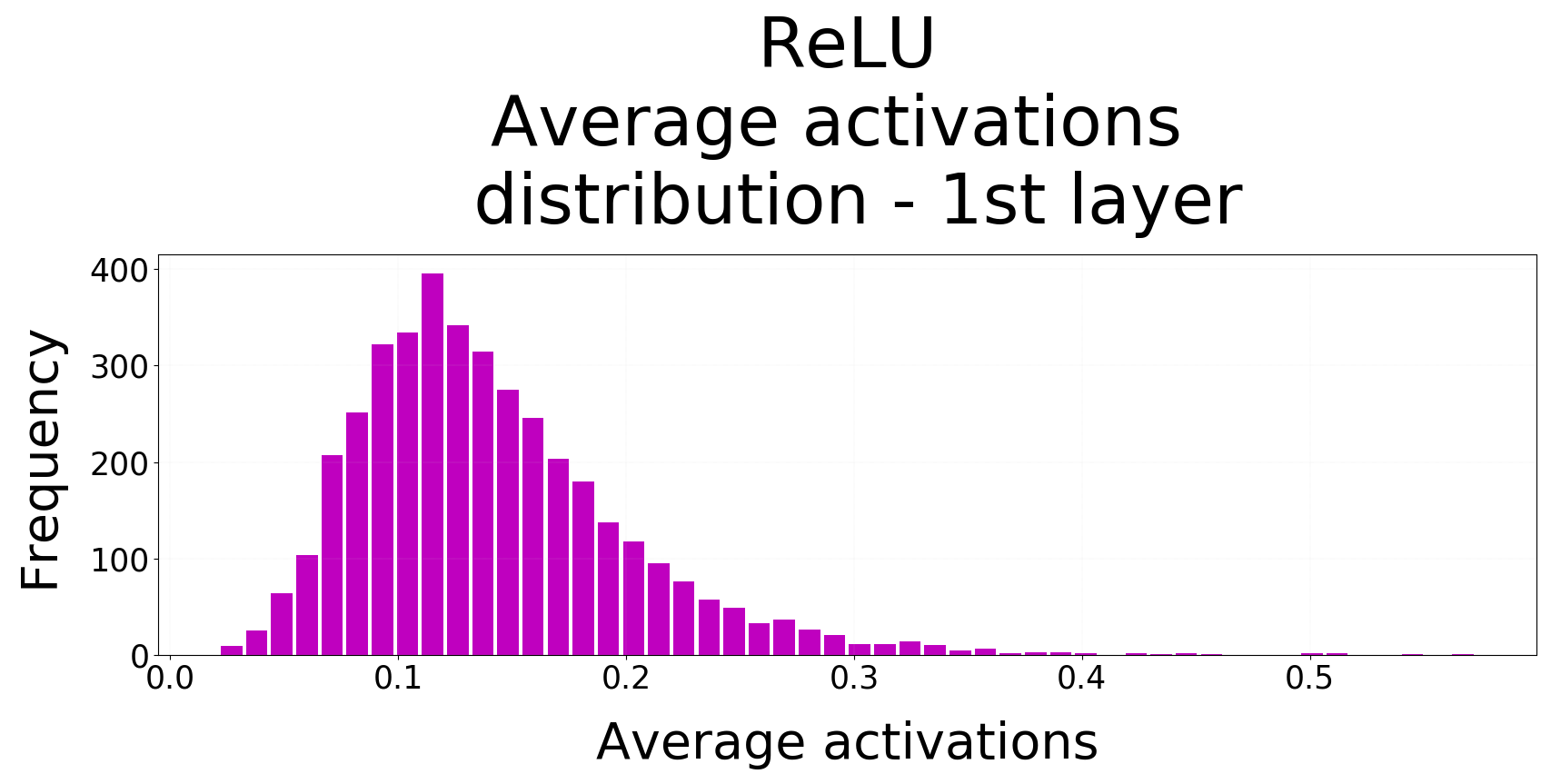}}
\hspace{\fill}
\subfloat{%
\includegraphics[width=0.3\textwidth]{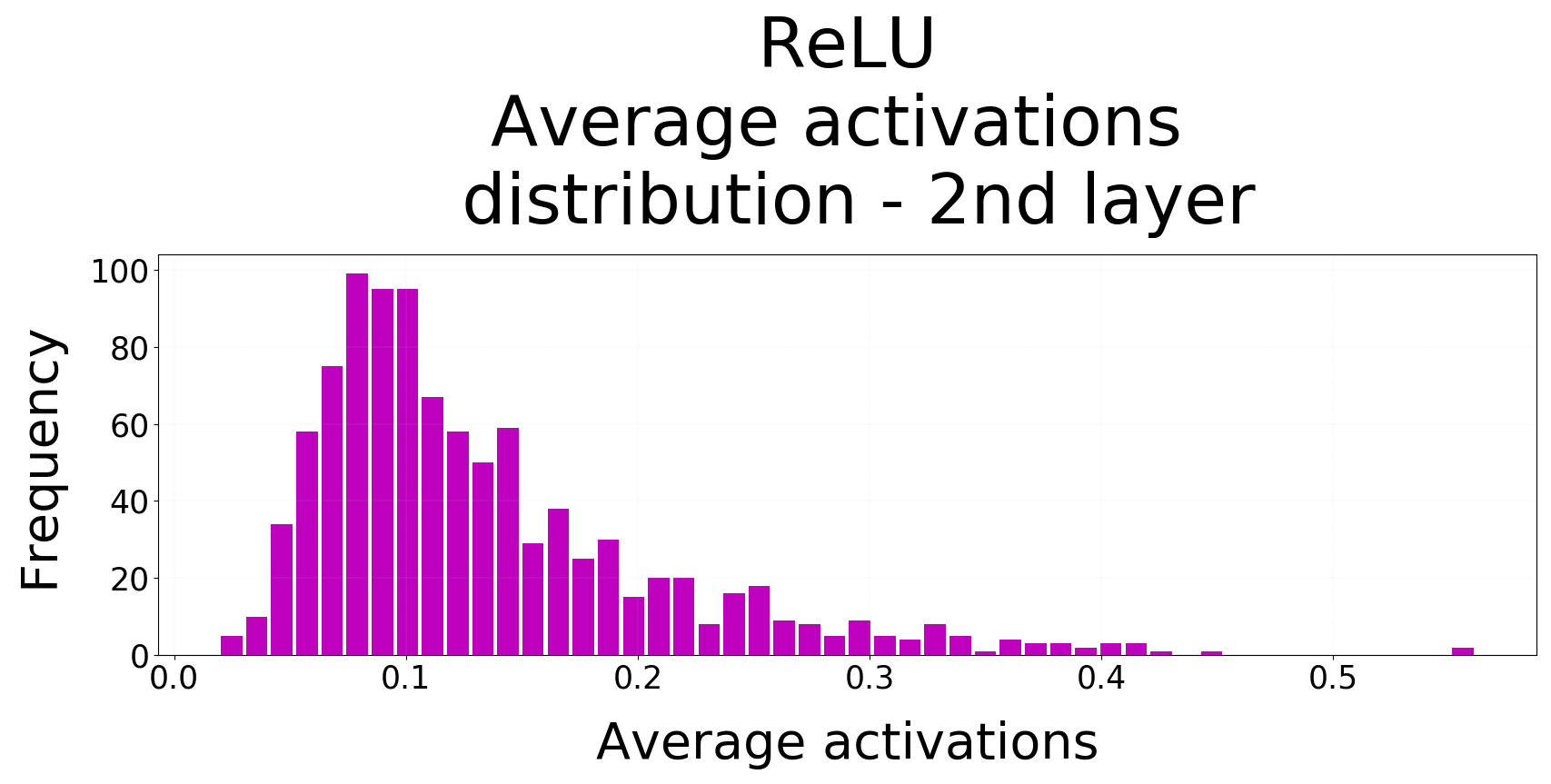}}
\hspace{\fill}
\subfloat{%
\includegraphics[width=0.3\textwidth]{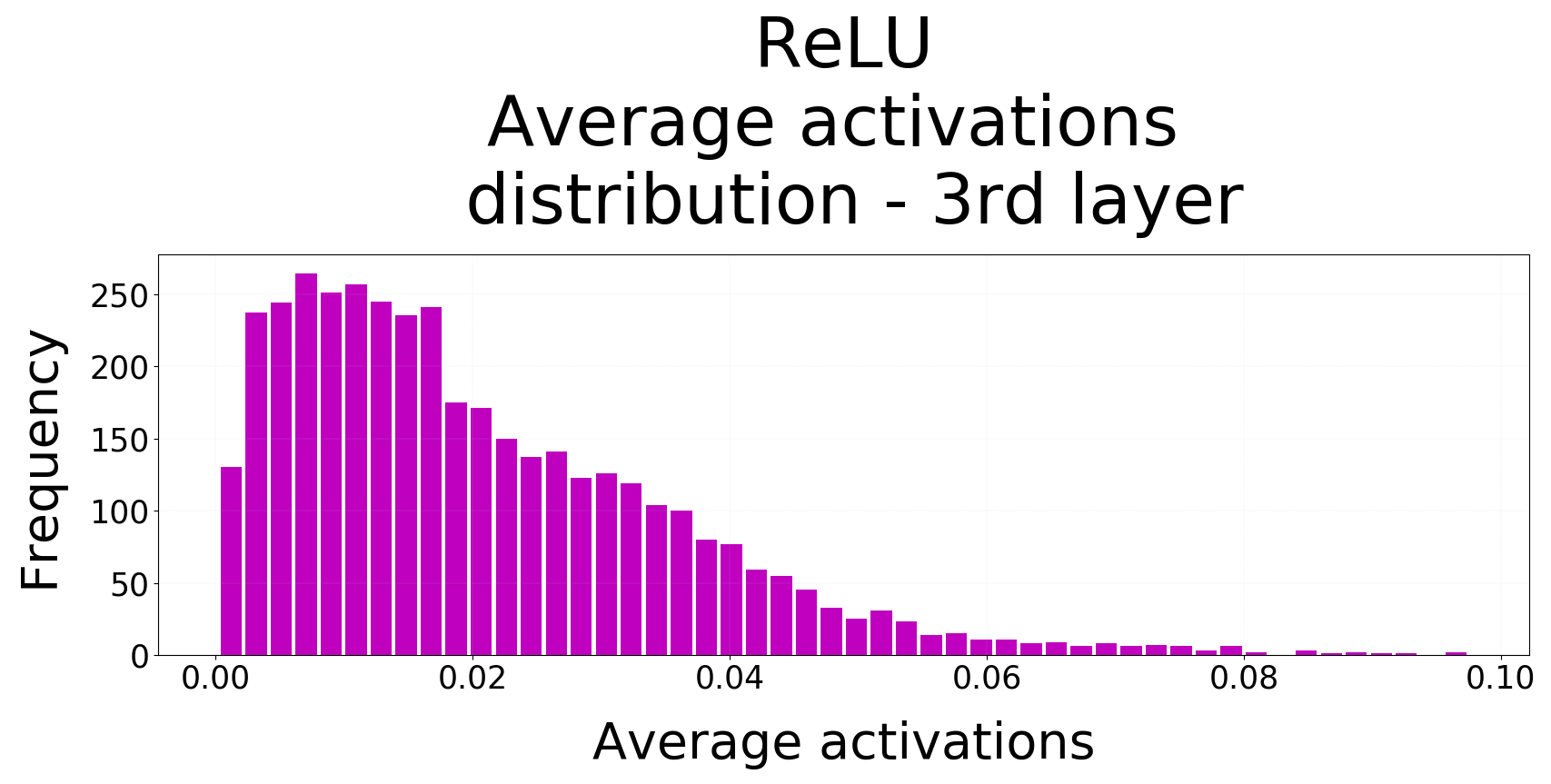}}
\caption{Average activations distribution for a 3-layers sparse MLP on CIFAR10 trained with ReLU and \textit{momentum} SGD after 1000 epochs.}
\label{fig:activations_1}
\end{figure*}

\begin{figure*}[!htbp]
\subfloat{%
\includegraphics[width=0.3\textwidth]{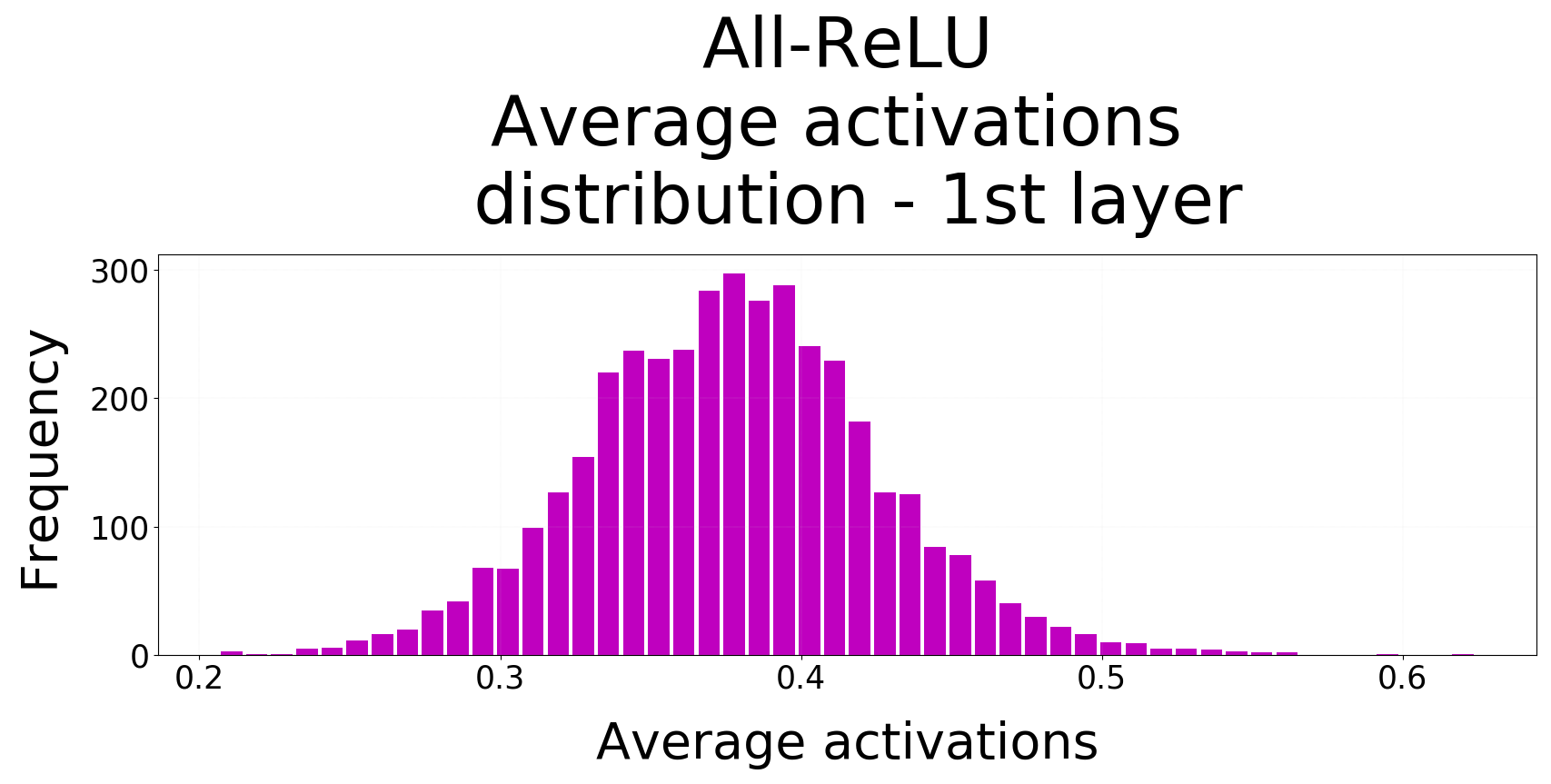}}
\hspace{\fill}
\subfloat{%
\includegraphics[width=0.3\textwidth]{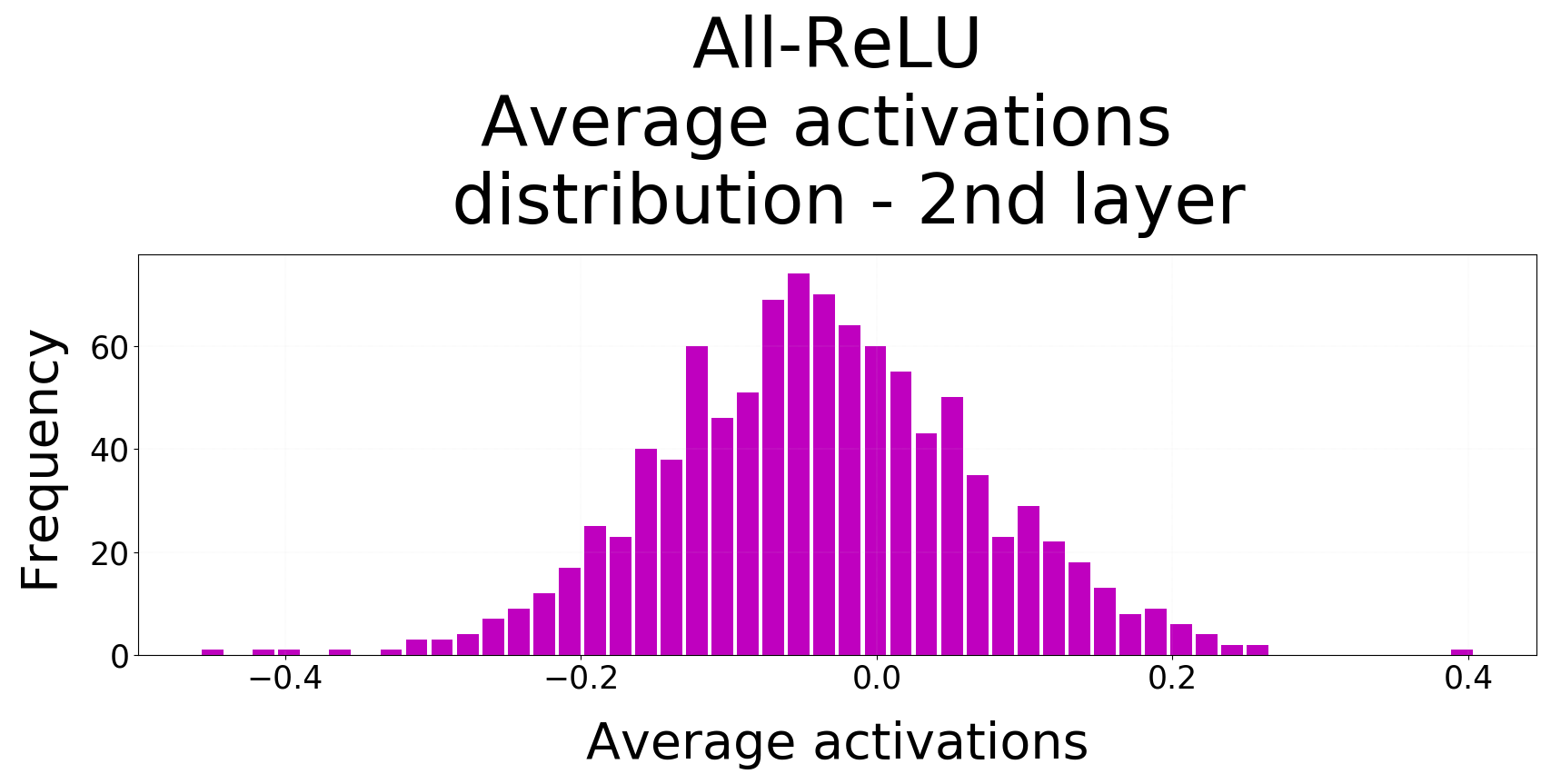}}
\hspace{\fill}
\subfloat{%
\includegraphics[width=0.3\textwidth]{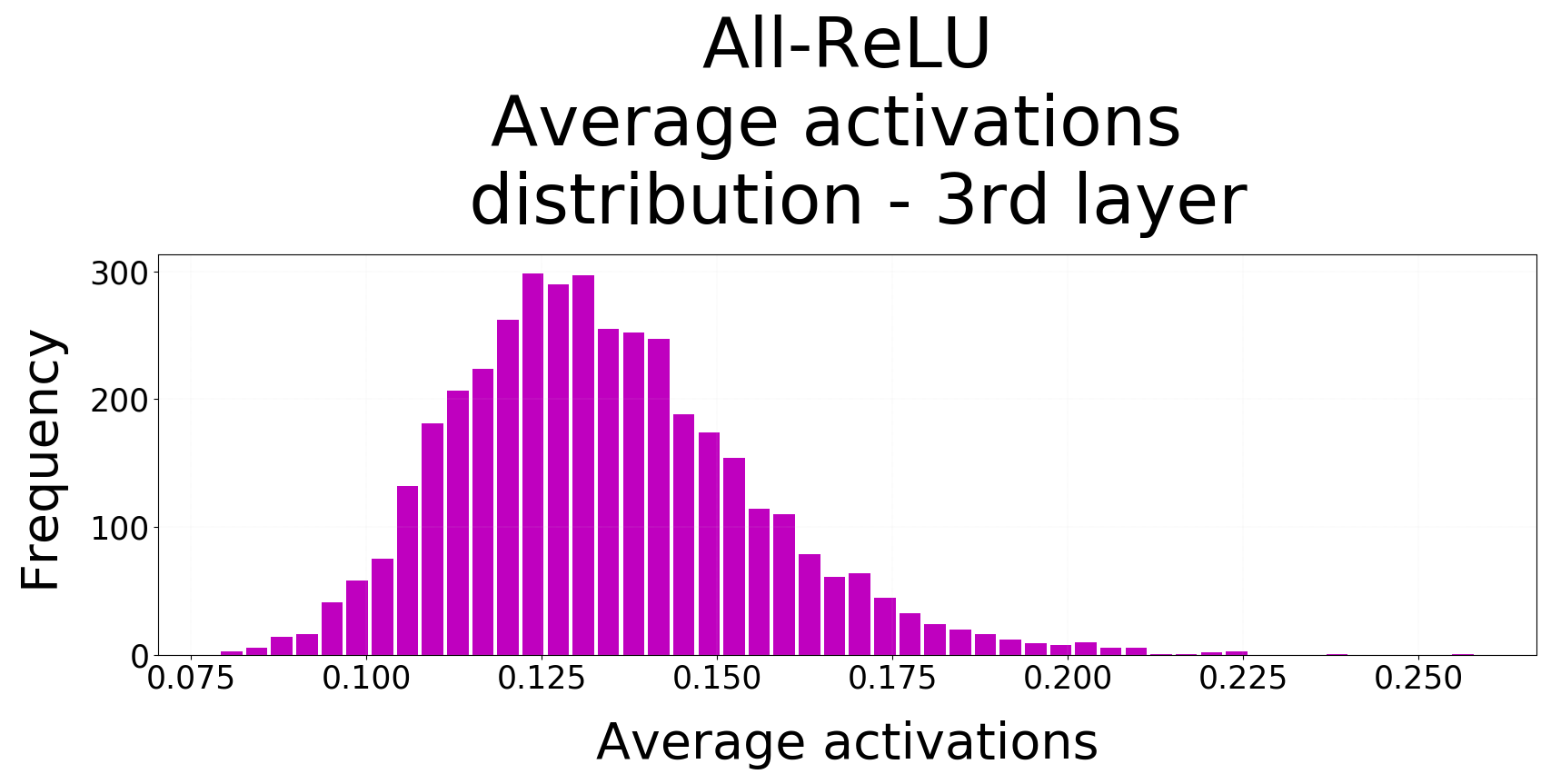}}
\caption{Average activations distribution for a 3-layers sparse MLP on CIFAR10 trained with All-ReLU and \textit{momentum} SGD after 1000 epochs.}
\label{fig:activations_2}
\end{figure*}

\begin{figure*}[!htbp]
\subfloat{%
\includegraphics[width=0.3\textwidth]{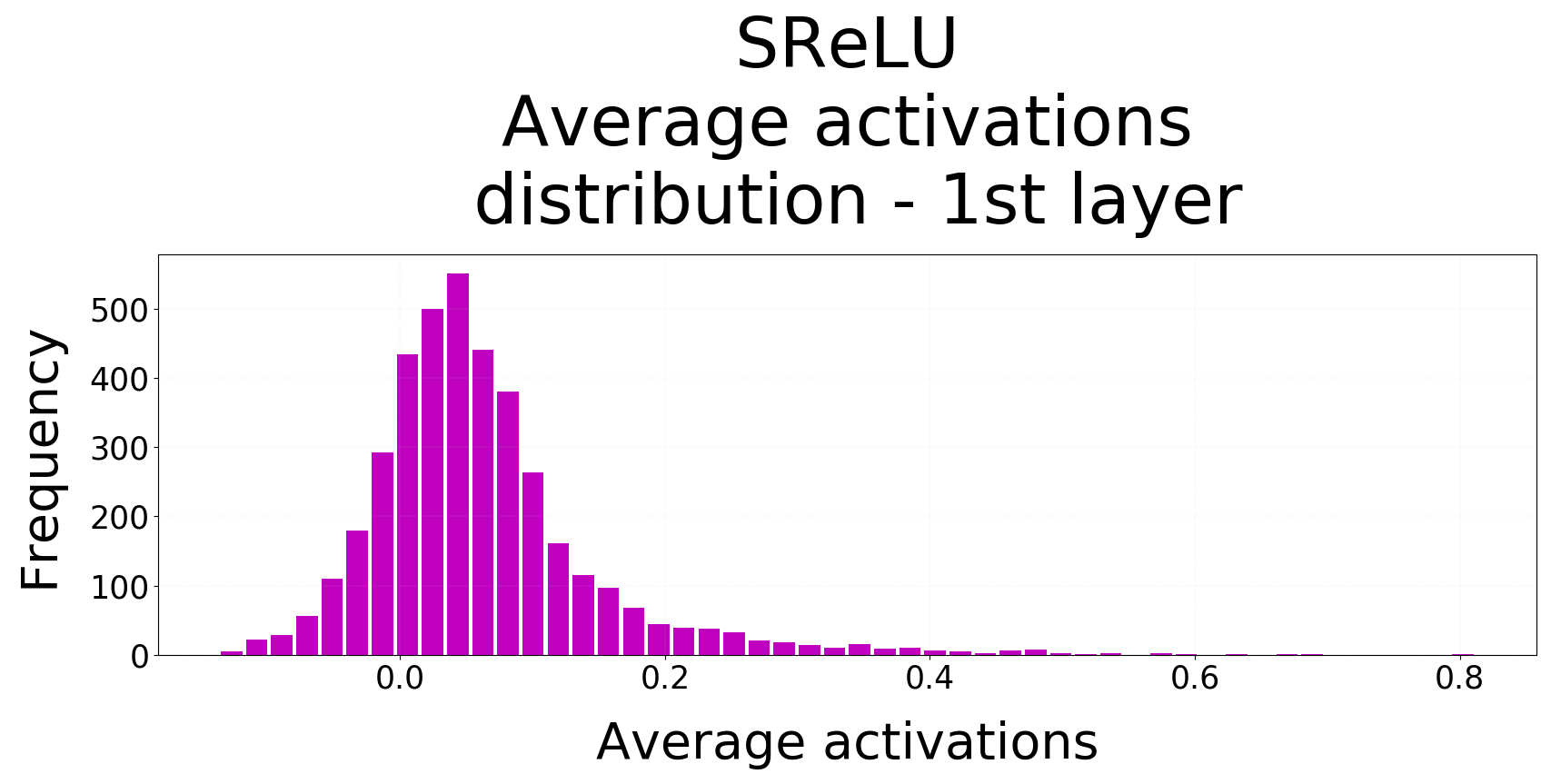}}
\hspace{\fill}
\subfloat{%
\includegraphics[width=0.3\textwidth]{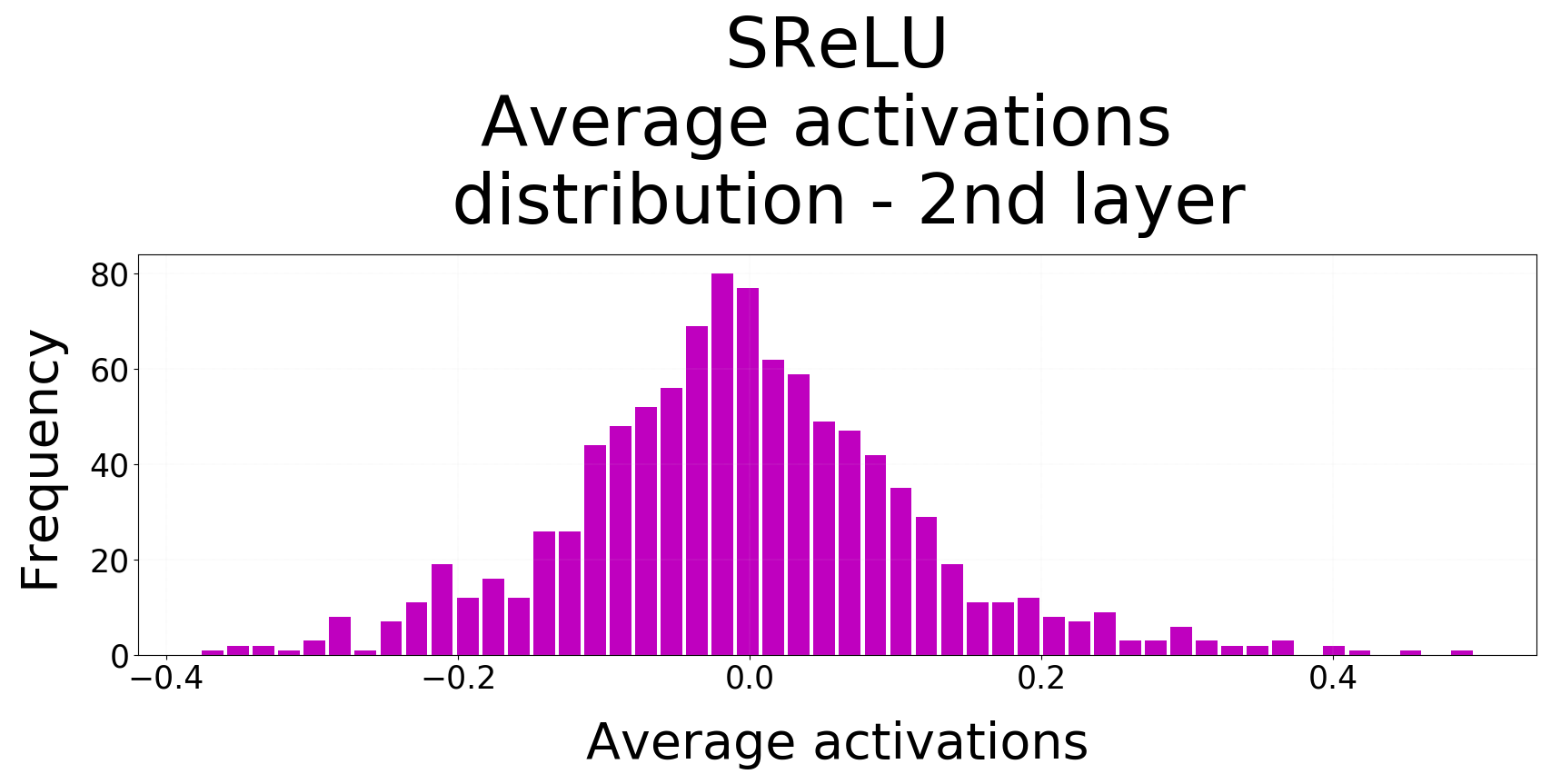}}
\hspace{\fill}
\subfloat{%
\includegraphics[width=0.3\textwidth]{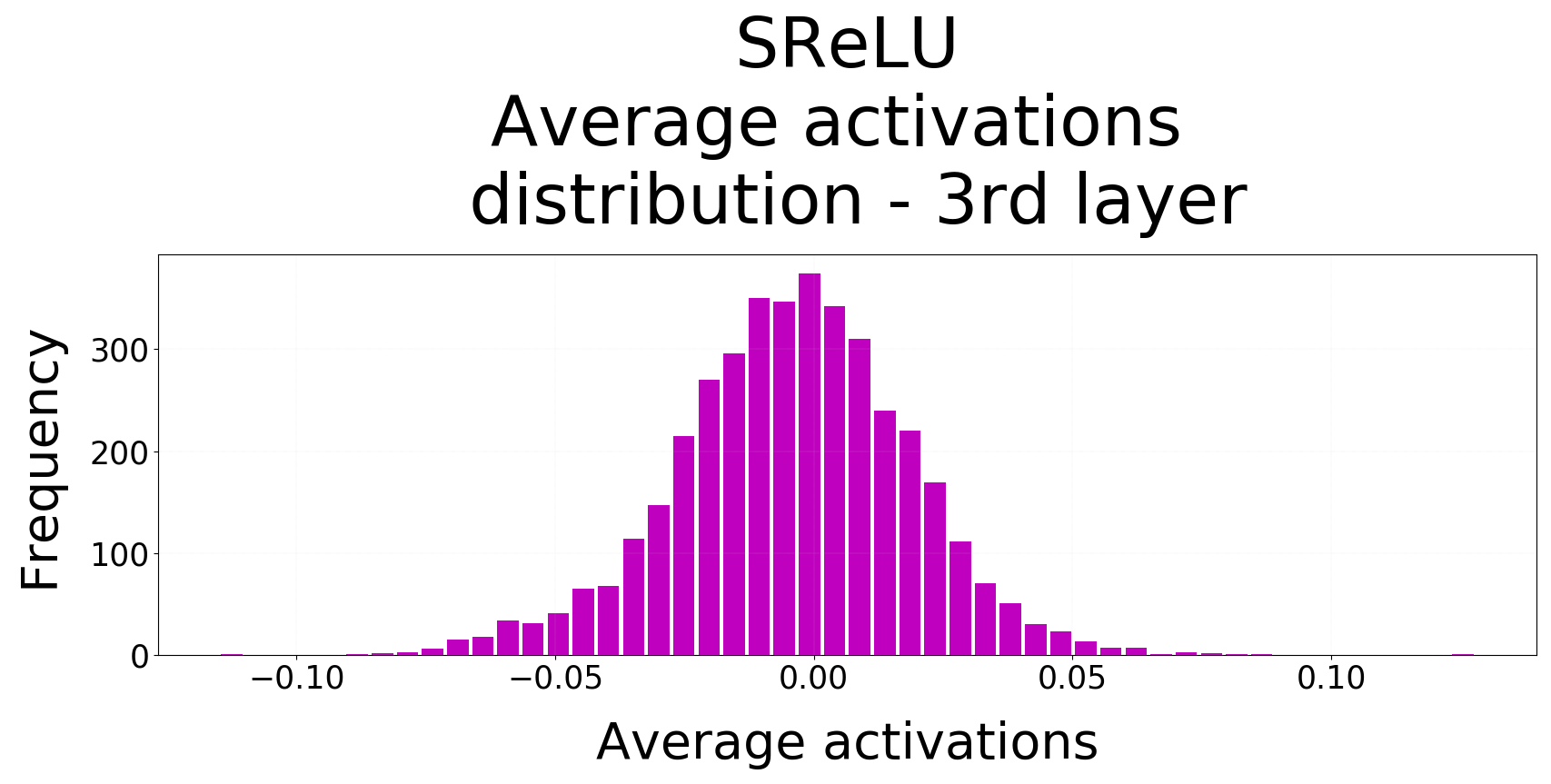}}
\caption{Average activations distribution for a 3-layers sparse MLP on CIFAR10 trained with SReLU and \textit{momentum} SGD after 1000 epochs.}
\label{fig:activations_3}
\end{figure*}

\begin{figure*}[!htbp]
\subfloat{%
\includegraphics[width=0.3\textwidth]{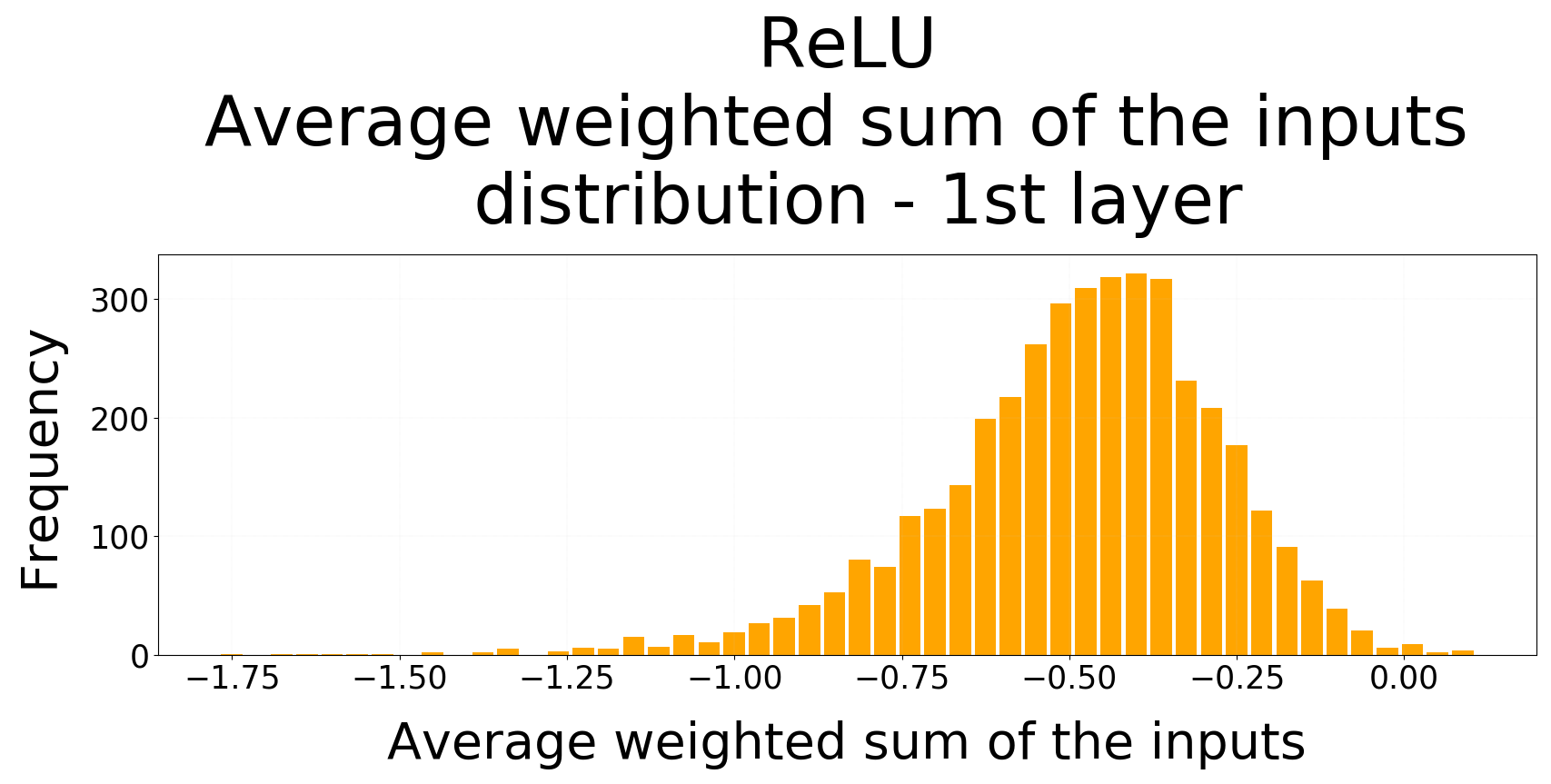}}
\hspace{\fill}
\subfloat{%
\includegraphics[width=0.3\textwidth]{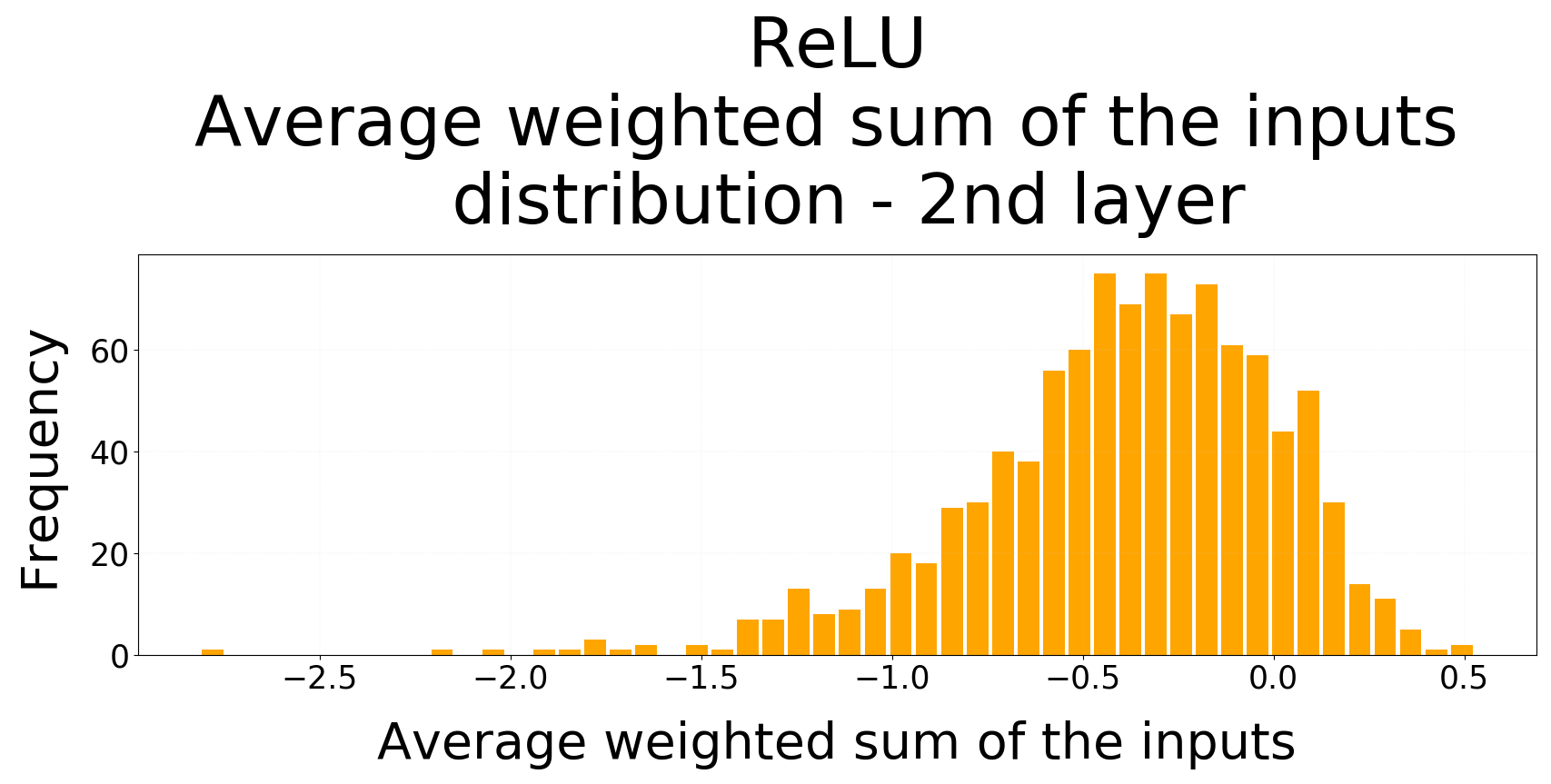}}
\hspace{\fill}
\subfloat{%
\includegraphics[width=0.3\textwidth]{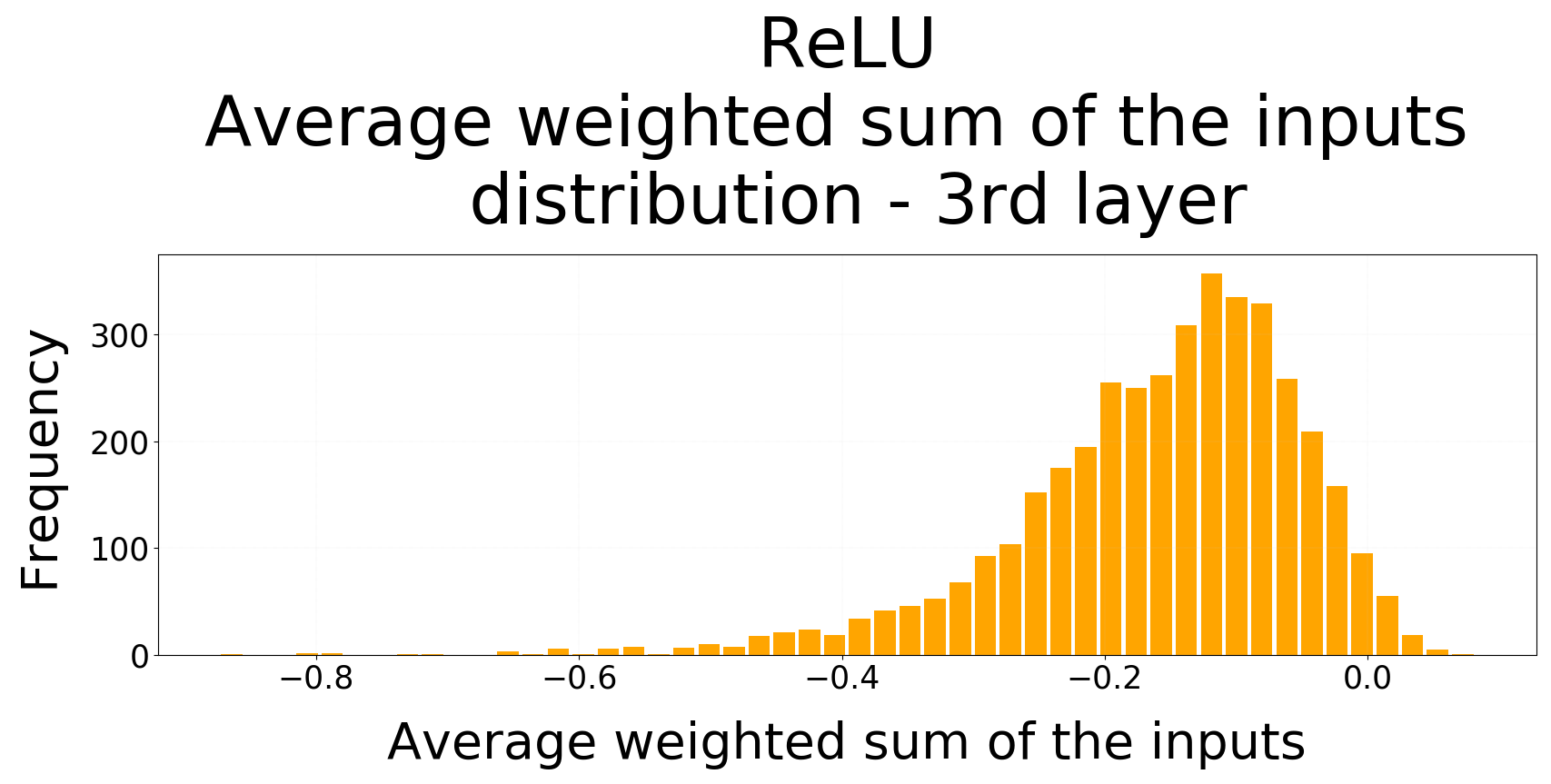}}
\caption{Average weighted sum of the inputs distribution for a 3-layers sparse MLP on CIFAR10 trained with ReLU and \textit{momentum} SGD after 1000 epochs.}
\label{fig:activations_4}
\end{figure*}

\begin{figure*}[!htbp]
\subfloat{%
\includegraphics[width=0.3\textwidth]{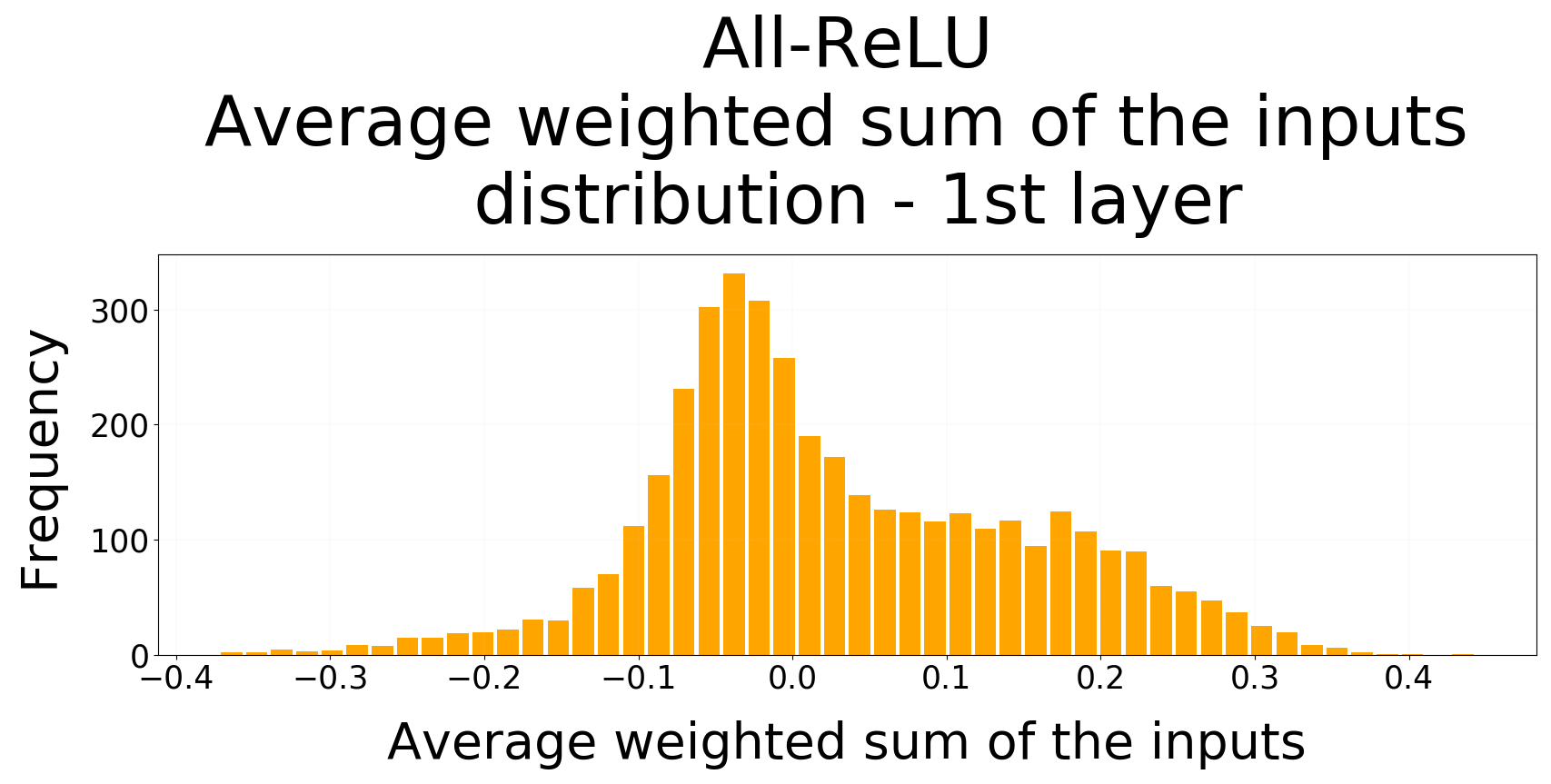}}
\hspace{\fill}
\subfloat{%
\includegraphics[width=0.3\textwidth]{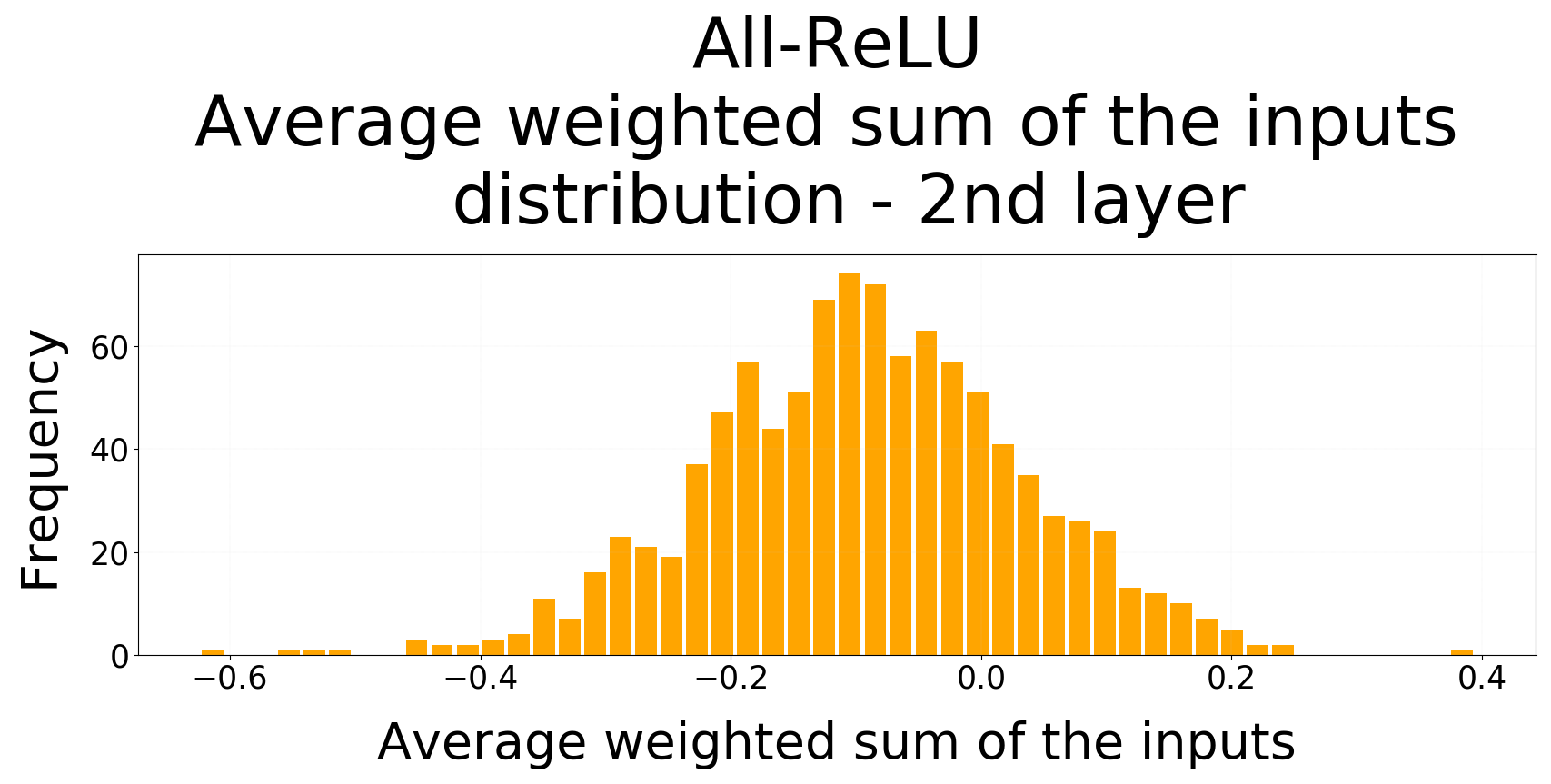}}
\hspace{\fill}
\subfloat{%
\includegraphics[width=0.3\textwidth]{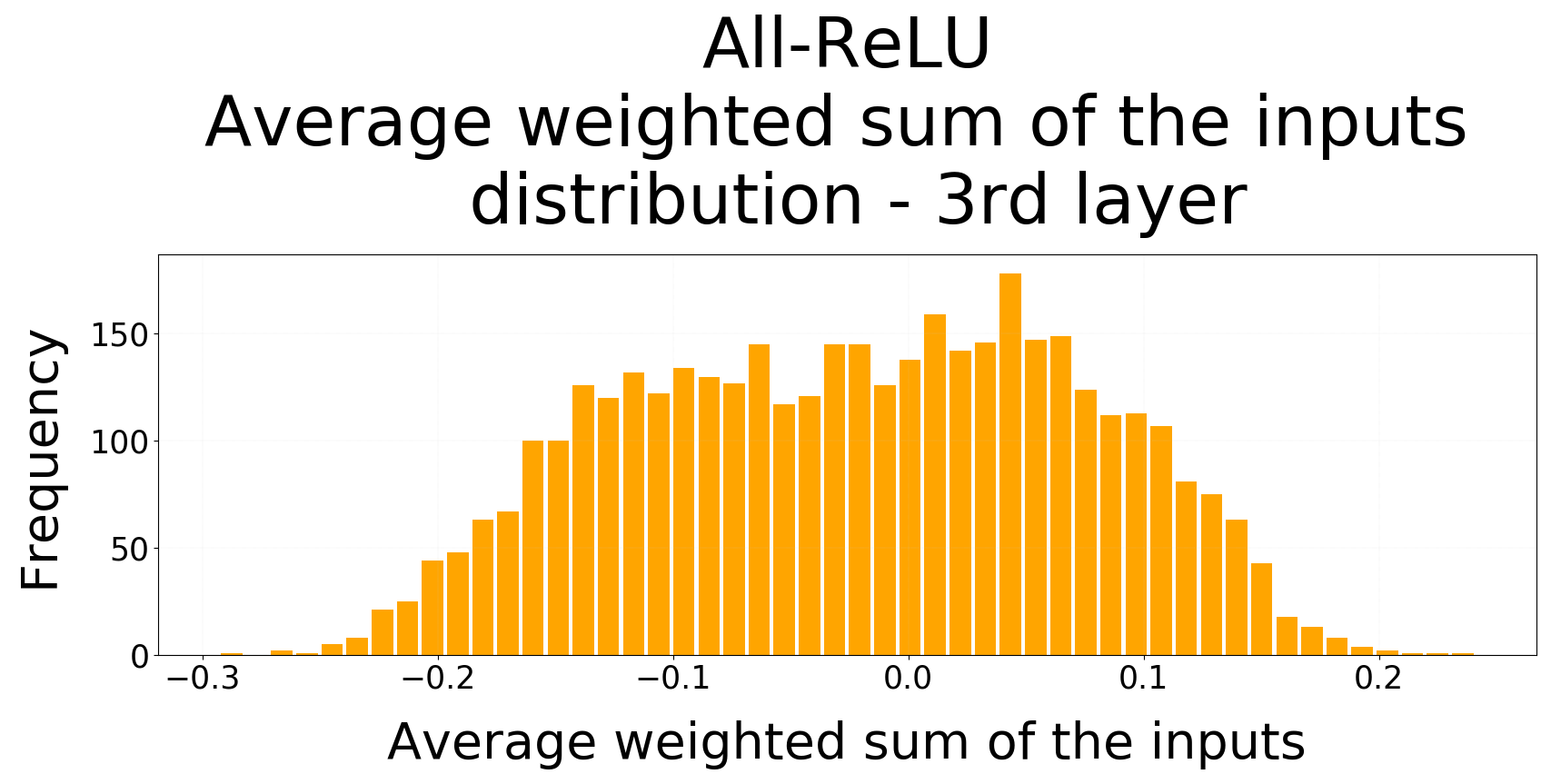}}
\caption{Average weighted sum of the inputs distribution for a 3-layers sparse MLP on CIFAR10 trained with All-ReLU and \textit{momentum} SGD after 1000 epochs.}
\label{fig:activations_5}
\end{figure*}

\begin{figure*}[!htbp]
\subfloat{%
\includegraphics[width=0.3\textwidth]{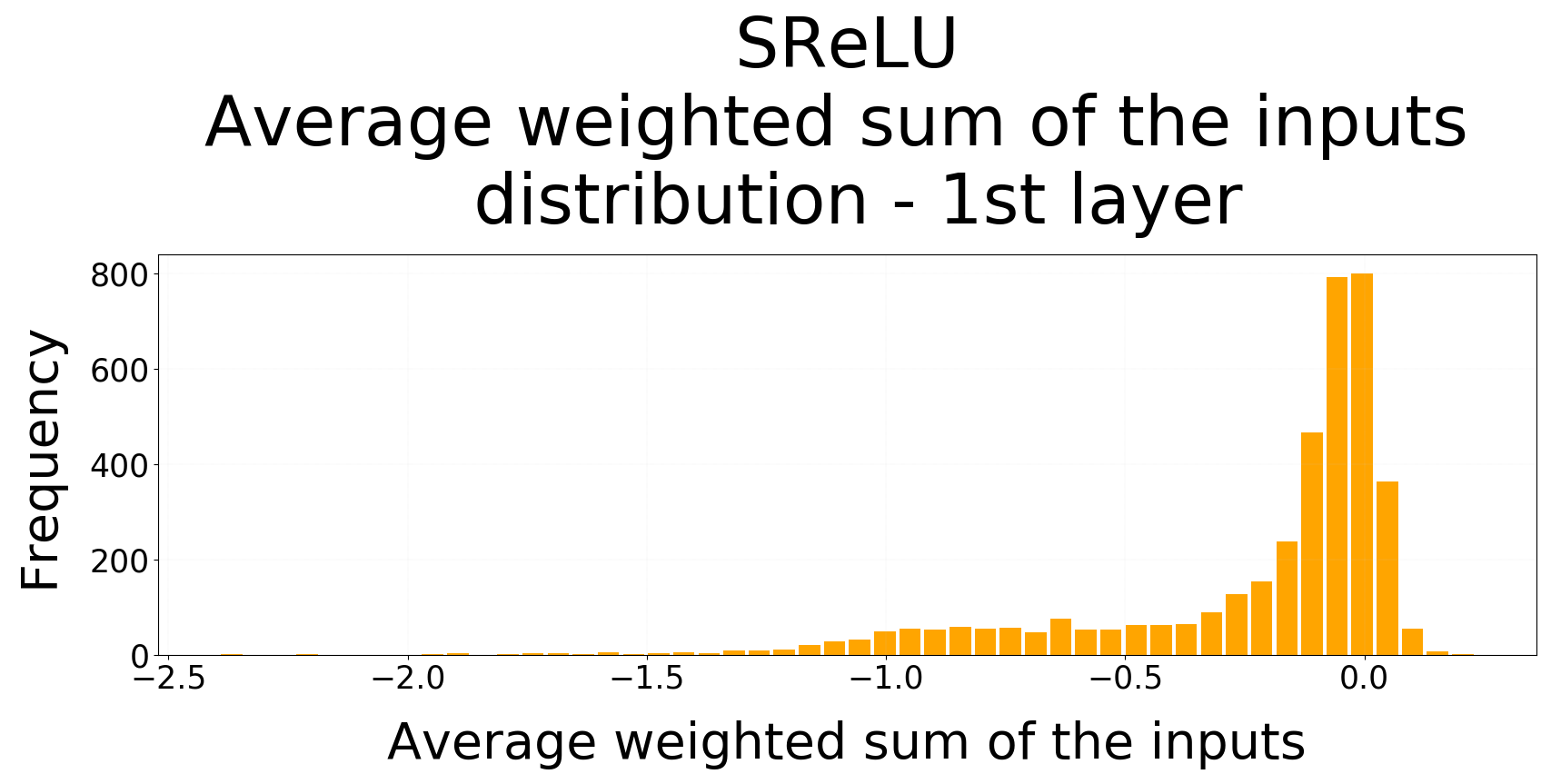}}
\hspace{\fill}
\subfloat{%
\includegraphics[width=0.3\textwidth]{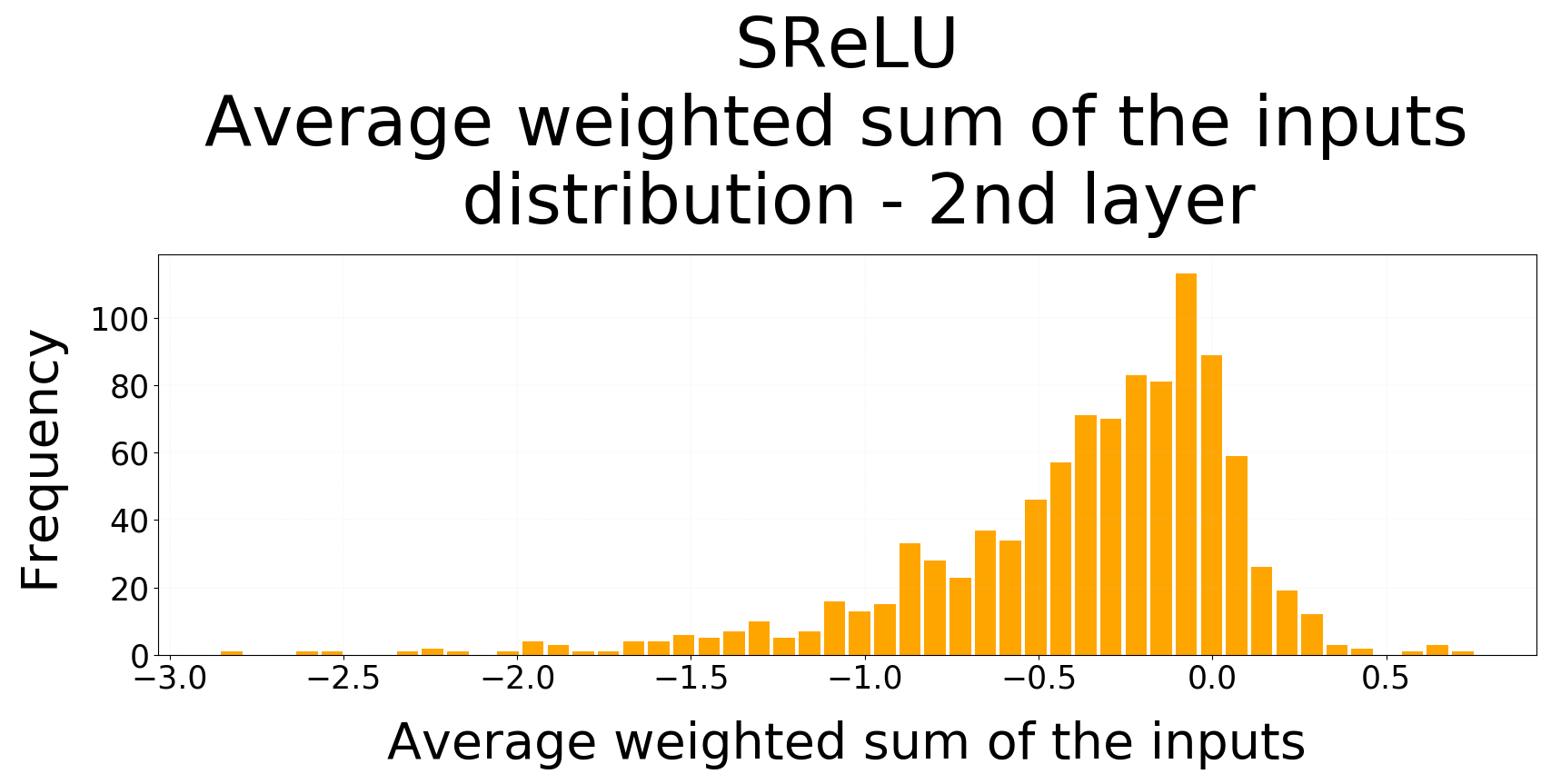}}
\hspace{\fill}
\subfloat{%
\includegraphics[width=0.3\textwidth]{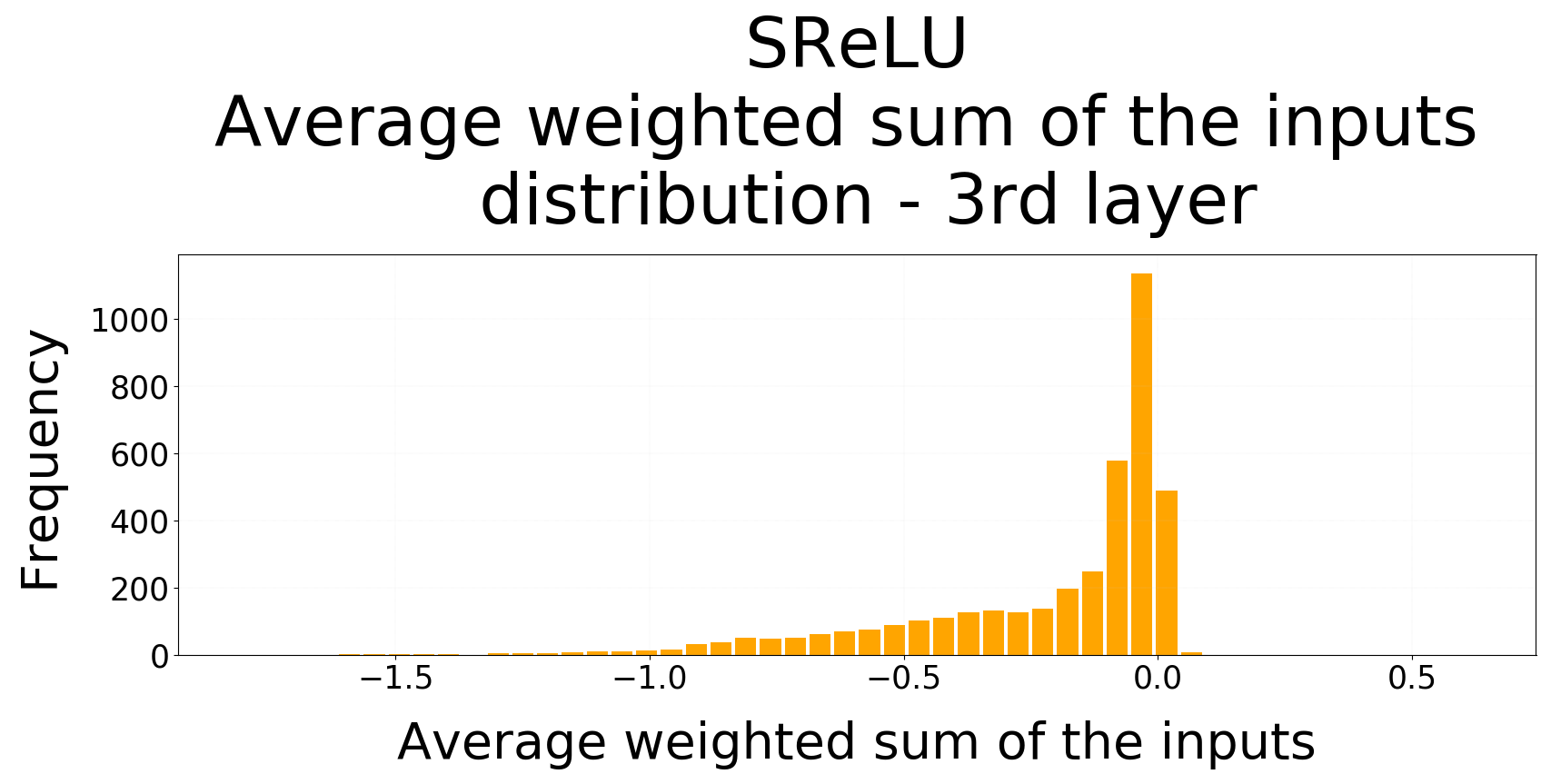}}
\caption{Average weighted sum of the inputs distribution for a 3-layers sparse MLP on CIFAR10 trained with SReLU and \textit{momentum} SGD after 1000 epochs.}
\label{fig:activations_6}
\end{figure*}

\begin{figure*}[!htbp]
\subfloat{%
\includegraphics[width=0.3\textwidth]{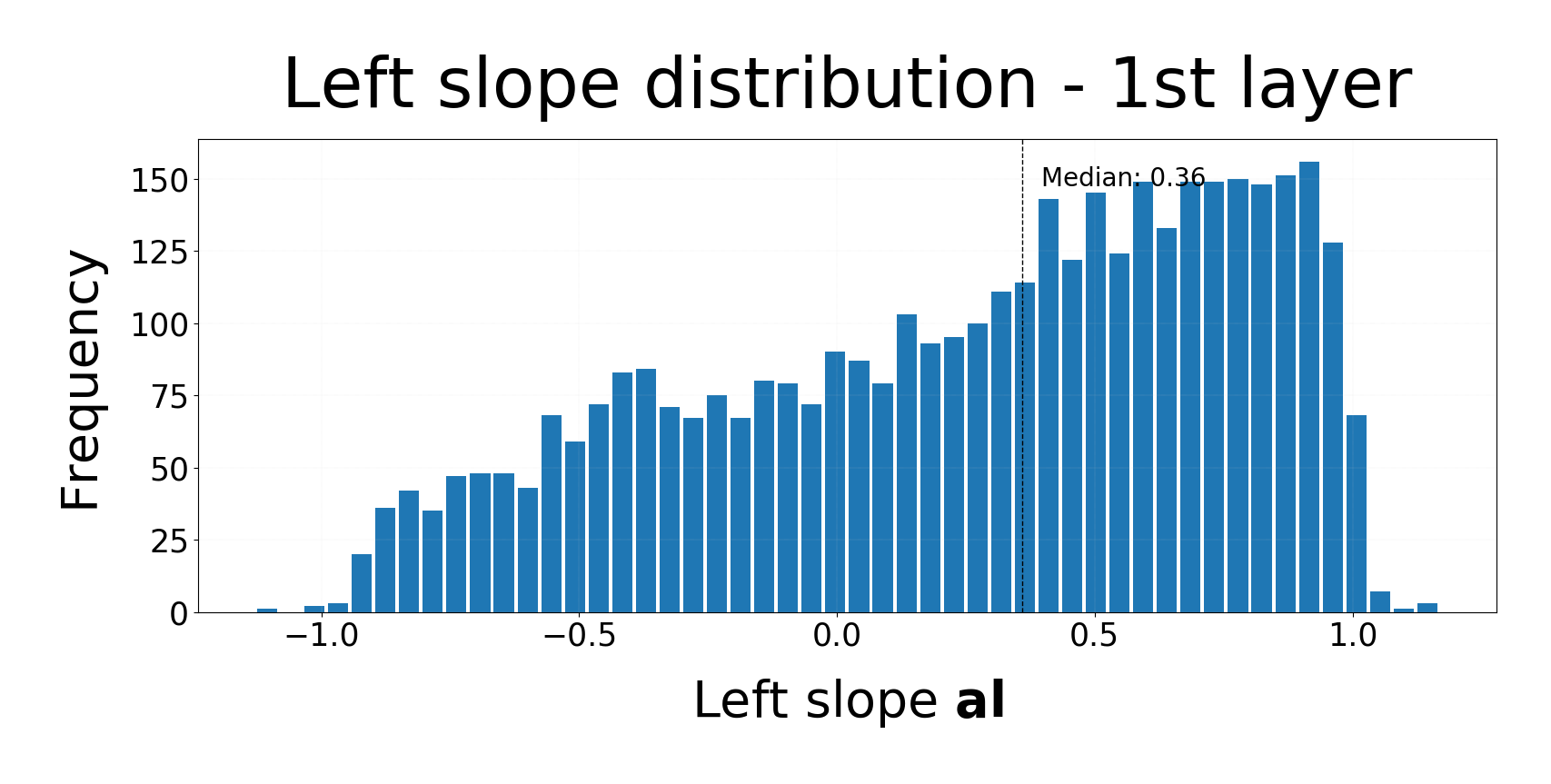}}
\hspace{\fill}
\subfloat{%
\includegraphics[width=0.3\textwidth]{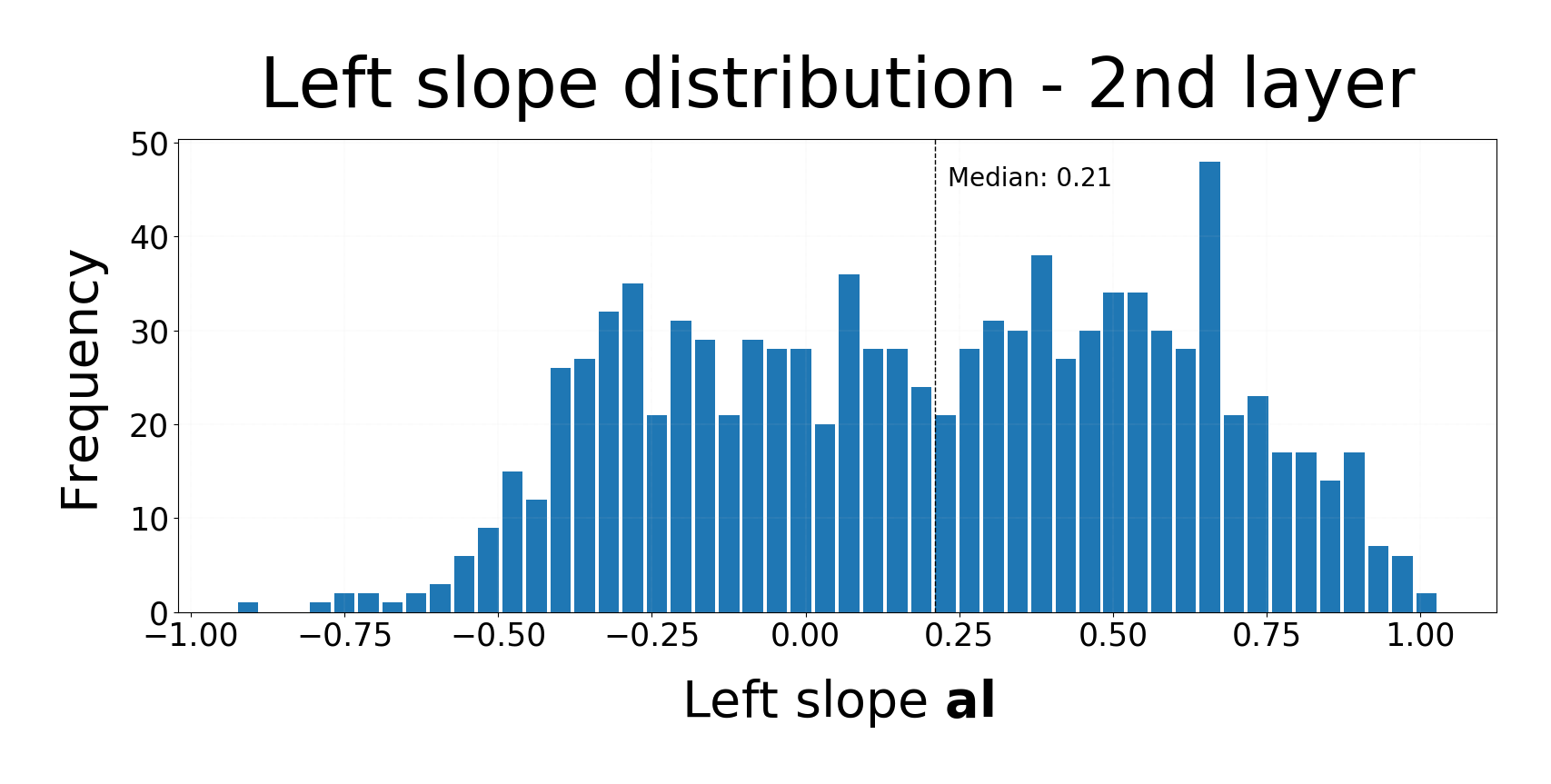}}
\hspace{\fill}
\subfloat{%
\includegraphics[width=0.3\textwidth]{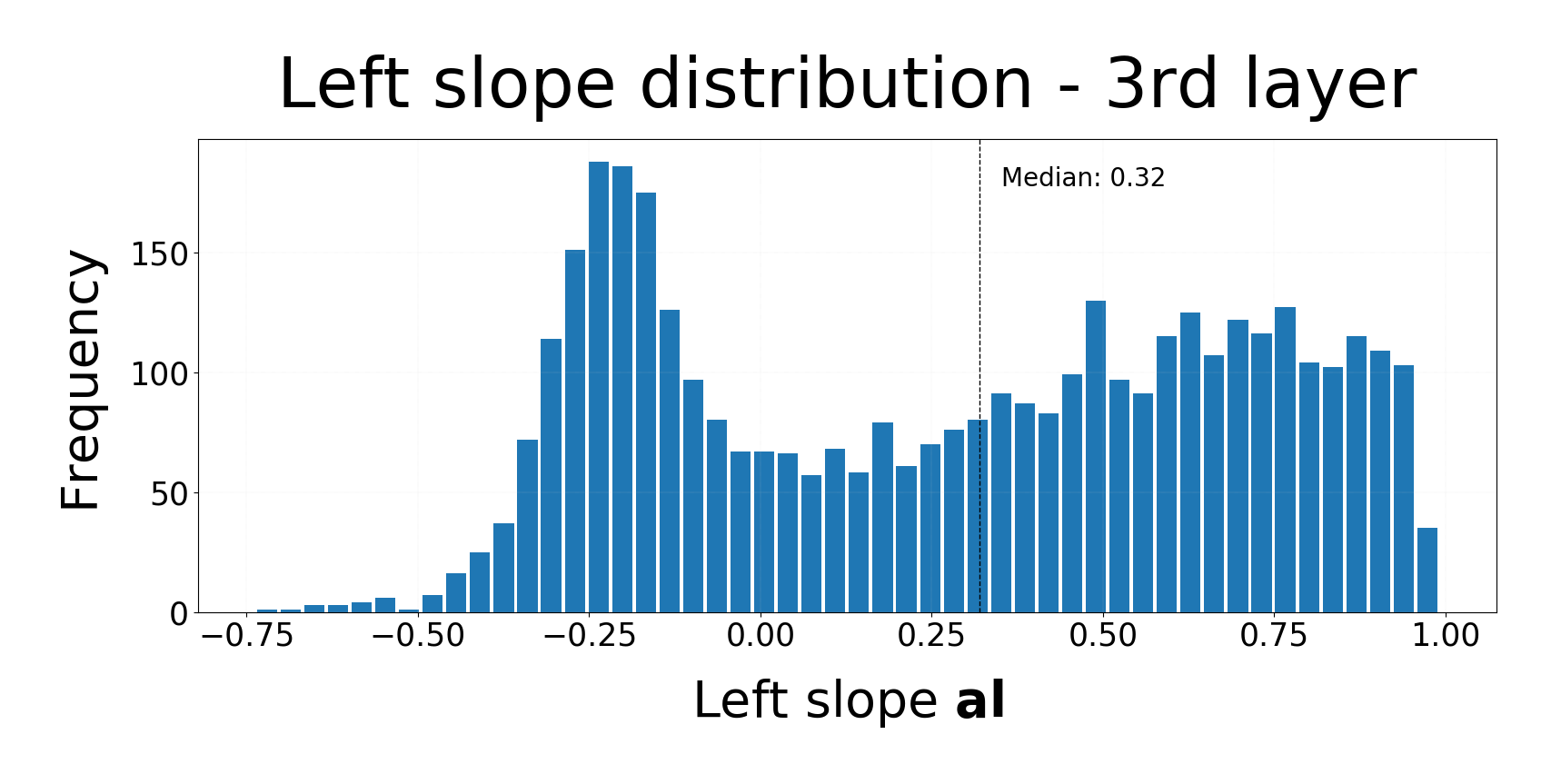}}
\caption{Left slopw $a_l$ distribution for a 3-layers sparse MLP on CIFAR10 trained with SReLU and \textit{momentum} SGD after 1000 epochs.}
\label{fig:activations_7}
\end{figure*}

\begin{figure*}[!htbp]
\subfloat{%
\includegraphics[width=0.3\textwidth]{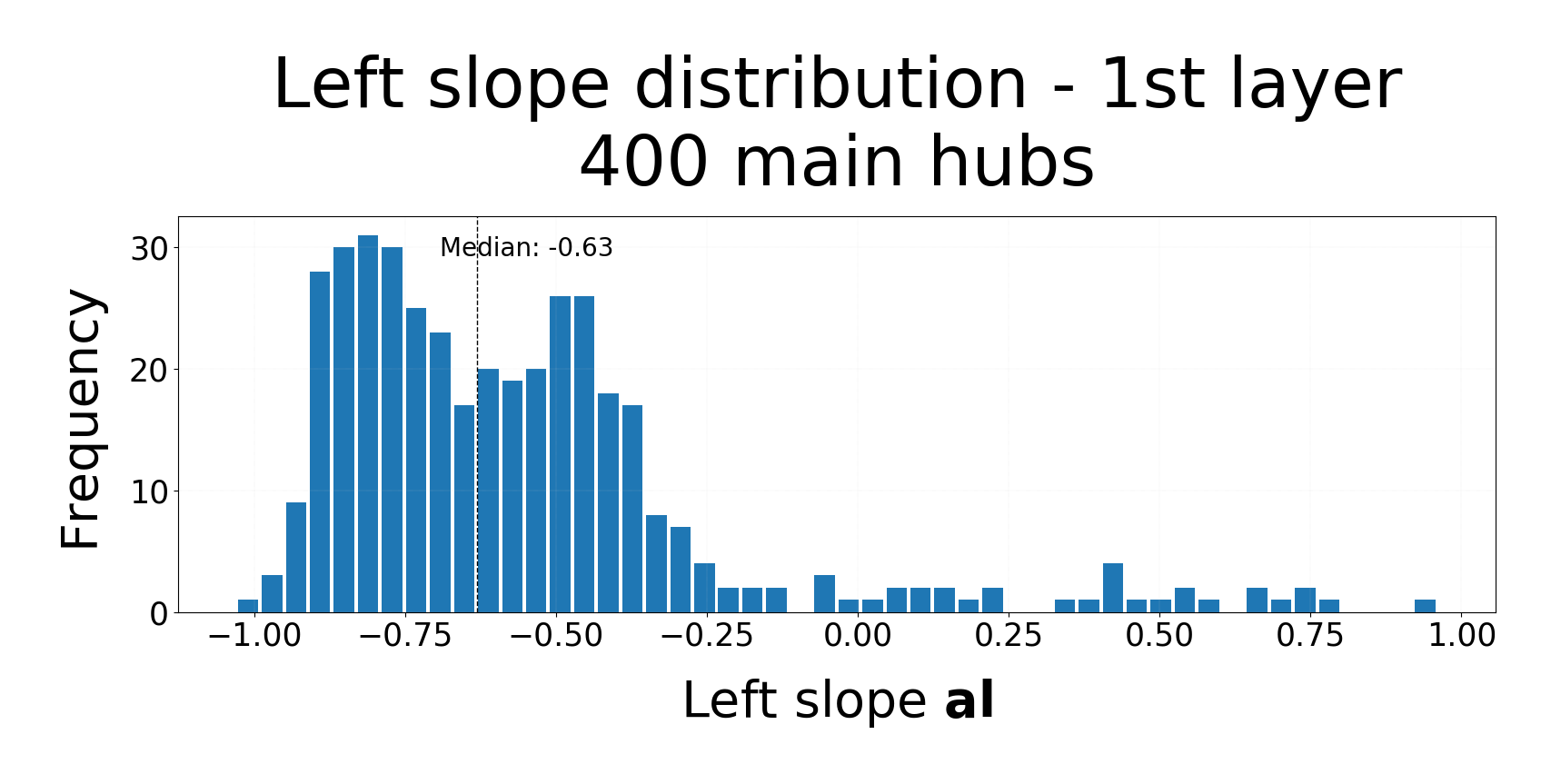}}
\hspace{\fill}
\subfloat{%
\includegraphics[width=0.3\textwidth]{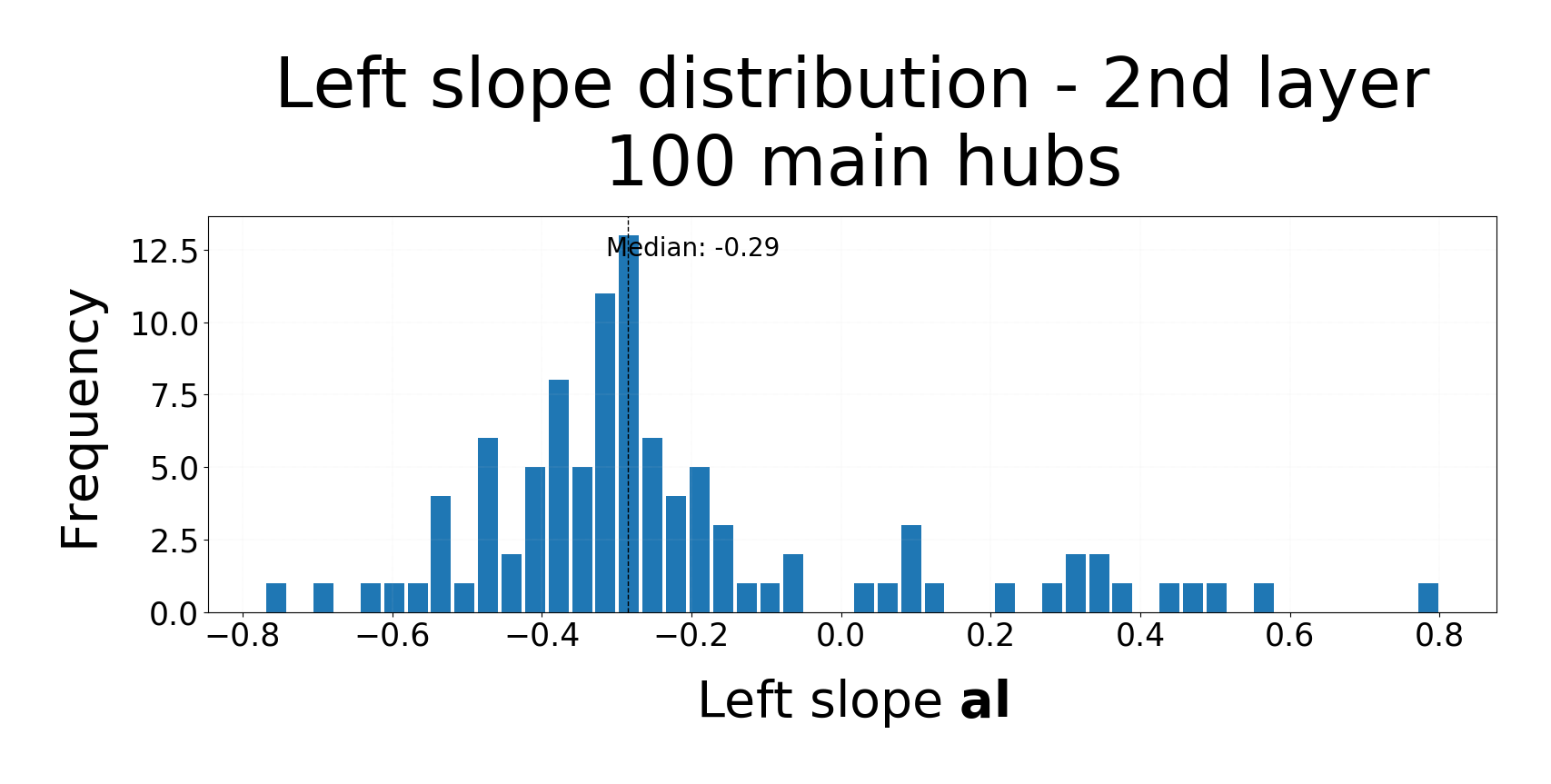}}
\hspace{\fill}
\subfloat{%
\includegraphics[width=0.3\textwidth]{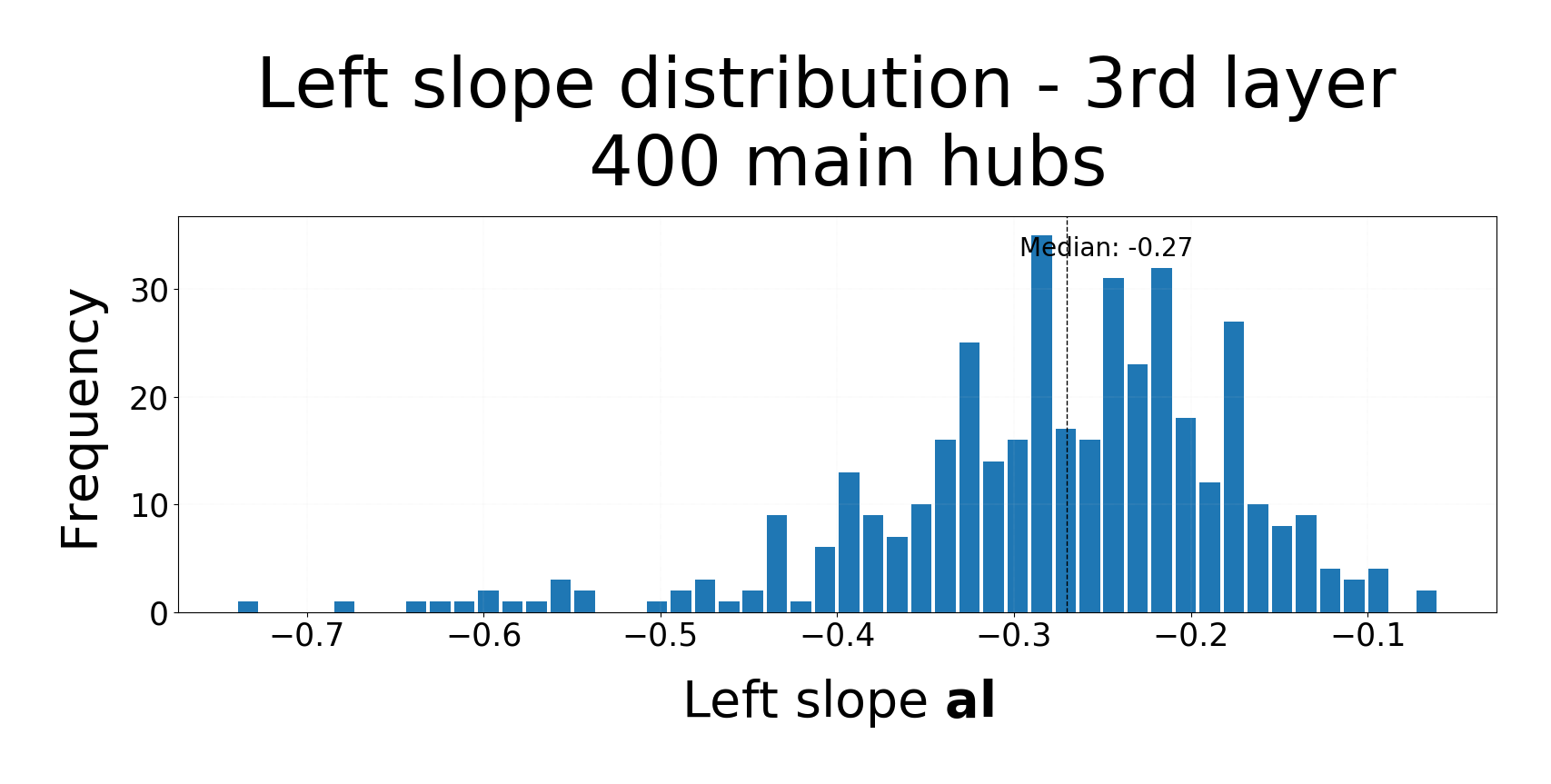}}
\caption{Left slopw $a_l$ distribution for a 3-layers sparse MLP on CIFAR10 trained with SReLU and \textit{momentum} SGD after 1000 epochs.}
\label{fig:activations_8}
\end{figure*}

\begin{figure*}[!htbp]
\subfloat{%
\includegraphics[width=0.3\textwidth]{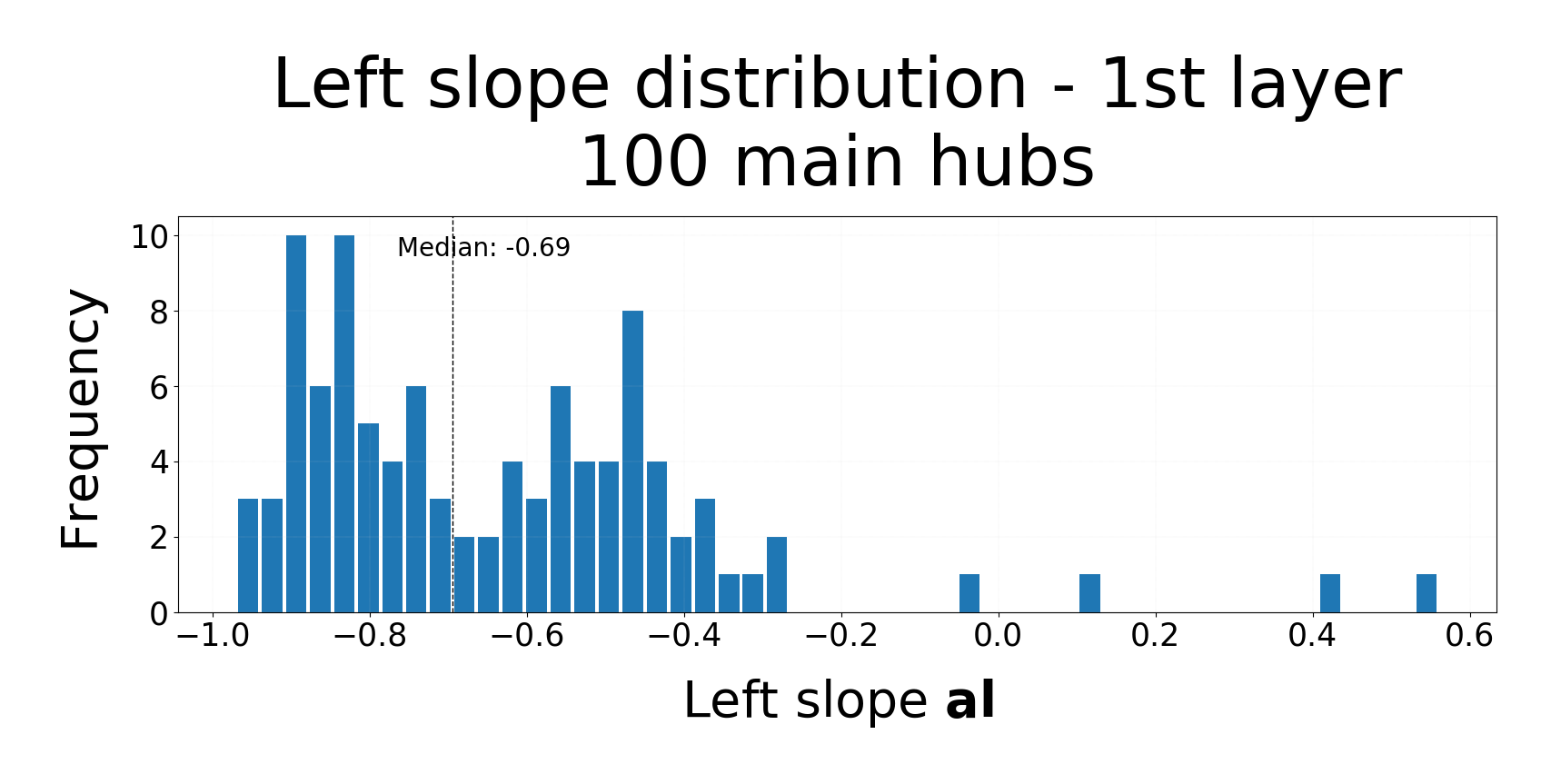}}
\hspace{\fill}
\subfloat{%
\includegraphics[width=0.3\textwidth]{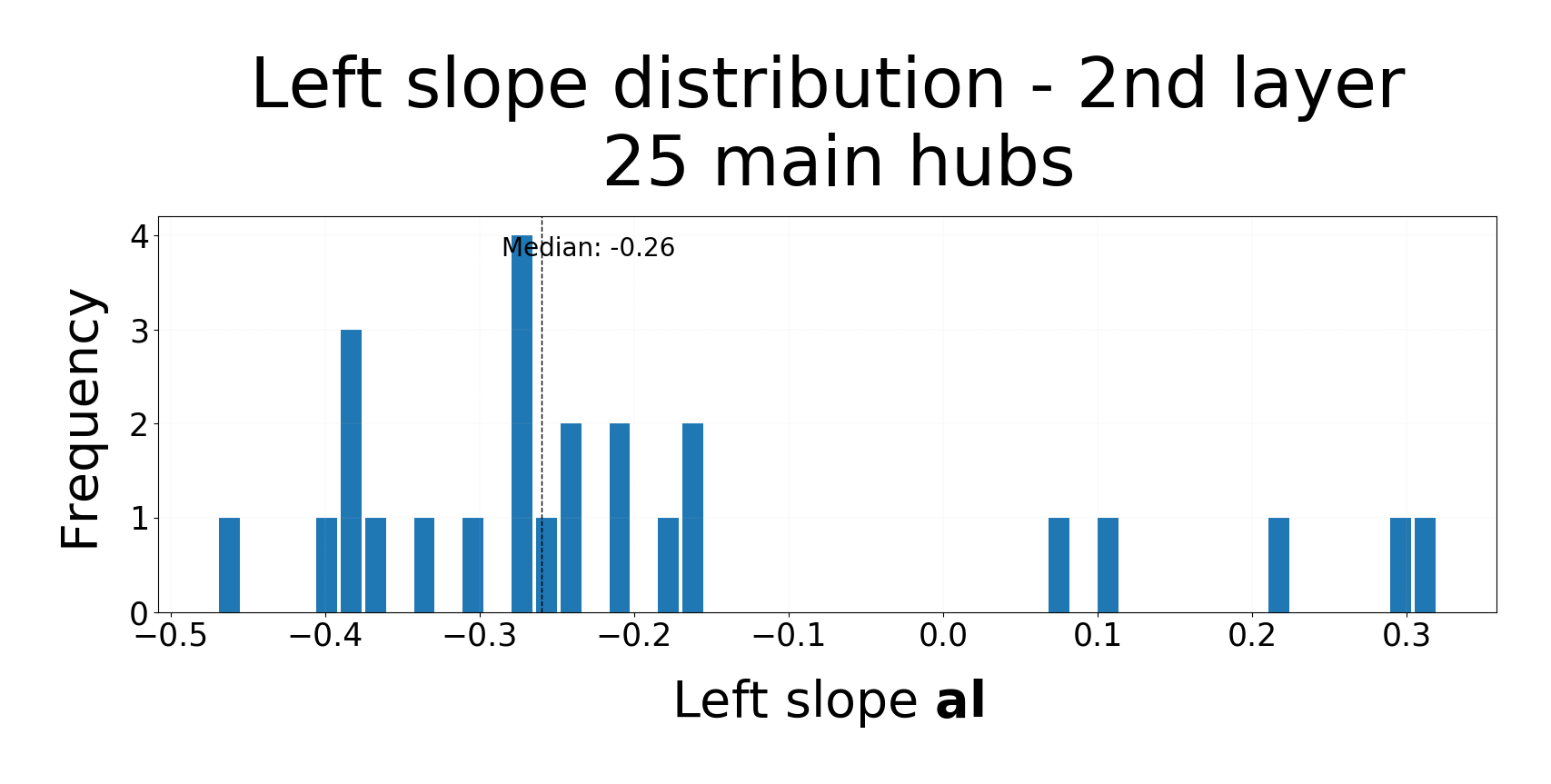}}
\hspace{\fill}
\subfloat{%
\includegraphics[width=0.3\textwidth]{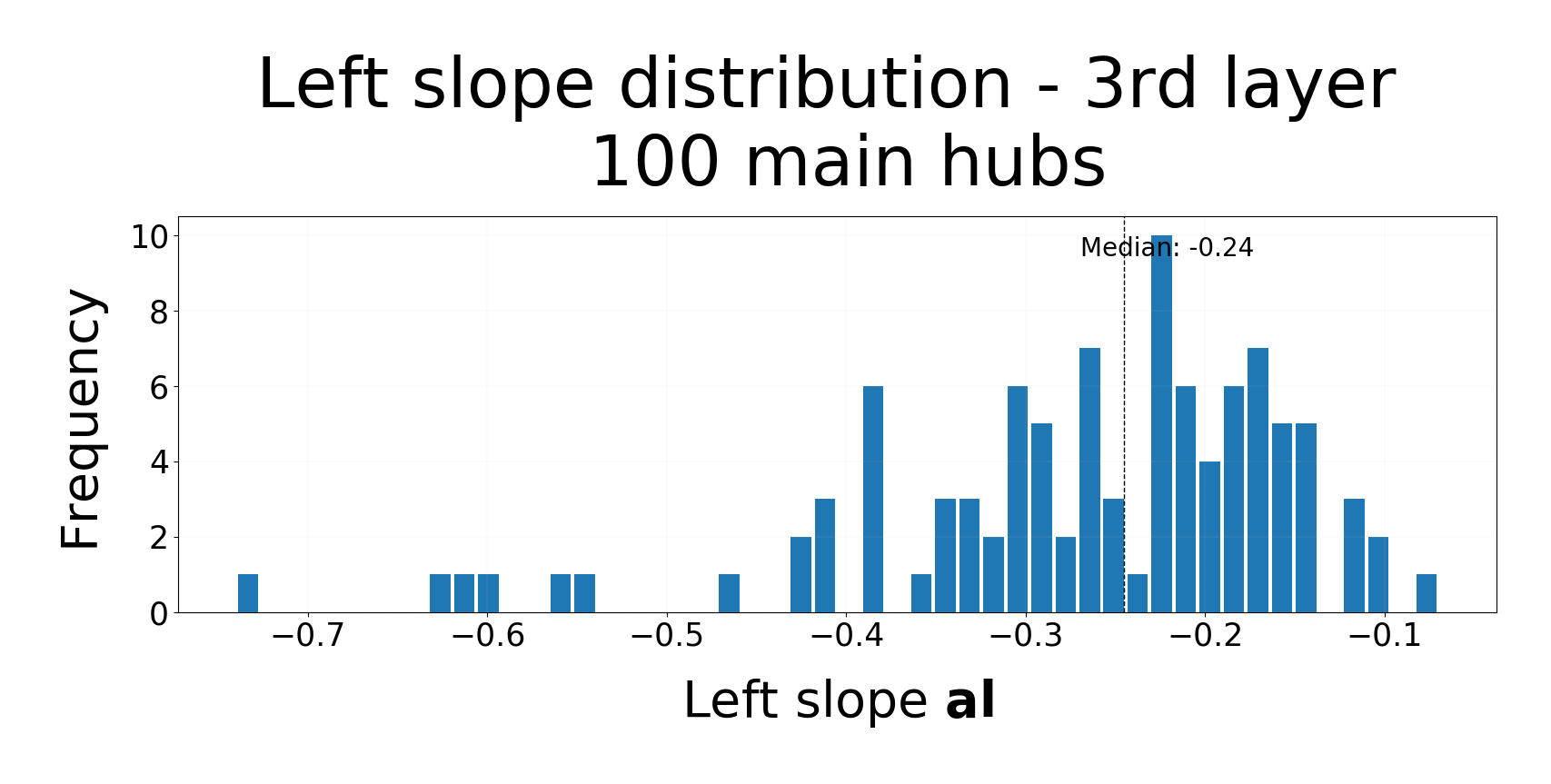}}
\caption{Left slopw $a_l$ distribution for a 3-layers sparse MLP on CIFAR10 trained with SReLU and \textit{momentum} SGD after 1000 epochs.}
\label{fig:activation_9}
\end{figure*}

\begin{figure*}[!htbp]
\subfloat{%
\includegraphics[width=0.3\textwidth]{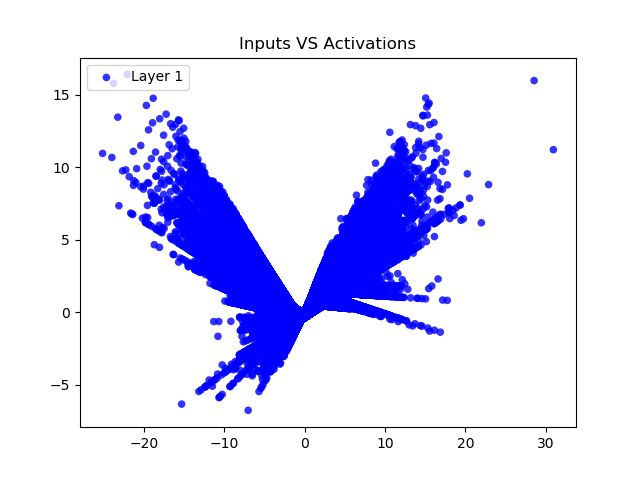}}
\hspace{\fill}
\subfloat{%
\includegraphics[width=0.3\textwidth]{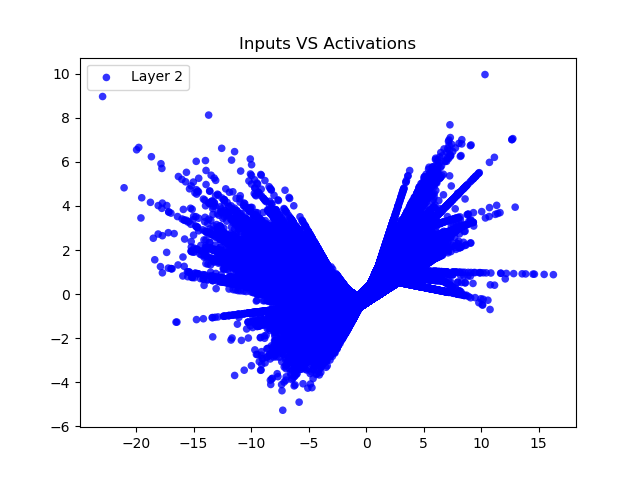}}
\hspace{\fill}
\subfloat{%
\includegraphics[width=0.3\textwidth]{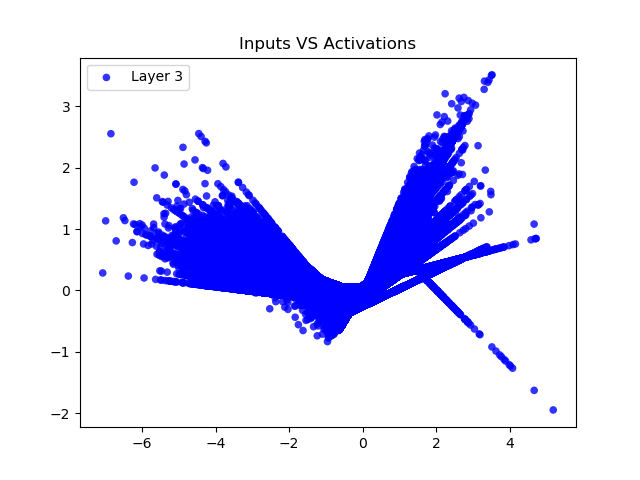}}
\caption{Input vs Activations for a 3-layers sparse MLP on CIFAR10 trained with SReLU and \textit{momentum} SGD after 1000 epochs.}
\label{fig:activations_10}
\end{figure*}

\newpage

\subsection{Finding the Best Slope for All-ReLU}
\label{subsection:gridsearch}

\begin{figure}[!htbp]
\centering
\includegraphics[width=0.8\textwidth]{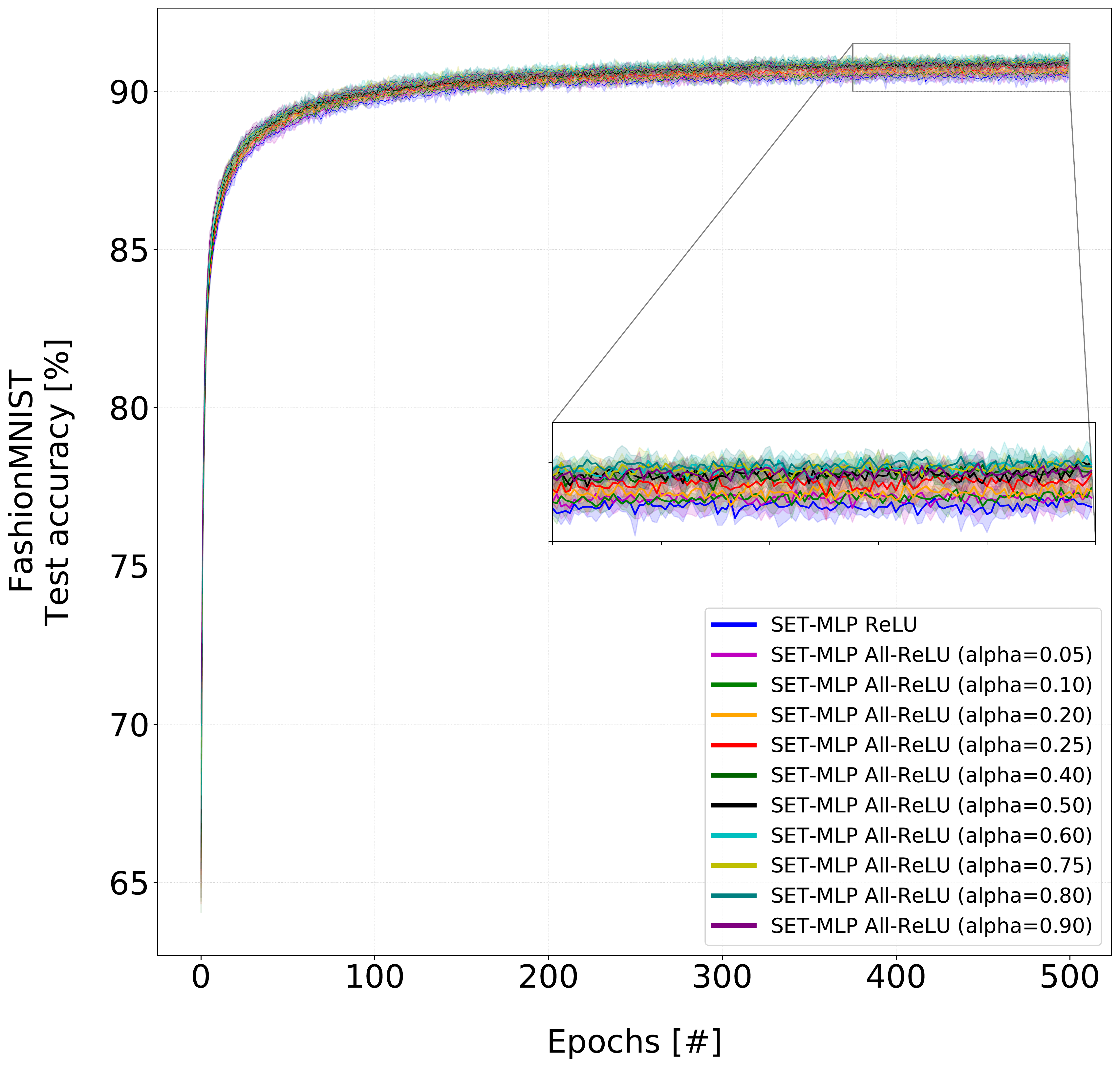}
\caption{Grid Search for slope $\alpha$ on FashionMNIST for the proposed All-ReLU activation function. The plot reports the accuracy on the test dataset when training the various SET MLPs models for 500 epochs. Each run is repeated five times and averaged to visualize the learning curves.}
\label{tuningslope}
\end{figure}

In this appendix, we illustrate the tuning of the hyperparameter $\alpha$ required by the proposed All-ReLU function when considering the FashionMNIST dataset. In Figure \ref{tuningslope}, we report the learning curve for different values of the slope, averaged over five runs, while in Table \ref{table:slopesearch} are shown the best results in term of accuracy. The best value of $\alpha$ for training a SET model on FashionMNIST is 0.6. To avoid the expensive grid search, we want to propose a practical method to narrow down the choices of this parameter. Based on our intuition, we propose to pick the slope $\alpha$ based on the observed skewness of the input data distribution. When the input is skewed on the left side, the chosen slope should reflect the level of distortion or asymmetry in the data distribution.

\begin{table}[!htbp]
\centering
\resizebox{\textwidth}{!}{%
\setlength{\tabcolsep}{0.8em} 
{\renewcommand{\arraystretch}{1.6}
\begin{tabular}{|c|c|c|c|c|c|c|c|c|c|c|}
\hline
& $\mathbf{\alpha=0}$ & $\mathbf{\alpha=0.5}$ & $\mathbf{\alpha=0.1}$ & $\mathbf{\alpha=0.2}$ & $\mathbf{\alpha=0.25}$ & $\mathbf{\alpha=0.5}$ & $\mathbf{\alpha=0.6}$ & $\mathbf{\alpha=0.75}$ & $\mathbf{\alpha=0.8}$ & $\mathbf{\alpha=0.9}$ \\ \hline
\textbf{Best accuracy [\%]} & 90.48 & 90.86 & 90.87 & 90.75 & 91 & 91.08 & 91.38 & 91.09 & 91.16 & 91.02 \\ \hline
\end{tabular}}}
\caption{The table reports the best accuracy obtained by the sparse models on the test dataset when adopting All-ReLU for different values of $\alpha$.}
\label{table:slopesearch}
\end{table}

It is essential to notice that any alpha level greater than 0.05 leads to better results than ReLU. Hence, even with suboptimal values is possible to attain satisfactory results.

\newpage
\subsection{Importance Pruning Post-Training}
\label{subsection:posttraining}
The proposed method \textit{Importance Pruning} based on our neuron importance metric can be easily applied one time only, once the sparse training procedure is concluded. With the experiments reported in Table \ref{table:postpruning}, we empirically demonstrate that the \textbf{Importance Pruning} technique should be integrated during training to gain the best results in terms of the memory footprint, running time and accuracy. 
\begin{table}[!htbp]
\resizebox{\textwidth}{!}{%
\begin{tabular}{llllll}

\toprule
\textbf{Dataset} & \textbf{Model Accuracy [\%]} &
\textbf{Parameters [\#]} &
\textbf{Pruning threshold \textit{t}} & \textbf{Results} & \\ \cmidrule{5-6}
& & & &

\textbf{Accuracy [\%]} & \textbf{$\mathbf{end\_n^W}$ [\#]} 

\\ \midrule
\textbf{Leukemia} & 86.42 & 1684944 & \textbf{5}\textsuperscript{th} percentile & \textbf{86.89} & 1074251 \\
& & & \textbf{10}\textsuperscript{th} percentile & 86.69 & 1034977 \\
& & & \textbf{15}\textsuperscript{th} percentile & 85.84 & 992711 \\
& & & \textbf{20}\textsuperscript{th} percentile & 85.84 & 947981 \\
& & & \textbf{25}\textsuperscript{th} percentile & 85.69 & 901289 \\

\textbf{Higgs} & 73.67 & 50224 & \textbf{5}\textsuperscript{th} percentile & 73.00 & 48508 \\
& & & \textbf{10}\textsuperscript{th} percentile & \textbf{73.16} & 46693 \\
& & & \textbf{15}\textsuperscript{th} percentile & 72.61 & 44897 \\
& & & \textbf{20}\textsuperscript{th} percentile & 72.27 & 42601 \\
& & & \textbf{25}\textsuperscript{th} percentile & 71.51 & 40491 \\

\textbf{Madelon} & 71.33 & 19006 & \textbf{5}\textsuperscript{th} percentile & 71.33 & 18385 \\
& & & \textbf{10}\textsuperscript{th} percentile & \textbf{72.00} & 17661 \\
& & & \textbf{15}\textsuperscript{th} percentile & 70.33 & 16883 \\
& & & \textbf{20}\textsuperscript{th} percentile & 69.16 & 16106 \\
& & & \textbf{25}\textsuperscript{th} percentile & 65.33 & 15297 \\

\textbf{FashionMNIST} & 91.38 & 125901 & \textbf{5}\textsuperscript{th} percentile & \textbf{90.66} & 120894 \\
& & & \textbf{10}\textsuperscript{th} percentile & 90.12 & 115486 \\
& & & \textbf{15}\textsuperscript{th} percentile & 89.37 & 109923 \\
& & & \textbf{20}\textsuperscript{th} percentile & 88.7 & 104158 \\
& & & \textbf{25}\textsuperscript{th} percentile & 85.97 & 98360 \\

\textbf{CIFAR10} & 69.83 & 381758 & \textbf{5}\textsuperscript{th} percentile & \textbf{69.53} & 368062 \\
& & & \textbf{10}\textsuperscript{th} percentile & 68.61 & 352791 \\
& & & \textbf{15}\textsuperscript{th} percentile & 68.38 & 336968 \\
& & & \textbf{20}\textsuperscript{th} percentile & 67.35 & 320316 \\
& & & \textbf{25}\textsuperscript{th} percentile & 66.57 & 276349 \\

\bottomrule
\end{tabular}}
\vspace{5pt}
\caption{On each dataset, we report the classification accuracy obtained by each model on the test data when the pruning method based on neuron importance is applied solely once at the end of the training procedure. The models employed for this evaluation are the resulting sparse MLPs from Table \ref{table:results} trained with All-ReLU and no pruning. \textit{Parameters [\#]} and \textit{Model Accuracy [\%]} represent, respectively, the number of weights in the model and the final accuracy at the end of the training, while $end\_n^W$ represents the number of parameters in the pruned model. }
\label{table:postpruning}
\end{table}

Here, the pruning threshold \textit{t} is used to remove all neurons and related connection which have an importance value lower than \textit{t}. 
The post-pruning is carried out multiple times for different values of \textit{t} (pruning threshold) to observe the loss in accuracy as the threshold increases. If we compare the results with the one obtained in Table \ref{table:results} when importance pruning is employed while training, we can notice that we are not able to eliminate a significant number of parameters without reflecting a remarkable drop in accuracy. Hence, the \textit{Importance Pruning} technique has shown to have more advantages if integrated during training.

\newpage
\subsection{Experiments setup}
\subsubsection{Datasets}
We evaluate and discuss the performance of our proposed methods on sparse MLP models by considering five publicly available datasets listed in Table \ref{table:datasets}. We report both the number of data points used as training data and as testing data. The \textit{Leukemia} dataset is obtained from the NCBI GEO repository (\cite{ncbi}) with the accession number GSE13159. In (\cite{liu2019sparse}), the authors provide a table with class labels (unbalanced) and their corresponding number of test samples. \textit{HIGGS} (\cite{higgs}) dataset is a classification problem to distinguish between a signal process which produces Higgs bosons and a background process which does not. The data has been produced using Monte Carlo simulations. Due to the limited resources available, we merely considered a fraction of the full dataset. \textit{Madelon} (\cite{madelon}) is an artificial dataset that has five informative features and 15 linear combinations of those features. The other 480 features are probes that provide no information about the class label. Lastly, FashionMNIST (\cite{fashionmnist}) and CIFAR10 (\cite{cifar10}) are both image datasets.

\subsubsection{Implementation details}
The following environment has been selected to implement the parallel algorithm WASAP-SGD, based on the implementation provided by \cite{anderson2017mpibased} and \cite{liu2019sparse}:
\begin{description}
\item[Language:] Pure Python 3.7 for quick prototyping, where \textit{SciPy} and \textit{Numpy} are employed for sparse matrix operations while \textit{Numba} accelerates some critical part of the code such as backpropagation.
\item[Framework:] Message Passing Interface (MPI) standard.
\item[Library:] \texttt{mpi4py}, the library is constructed on top of the MPI-1/2 specifications and provides an object-oriented interface which directly follows MPI-2 C++ bindings.
\end{description}
All the experiments performed in this chapter are executed on a typical
laptop with the following configuration:
\begin{itemize}
\item Hardware configuration: CPU Intel Core i7-9750H, 2.60 GHz $\times$ 6, RAM 32 GB, Hard disk 1000 GB, NVIDIA GeForce GTX 1650 4GB.
\item Software used: Windows 10, Python 3.7, Numpy
1.19.1, SciPy 1.4.1, and Numba 0.48.0.
\end{itemize}

\begin{table}[!htbp]
\small
\centering
\begin{tabular}{lllllll}
\toprule
\textbf{Experiment} & \textbf{Dataset} & & & &\textbf{Hyper-parameters} & \\ \cmidrule{3-7}
& & $\mathbf{\epsilon}$ & $\mathbf{\eta}$ & $\mathbf{\mathcal{B}}$ & \textbf{Weight initialization} & $\mathbf{\alpha}$ \\ \midrule
\textbf{MLPs} & Leukemia & 10 & 0.005 & 5 & \textit{normal} & 0.75\\
& Higgs & 10 & 0.01 & 128 & \textit{xavier} & 0.05 \\
& Madelon & 10 & 0.01 & 32 & \textit{normal} & 0.5 \\ 
& FashionMNIST & 20 & 0.01 & 128 & \textit{he uniform} & 0.6 \\ 
& CIFAR10 & 20 & 0.01 & 128 &\textit{he uniform} & 0.75 \\ 
\bottomrule
\end{tabular}
\caption{List of hyperparameters used for the experiments.}
\label{table:hyperparameters}
\end{table}

\subsubsection{Experiments hyperparameters}
We primarily used the same configuration of parameters as in the original SET paper (\cite{Mocanu_2018}). Like SET, our methods were also tested on a multilayer perceptron, in which the fully connected layers have been replaced with sparse layers. The MLP models in \autoref{subsection:mlps} are trained with sequential momentum SGD (momentum is set to 0.9), weight decay, a dropout rate of 0.3 and fixed learning schedule. In \autoref{table:hyperparameters} we provide an overview of the primary hyperparameters. The slope $\alpha$ for All-ReLU has been mostly identified via grid search. Similarly, the epoch $\tau$, that determines the starting point of Importance Pruning is determined based on a local search within a limited set of value, and it is set to 200 for all models. The fraction $\zeta$ of the smallest positive weights and the largest negative weights to be removed is always set to $0.3$. Here, $\epsilon$ controls the sparsity level as discussed in \cite{Mocanu_2018}, $\eta$ is the learning rate and $\mathcal{B}$ the batch size. Furthermore, to improve the learning process of our networks, we standardise the features of our datasets such that each attribute has zero mean and unit variance. The models in \autoref{subsection:parallelmlps} employ the same hyperparameters configurations, but the training procedure is parallelised with WASAP-SGD.

\end{document}